\documentclass{article}

\usepackage[preprint, nonatbib]{neurips_2024}

\usepackage[utf8]{inputenc} % allow utf-8 input
\usepackage[T1]{fontenc}    % use 8-bit T1 fonts
\usepackage{hyperref}       % hyperlinks
\usepackage{url}            % simple URL typesetting
\usepackage{booktabs}       % professional-quality tables
\usepackage{amsfonts}       % blackboard math symbols
\usepackage{nicefrac}       % compact symbols for 1/2, etc.
\usepackage{microtype}      % microtypography
\usepackage{xcolor}         % colors
\usepackage{amsmath}
\usepackage[pdftex]{graphicx}
\usepackage{subcaption}
\definecolor{cvprblue}{rgb}{0.21,0.49,0.74}
\usepackage{multirow}
\usepackage{tabu}
\usepackage{xcolor,colortbl}
\usepackage{wrapfig}
\usepackage{amssymb}
\usepackage{algorithm,algpseudocode}
\algrenewcommand\algorithmicrequire{\textbf{Input:}}
\algrenewcommand\algorithmicensure{\textbf{Output:}}

\definecolor{tabfirst}{rgb}{1, 0.7, 0.7} % red
\definecolor{tabsecond}{rgb}{1, 0.85, 0.7} % orange
\definecolor{tabthird}{rgb}{1, 1, 0.7} % yellow
\definecolor{grey}{RGB}{80,80,80}
\newcommand{\method}{Sp\(^2\)360}

\title{\method: Sparse-view 360$^{\circ}$ Scene Reconstruction\\using Cascaded 2D Diffusion Priors}

\author{%
  Soumava Paul\textsuperscript{1, 2}, Christopher Wewer\textsuperscript{1}, Bernt Schiele\textsuperscript{1}, Jan Eric Lenssen\textsuperscript{1} \\
  {\textsuperscript{1} Max Planck Institute for Informatics, Saarland Informatics Campus, Germany}\\
  {\textsuperscript{2}Saarland University, Saarland Informatics Campus, Germany}\\
  {\tt\small soumava2016@gmail.com}, {\tt\small\{cwewer, schiele, jlenssen\}@mpi-inf.mpg.de}
}

\begin{document}

\maketitle

\begin{abstract}
  We aim to tackle sparse-view reconstruction of a $360^{\circ}$ 3D scene using priors from latent diffusion models (LDM). The sparse-view setting is ill-posed and underconstrained, especially for scenes where the camera rotates $360$ degrees around a point, as no visual information is available beyond some frontal views focused on the central object(s) of interest. In this work, we show that pretrained 2D diffusion models can strongly improve the reconstruction of a scene with low-cost fine-tuning. Specifically, we present \emph{SparseSplat360 (\method{})}, a method that employs a cascade of in-painting and artifact removal models to fill in missing details and clean novel views. Due to superior training and rendering speeds, we use an explicit scene representation in the form of 3D Gaussians over NeRF-based implicit representations. We propose an iterative update strategy to fuse generated pseudo novel views with existing 3D Gaussians fitted to the initial sparse inputs. As a result, we obtain a multi-view consistent scene representation with details coherent with the observed inputs. Our evaluation on the challenging Mip-NeRF360 dataset shows that our proposed 2D to 3D distillation algorithm considerably improves the performance of a regularized version of 3DGS adapted to a sparse-view setting and outperforms existing sparse-view reconstruction methods in $360^{\circ}$ scene reconstruction. Qualitatively, our method generates entire $360^{\circ}$ scenes from as few as $9$ input views, with a high degree of foreground and background detail.

  %\bernt{the term $360^{\circ}$ scenes is used often in the abstract - I have an idea what it might mean - but is that a term that everyone knows exactly what this is? If not, we need to define in the abstract shortly (e.g.) with a half-sentence or at least in the very first lines of the intro}
\end{abstract}
\vspace{-4.0mm}
\section{Introduction}
Obtaining high-quality 3D reconstructions or novel views from a set of images has been a long-standing goal in computer vision and has received increased interest recently.
Recent 3D reconstruction methods, such as those based on Neural Radiance Fields (NeRF)~\cite{nerf}, Signed Distance Functions (SDFs)~\cite{neus}, or the explicit 3D Gaussian Splatting (3DGS)~\cite{3dgs}, are now able to produce photorealistic novel views of $360^{\circ}$ scenes.
However, in doing so, they rely on hundreds of input images that densely capture the underlying scene.  This requirement is both time-consuming and often an unrealistic assumption for complex, large-scale scenes. Ideally, one would like a 3D reconstruction pipeline to offer generalization to unobserved parts of the scene and be able to successfully reconstruct areas that are only observed a few times. In this work, we present a method to efficiently obtain high-quality 3D Gaussian representations from just a few views, moving towards this goal.
%Neural Radiance Fields (NeRF) \cite{nerf} introduced a breakthrough in the field of differentiable rendering and learning 3D representations from 2D images, enabling the generation of realistic, high-quality renderings from novel viewpoints of a scene. 
%NeRFs are parameterized by a simple neural network lke MLP that map 3D locations and viewing directions to a color and density at a certain position in 3D space. From this implicit representation, one obtains images at different viewpoints via differentiable volumetric rendering. 
%Recently, 3DGS \cite{3dgs} showed that it is possible to achieve real-time rendering speeds and drastically shorter training times using an explicit point-based representation modeled by 3D Gaussians (ellipsoids) with attributes that can model both geometry and view-dependent appearances. 
%A fast tile-based rasterizer and differentiable splatting are the key components behind achieving real-time rendering while also maintaining visual quality comparable with state-of-the-art implicit methods \cite{barron2022mipnerf360}. 
%However, both implicit and explicit approaches require dense captures of a 3D scene where each region is imaged multiple times from different angles. This dense capture setting necessitates the availability of hundreds of images to obtain a high-fidelity reconstruction for even forward-facing scenes or scenes containing only a few objects. Under-observed and/or occluded regions are usually prone to floaters and foggy artifacts. \\

Standard 3DGS, much like NeRF, is crippled in a sparse observation setting. In the absence of sufficient observations and global geometric cues, 3DGS invariably overfits to training views. This leads to severe artifacts and background collapse already in nearby novel views due to the inherent depth ambiguities associated with inferring 3D structure from few-view 2D images. There exists a long line of work to improve performance of both NeRF and 3DGS in sparse novel view synthesis~\cite{dsnerf, roessle2022depthpriorsnerf, dngaussian, sparsegs, fsgs, depthreggs, sparsenerf, dietnerf, regnerf, diffusionerf}. 
% \CW{Writing about it in related work} Existing sparse-view methods for NeRFs and 3DGS propose regularizers based on depth \cite{dsnerf, dngaussian, sparsegs, fsgs, roessle2022depthpriorsnerf, sparsenerf}, appearance \cite{regnerf}, semantics \cite{dietnerf}, or the frequency domain of NeRF inputs \cite{Yang_2023_CVPR}, to prevent the optimization from converging to a local minima, as they are simultaneously many solutions that can satisfy the sparse constraints.
While these works can reduce artifacts in sparsely observed regions, they are not able to fill in missing details due to the use of simple regularizations or weak priors.

In the setting of large-scale $360^{\circ}$ scenes, the problem is even more ill-posed and under-constrained. Much stronger priors are needed, such as those from large pre-trained 2D diffusion models~\cite{glide, dalle2, imagen, ldm}, capturing knowledge about typical structures in the 3D world. Recent approaches~\cite{zero1to3, zeronvs, reconfusion} add view-conditioning to these image generators by fine-tuning them on a large mixture of real-world and synthetic multi-view datasets. %Recently, ReconFusion~\cite{reconfusion} became the first method to attempt sparse-view reconstruction of large-scale $360^{\circ}$ scenes by supplementing a pretrained Latent Diffusion Model (LDM) with additional conditioning pathways for pixel-aligned view conditioning features~\cite{pixelnerf} and CLIP~\cite{clip} embeddings, and fine-tuning it on a huge mixture of real-world and synthetic multi-view datasets.
Leveraging these strong priors for optimization of radiance fields yields realistic reconstructions in unobserved areas of challenging $360^{\circ}$ scenes.  
%Integrating this strong prior with a ZipNeRF~\cite{zipnerf} backbone enables diverse and realistic reconstruction in unobserved areas of challenging $360^{\circ}$ scenes. 
In this work, we propose to forego augmenting a 2D diffusion model with additional channels for pose or context to make it 3D-aware. Instead, we perform low-cost fine-tuning of pretrained models to adapt them to specific sub-tasks of few-view reconstruction. This weakens the assumption of large-scale 3D training data, which is expensive to obtain.
%This removes dependency on the availability of large-scale 3D data but makes the problem more challenging. \SP{Please check if this looks ok now.}

We present SparseSplat360 (\method{}), an efficient method that addresses the given task of sparse 3D reconstruction by iteratively adding synthesized views to the training set of the 3D representation. The generation of new training views is divided and conquered as the 2D sub-tasks of (1) \emph{in-painting missing areas} and (2) \emph{artifact elimination}.  The in-painting model for the first stage is pre-trained on large 2D datasets and efficiently fine-tuned on the sparse views of the given scene. The artifact elimination network for the second stage is an image-to-image diffusion model, fine-tuned to specialize on removing typical artifacts appearing in sparse 3DGS. Thus, both stages utilize 2D diffusion models that are fine-tuned on small amounts of data. In each iteration, the models are conditioned on rendered novel views of the already existing scene, thus using 2D images as input and output, avoiding the requirement of training on large datasets of 3D scenes. In contrast to previous works, our method leverages stronger priors than simple regularizers and does not rely on million-scale multi-view data or huge compute resources to train a 3D-aware diffusion model. %\SP{Please check if this looks ok now.}

In summary, our contributions are:
\begin{itemize}
    \item We present a novel systematic approach to perform sparse 3D reconstruction of $360^{\circ}$ scenes by autoregressively adding generated novel views to the training set.
    \item We introduce a two-step approach for generating novel training views by performing \emph{in-painting} and \emph{artifact removal} with 2D diffusion models, which avoids fine-tuning on large-scale 3D data.
    %\item We design a method for obtaining camera trajectories of artificial novel views based on B-spline and SLERP interpolation, which is crucial for autoregressive scene generation. \JEL{Not sure if we can sell this as contribution if it is just standard SLERP}
    \item We introduce a sparse 3DGS baseline that improves reconstruction from sparse observation by applying regularization techniques without the need for pre-trained models.
    \item We show that \method{} outperforms recent works in reconstructing large 3D scenes from as few as $9$ input views.
\end{itemize}

\section{Related Work}
\subsection{Sparse-View Radiance Fields}
Following the breakthroughs of NeRF~\cite{nerf} and 3D Gaussian Splatting~\cite{3dgs} for inverse rendering of radiance fields, there have been many approaches to weaken the requirement of dense scene captures to sparse input views only. These methods can be categorized into regularization techniques and generalizable reconstruction priors.

\paragraph{Regularization Techniques}
Fitting a 3D representation from sparse observations only is an ill-posed problem and very prone to local minima. In the case of radiance fields, this is typically visible as 'floaters' during rendering of novel views.
A classical technique for training with limited data is regularization.
Many existing methods leverage depth from Structure-from-Motion~\cite{dsnerf, roessle2022depthpriorsnerf}, monocular estimation~\cite{dngaussian, sparsegs, fsgs, depthreggs}, or \mbox{RGB-D} sensors~\cite{sparsenerf}.
DietNeRF~\cite{dietnerf} proposes a semantic consistency loss based on CLIP~\cite{clip} features. FreeNeRF~\cite{Yang_2023_CVPR} regularizes frequency range of NeRF inputs by increasing the frequencies of positional-encoding features in a coarse-to-fine manner.
Moving closer to generative priors, RegNeRF~\cite{regnerf} and DiffusioNeRF~\cite{diffusionerf} maximize likelihoods of rendered patches under a trained normalizing flow or diffusion model, respectively. 

\paragraph{Generalizable Reconstruction}
In the case of very few or even a single view only, regularization techniques are usually not strong enough to account for the ambiguity in reconstruction.
Therefore, another line of research focuses on training priors for novel view synthesis across many scenes.
pixelNeRF~\cite{pixelnerf} extracts pixel-aligned CNN features from input images at projected sample points during volume rendering as conditioning for a shared NeRF MLP.
Similarly, many approaches~\cite{grf, mvsnerf, wcr, visionnerf} define different NeRF conditionings on 2D or fused 3D features.
Following the trend of leveraging explicit data structures for accelerating NeRFs, further priors have been learned on triplanes~\cite{neo360}, voxel grids~\cite{fe-nvs}, and neural points~\cite{simnp}.
Building upon the success of 3D Gaussian Splatting~\cite{3dgs} and its broad applications to, e.g., surface~\cite{sugar, 2dgs} or non-rigid reconstruction~\cite{dynamicgaussians, 4dgs, npg}, recent methods like pixelSplat~\cite{pixelsplat} and MVSplat~\cite{mvsplat} achieve state-of-the-art performance in stereo view interpolation while enabling real-time rendering.
However, all of these works train regression models that infer blurry novel views in case of high uncertainty.
Bridging the gap of generalizable and generative priors, GeNVS~\cite{genvs} and latentSplat~\cite{latentsplat} render view-conditioned feature fields followed by a 2D generative decoding to obtain a novel view.

\subsection{Generative Priors}
In case of ambiguous novel views, the expectation over all possible reconstructions might itself not be a reasonable prediction. Therefore, regression-based approaches fail. Generative methods, on the other hand, try to sample from this possibly multi-modal distribution.

\paragraph{Diffusion Models}
In recent years, diffusion models~\cite{ddpm, guideddiffusion} emerged as the state-of-the-art for image synthesis.
They are characterized by a pre-defined forward noising process that gradually destroys data by adding random (typically Gaussian) noise. The objective is to learn a reverse denoising process with a neural network that, after training, can sample from the data distribution given pure noise.
Important improvements include refined sampling procedures~\cite{ddim, edm} and the more efficient application in a spatially compressed latent space compared to the high-resolution pixel space~\cite{ldm}. Their stable optimization, in contrast to GANs, enabled today's text-to-image generators~\cite{glide, dalle2, imagen, ldm} trained on billions of images~\cite{laion5b}.

\paragraph{2D Diffusion for 3D}
While diffusion models have been applied directly on 3D representations like triplanes~\cite{triplanediffusion, ssdnerf}, voxel grids~\cite{diffrf}, or (neural) point clouds~\cite{pvdiffusion, pc2, npcd}, 3D data is scarce.
Given the success of large-scale diffusion models for image synthesis, there is a great research interest in leveraging them as priors for 3D reconstruction and generation.
DreamFusion~\cite{dreamfusion} and follow-ups~\cite{scorejacobianchaining, magic3d, fantasia3d, nerdi, dreamgaussian} employ score distillation sampling (SDS) to iteratively maximize the likelihood of radiance field renderings under a conditional 2D diffusion prior.
For sparse-view reconstruction, existing approaches incorporate view-conditioning via epipolar feature transform~\cite{sparsefusion}, cross-attention to encoded relative poses~\cite{zero1to3, zeronvs}, or pixelNeRF~\cite{pixelnerf} feature renderings~\cite{reconfusion}. However, this fine-tuning is expensive and requires large-scale multi-view data, which we circumvent with \method.

\vspace{-1.5mm}
\section{Method}
\label{sec:method}
In this section, we describe our method in detail. The section begins with a general overview of \method{} in Sec.~\ref{subsec:overview}, outlining the autoregressive algorithm for training view generation. In the following sections, the individual parts of the system are introduced: the in-painting module in Sec.~\ref{subsec:in-painting}, the artifact removal procedure in Sec.~\ref{subsec:img2img_diffusion}, and the sparse 3DGS baseline method in Sec.~\ref{subsec:geometry_baseline}.

\vspace{-1.0mm}
\subsection{\method{} for Sparse-View 3D Reconstruction}
\label{subsec:overview}
Given a sparse set of input images $\mathcal{I} = \{\mathbf{I}_1, \mathbf{I}_2, ..., \mathbf{I}_M\}$ with camera poses $\{\pi_1, \pi_2, ..., \pi_M\}$, and a sparse point cloud $\mathbf{P} \in \mathbb{R}^{S \times 3}$, estimated by Structure-from-Motion (SfM)~\cite{schoenberger2016sfm, schoenberger2016mvs}, our goal is to obtain a 3D Gaussian representation of the scene, which enables the rendering of novel views from camera perspectives that are largely different from given views in $\mathcal{I}$. The tackled scenario of sparse-view inputs is extremely challenging, as the given optimization problem is heavily under-constrained and leads to severe artifacts if done naively.

\begin{figure*}[!t]
    \centering
    \includegraphics[width=\textwidth]{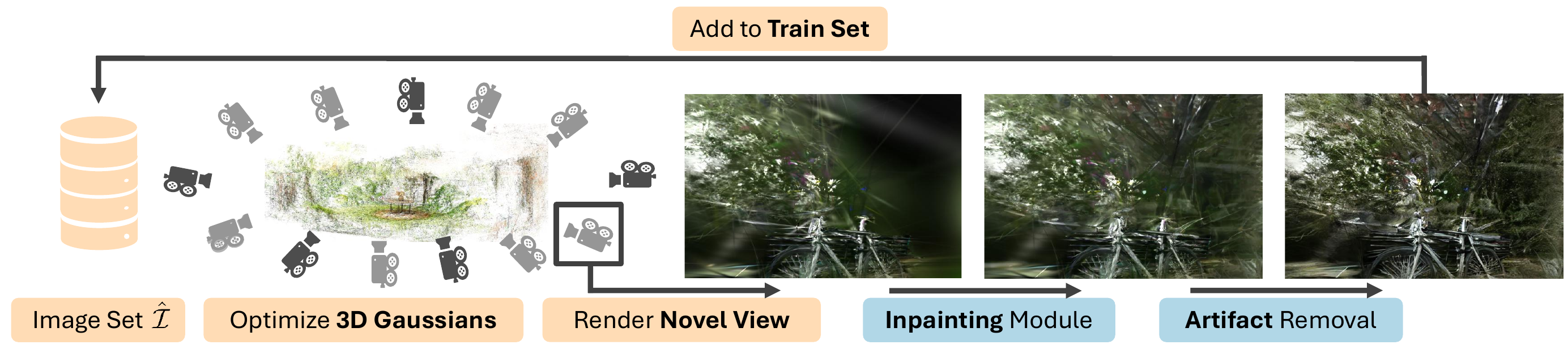}
    \caption{\textbf{Overview of \method{}}. We render 3D Gaussians fitted to our sparse set of $M$ views from a novel viewpoint. The image has missing regions and Gaussian artifacts, which are fixed by a combination of in-painting and denoising diffusion models. This then acts as pseudo ground truth to spawn and update 3D Gaussians and satisfy the new view constraints. This process is repeated for several novel views spanning the $360^{\circ}$ scene until the representation becomes multi-view consistent. }
    \label{fig:pipeline}
    \vspace{-3.0mm}
\end{figure*}

\begin{wrapfigure}{R}{8cm}
\vspace{-0.3cm}
\begin{minipage}{0.55\textwidth}
\begin{algorithm}[H]
\caption{\method{} Algorithm}
    \begin{algorithmic}[1]
    \Require Sparse input image set $\mathcal{I}$, camera poses $\{\pi_1, \pi_2, ..., \pi_M\}$, sparse point cloud $\mathbf{P} \in \mathbb{R}^{S \times 3}$
    \Ensure Set of 3D Gaussians $\mathcal{G}$
    \State $\hat{\mathcal{I}} \leftarrow \mathcal{I}$
    \State $\mathcal{G} \leftarrow$ Optimize \emph{Sparse 3DGS} for $k$ iterations
    \For{$N$ iterations}
        \State $\pi \leftarrow$ Sample novel camera pose.
        \State $\mathbf{I} \leftarrow R_\pi (\mathcal{G})$  - Render from camera $\pi$
        \State $\mathbf{I} \leftarrow$ In-paint$(\mathbf{I})$
        \State $\mathbf{I} \leftarrow$ ArtifactRemoval$(\mathbf{I})$
        \State $\hat{\mathbf{I}} \leftarrow \ \hat{\mathbf{I}} \cup \{\mathbf{I}\}$
        \State $\mathcal{G} \leftarrow$ Optimize \emph{Sparse 3DGS} for $k$ iterations
    \EndFor
    \end{algorithmic}
    \label{alg:main} 
\end{algorithm}
\end{minipage}
\vspace{-1cm}
\end{wrapfigure} 

An overview of \method{} is given in Fig.~\ref{fig:pipeline} and Alg.~\ref{alg:main}. The approach begins by optimizing a set of 3D Gaussians~\cite{3dgs} to reconstruct the initial sparse set of input images $\hat{\mathcal{I}} = \mathcal{I}$. For this, we introduce a \emph{Sparse 3DGS} baseline (c.f. Sec.~\ref{subsec:geometry_baseline}), which combines best practices from previous works on NeRFs.
The obtained representation serves as initial prior for $360^{\circ}$ reconstruction. Next, we autoregressively add new generated views to our training set: (1) we sample novel cameras and render novel views with artifacts and missing areas, make them look plausible by (2) performing in-painting (c.f.~\ref{subsec:in-painting}) and then (3) artifact removal (c.f.~\ref{subsec:img2img_diffusion}), before (4) adding them to set $\hat{\mathcal{I}}$ and continuing optimizing our 3D representation for $k$ iterations.

For training 3DGS in an iterative fashion, as outlined above, special precaution has to be taken to prevent overfitting on the initial views. An optimal schedule involves finding the number of iterations per cycle $k$ and the hyperparameters for the original 3DGS~\cite{3dgs}. We provide a detailed evaluation in App.~\ref{subsec:iterative_update}. In the following, we will detail the in-painting and artifact removal stages of our pipeline.

\vspace{-2.0mm}
\subsection{In-Painting Novel Views}
\label{subsec:in-painting}
When using only a sparse observation set, many areas remain unobserved, leading to areas of no Gaussians in the 3D representation and zero opacity regions in some novel views (c.f. Fig.\ref{fig:pipeline}). Regularization techniques cannot help with inferring details. Instead, we incorporate a generative in-painting diffusion model to fill such regions indicated by a binary mask $\phi$, which is obtained by rendering opacity from our current Gaussian representation $\mathcal{G}$ and binarizing it with a threshold $\tau$.

We fine-tune Stable Diffusion $2$~\cite{ldm} to perform in-painting on our novel view renderings. The current training images $\hat{\mathcal{I}}$ are used as training data to adapt it to the current scene. For the fine-tuning technique, we get inspired by recent work~\cite{tang2023realfill} that uses LoRA~\cite{lora} adapters for the UNet $\boldsymbol{\epsilon_\theta}$ and text encoder $c_\theta(y)$.
%Low-rank learnable residual layers $\Delta \mathbf{W} = \mathbf{AB}$ are injected into each weight matrix $\mathbf{W}$, where $\mathbf{A} \in \mathbb{R}^{n \times r}$ and $\mathbf{B} \in \mathbb{R}^{r \times n}$ ($r << n$) are 2 low-rank matrices. 
Given images $\hat{\mathcal{I}} = \{\mathbf{I}_1, ..., \mathbf{I}_M\}$, we create artificial masks $\{\phi_1, ..., \phi_M\}$ by creating random rectangular masks over each image and taking either their union or the complement of the union. Then, the adapter weights are fine-tuned using the following objective:

\begin{equation}
\label{eq:realfill_loss}
\begin{aligned}
    \mathcal{L} = \mathbb{E}_{i\sim\mathcal{U}(M), y, \epsilon \sim \mathcal{N}(\mathbf{0}, \mathbf{1}), t}\Big[\Vert \epsilon_t - \boldsymbol{\epsilon_\theta}(\mathbf{z}_t; t, \phi_i, c_\theta(y)) \Vert_{2}^{2}\Big] \textnormal{,}
\end{aligned}
\end{equation}

where $\mathbf{z}_t$ is the diffused latent encoding of image $\mathbf{I}_i$ at step $t$.
Here, $y$ is a simple text prompt - ``A photo of [V]'', where [V] is a rare token like in DreamBooth \cite{dreambooth} whose embedding is optimized for in-painting. For views $\mathbf{I} \notin \mathcal{I}_i$ (not from the original set), Eq \ref{eq:realfill_loss} is only evaluated for regions that have a rendered opacity $> \tau$. This in-painting objective for fine-tuning enables our model to in-paint missing regions in a novel view rendering with details faithful to the observed $M$ views. At inference, $\boldsymbol{\epsilon_\theta}$ predicts the noise in $\mathbf{z}_t$ as:

\begin{equation}
\label{eq:noise_pred}
\begin{aligned}
    \hat{\epsilon_t} = \boldsymbol{\epsilon_\theta}(\mathbf{z}_t; t, \phi_i, c_\theta(y)),
\end{aligned}
\end{equation}

which is used to progressively obtain less noisy latents in $s$ DDIM~\cite{ddim} sampling steps starting from $t$. After passing the denoised latent $\hat{z}_{0}$ through a VAE decoder $\mathcal{D}$. we obtain the in-painted image $x_{0}$.

\subsection{Removing Sparse-View Artifacts}
\label{subsec:img2img_diffusion}

The in-painting technique can fill in plausible details in low-opacity areas of novel view renders. However, it cannot deal with the dominance of blur, floaters, and color artifacts, which are not detected by the in-painting mask $\phi$. As such, we resort to a diffusion-based approach to learn how typical artifacts in 3D Gaussian representations from sparse views look like and how to remove them. 

We fine-tune an image-conditioned diffusion model~\cite{instructpix2pix} to edit images based on short, user-friendly edit instructions. Here, the UNet $\boldsymbol{\epsilon_\theta}$ is trained to predict noise in $z_t$ conditioned on a given image $c$, in addition to the usual text conditioning $c_{\theta}(y)$. To enable this, additional input channels are added to the first convolutional layer of $\boldsymbol{\epsilon_\theta}$ so that $z_t$ and $\mathcal{E}(c)$ can be concatenated. Weights of these channels are initialized to zero, whereas rest of the model is initialized from the pre-trained Stable Diffusion v1.5 checkpoint. For architectural design choices, we resort to those made by Instruct-Pix2Pix~\cite{instructpix2pix}.

\begin{figure*}[!t]
    \centering
    \includegraphics[width=\textwidth]{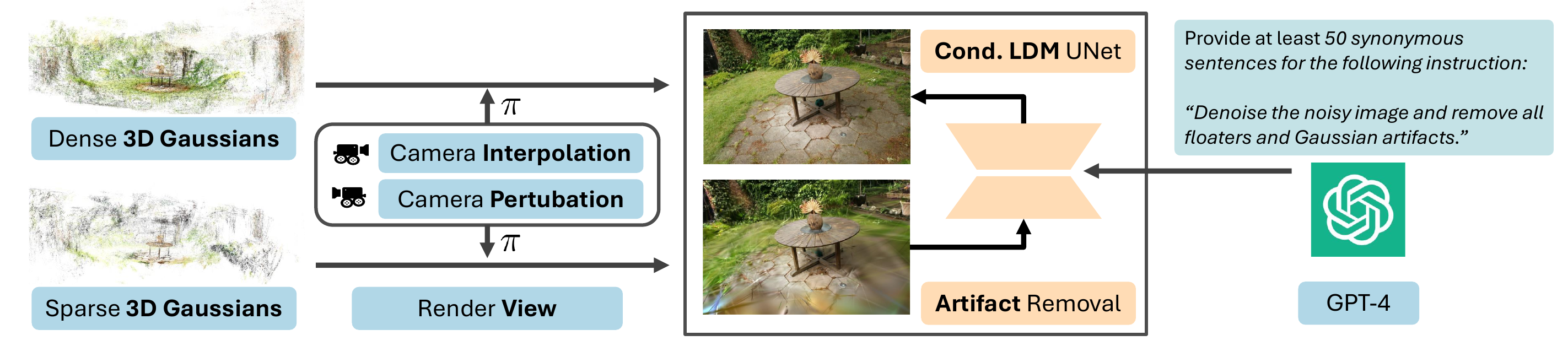}
    \caption{\textbf{Artifact removal fine-tuning.} Pairs of clean images and images with artifacts are obtained from 3DGS fitted to sparse and dense observations, respectively, across $36$ scenes. These are combined with one of 51 synonymous prompts generated by GPT-4~\cite{achiam2023gpt} from a base instruction. SD v1.5~\cite{ldm} is then fine-tuned with a dataset of $10.5K$ samples for the Gaussian artifact removal task.}
    \label{fig:dataset_creation}
    \vspace{-3.0mm}
\end{figure*}

\paragraph{Dataset Creation} \looseness=-1 For training, we rely on a set $\mathcal{X} = \{(\mathbf{x}^i, \mathbf{c}^i, y^i)_{i=1}^N\}$ of data triplets, each containing a clean image $\mathbf{x}^i$, an image with artifacts $\mathbf{c}^i$, and the corresponding edit instruction for fine-tuning $y^i$, to ``teach'' a diffusion model how to detect Gaussian artifacts and generate a clean version of the conditioning image. For this, we build an \emph{artifact simulation engine} comprising a 3DGS model fitted to dense views, \emph{Sparse 3DGS} fitted to few views, and camera interpolation and perturbation modules to use supervision of the dense model at viewpoints beyond ground truth camera poses. The fine-tuning setup is illustrated in Fig.~\ref{fig:dataset_creation}. For a given scene, we fit sparse models for \(M \in \{3, 6, 9, 18\}\) number of views. For larger $M$, we observe that there are very few artifacts beyond standard Gaussian blur. To have diversity in edit instructions, we start with a base instruction - ``Denoise the noisy image and remove all floaters and Gaussian artifacts.'' and ask GPT-4 to generate 50 synonymous instructions. During training, each clean, artifact image pair is randomly combined with one of these 51 instructions.

\paragraph{Training the Artifact Removal Module} Using our synthetically curated dataset $\mathcal{X}$, we fine-tune SD v1.5 as follows:

\begin{equation}
\label{eq:ip2p_fine-tune}
\begin{aligned}
    \mathcal{L} = \mathbb{E}_{i\sim\mathcal{U}(N), \epsilon \sim \mathcal{N}(\mathbf{0}, \mathbf{I}), t}\Big[\Vert \epsilon_t - \boldsymbol{\epsilon_\theta}(\mathbf{z}^i_t; t, \mathcal{E}(\mathbf{c}^i), c_\theta(y^i)) \Vert_{2}^{2}\Big]
\end{aligned}
\end{equation}

where $\mathbf{z}_t^i$ is the encoded image $\mathbf{x}^i$ diffused with sampled noise $\epsilon$ at time step $t$. Thus, the model is fine-tuned to generate clean images $\mathbf{x}^i$, conditioned on artifact images $\mathbf{c}^i$ and text prompts $y^i$.

\paragraph{Generating Clean Renders} Given an in-painted rendering $\mathbf{x}_0$ and the base prompt $y=$"Denoise the noisy image and remove all floaters and Gaussian artifacts" at inference time, the fine-tuned artifact removal UNet $\boldsymbol{\epsilon_\theta}$ predicts the noise in latent $\mathbf{z}_t$ according to $t\sim\mathcal{U}[t_{min},t_{max}]$ as:

\begin{equation}
\label{eq:noise_pred_ip2p}
\begin{aligned}
\hat{\epsilon}_t = & \, \boldsymbol{\epsilon_\theta}(\mathbf{z}_t; t, \varnothing, \varnothing) \\
                   & + s_I \cdot (\boldsymbol{\epsilon_\theta}(\mathbf{z}_t; t, \mathcal{E}(\mathbf{x}_0), \varnothing) - \boldsymbol{\epsilon_\theta}(\mathbf{z}_t; t, \varnothing, \varnothing)) \\
                   & + s_T \cdot (\boldsymbol{\epsilon_\theta}(\mathbf{z}_t; t, \varnothing, c_\theta(y)) - \boldsymbol{\epsilon_\theta}(\mathbf{z}_t; t, \mathcal{E}(\mathbf{x}_0), \varnothing))
\end{aligned}
\end{equation}

where $s_I$ and $s_T$ are the image and prompt guidance scales, dictating how strongly the final multistep reconstruction agrees with the in-painted render $\mathbf{x}_0$ and the edit prompt $y$, respectively. After $s$ DDIM~\cite{ddim} sampling steps, we obtain our final image by decoding the denoised latent.

\subsection{Sparse 3DGS}
\label{subsec:geometry_baseline}

The sparse 3DGS baseline serves as our starting point that already improves reconstruction quality over standard 3DGS. It is inspired by several recent works on sparse neural fields~\cite{dsnerf, regnerf} and 3DGS~\cite{sparsegs, fsgs, dngaussian}. The baseline unifies depth regularization from monocular depth estimators and depth priors from pseudo views, while using specific hyperparameter settings, such as densification thresholds and opacity reset configurations for Gaussian splatting. It is outlined in detail in App.~\ref{subsec:geometry_baseline_app} and evaluated individually in the ablation studies in Sec.~\ref{subsec:ablation_studies}.

\vspace{-1.5mm}
\section{Experiments}

This section compares \emph{\method{}} with state-of-the-art sparse-view reconstruction techniques. We also provide detailed ablation studies motivating specific design choices across different components of our approach. The setup for \emph{Iterative 3DGS} (c.f.~\ref{subsec:iterative_update}) translates seamlessly to the iterative distillation procedure (c.f.~\ref{subsec:distillation}) with diffusion priors, and hence, we refrain from further analysis here.

\vspace{-1.0mm}
\subsection{Experimental Setup}

\paragraph{Evaluation Dataset} We evaluate \emph{\method{}} on the 9 scenes of the MipNeRF360 dataset \cite{barron2022mipnerf360}, comprising 5 outdoor and 4 indoor scenes. Each scene has a central object or area of complex geometry with an equally intricate background. This makes it the most challenging \(360^{\circ}\) dataset compared to CO3D \cite{reizenstein2021common}, RealEstate 10K \cite{zhou2018stereo}, DTU \cite{jensen2014large}, etc. We retain the train/test split of the MipNeRF360 dataset, where every $8th$ image is kept aside for evaluation. To create $M$-view subsets, we sample from the train set of each scene using a geodesic distance-based heuristic to encourage maximum possible scene coverage (see supplement for details).

\vspace{-1.0mm}
\paragraph{Fine-tuning dataset} We fine-tune the in-painting module only on the $M$ input views. For the artifact removal module, we train 3DGS on sparse and dense subsets of $360^{\circ}$ scenes from MipNeRF360~\cite{barron2022mipnerf360}, Tanks and Temples~\cite{10.1145/3072959.3073599} and Deep Blending~\cite{10.1145/3272127.3275084} across a total of 36 scenes to obtain $\sim 10.5K$ data triplets. We train 9 separate artifact removal modules holding out the MipNeRF360 scene we want to reconstruct. On a single A100 GPU, fine-tuning the in-painting and artifact removal modules takes roughly $2$h and $1$h, respectively.

\vspace{-1.0mm}
\paragraph{Baselines} We compare our approach against $7$ baselines. FreeNeRF~\cite{Yang_2023_CVPR}, RegNeRF~\cite{regnerf}, DiffusioNeRF~\cite{diffusionerf}, and DNGaussian~\cite{dngaussian} are few-view regularization methods based on NeRFs or 3D Gaussians. ZeroNVS~\cite{zeronvs} is a recent generative approach for reconstructing a complete 3D scene from a single image. We use the ZeroNVS$^*$ baseline introduced in ReconFusion~\cite{reconfusion}, designed to adapt ZeroNVS to multi-view inputs. Conditioning the diffusion model on the input view closest to the sampled random view enables scene reconstruction for a general $M$-view setting. We also compare against 3DGS, the reconstruction pipeline for \method{}, and against Sparse 3DGS, our self-created baseline. We cannot compare with ReconFusion~\cite{reconfusion} as their code is not publicly available, and their method builds on top of closed-source diffusion models. We expect to perform slightly worse due to weaker priors, but our approach is more lightweight and data-efficient. 

\vspace{-1.0mm}
\paragraph{Metrics} Due to the generative nature of our approach, we employ FID~\cite{heusel2017gans} and KID~\cite{bińkowski2018demystifying} to measure similarity of distribution of reconstructed novel views and ground truth images. We also compute two perceptual metrics - LPIPS~\cite{lpips} and DISTS~\cite{dists} - to measure similarity in image structure and texture in the feature space. Despite their known drawbacks as evaluators of generative techniques~\cite{genvs, zeronvs}, we additionally provide PSNR and SSIM scores for completeness. Both favor pixel-aligned blurry estimates over high-frequency details, making them ill-suited to our setting.

\vspace{-1.0mm}
\subsection{Implementation Details}
\label{subsec:implementation}
We implement our entire framework in Pytorch 1.12.1 and run all experiments on single $A100$ or $A40$ GPUs. We work with image resolutions in the 400-600 pixel range as this is closest to the output resolution of $512$ for both diffusion models. We set $\lambda_1 = 0.2$ (same as 3DGS) and vary $\lambda_{depth}, \lambda_{pseudo}$ in $\{0.0, 0.05, 0.1\}$ for \emph{Sparse 3DGS}. For in-painting, we set $\tau = 0.8$ and fine-tune the in-painting module at \(512 \times 512\) resolution for 3000-5000 steps with LoRA modules of rank in $\{8, 16, 32\}$ (depending on $M$). We use a batch size of $16$ and learning rates of 2e-4 for the UNet and 4e-5 for the text encoder. The artifact removal module is trained at \(256 \times 256\) resolution for $2500$ iterations with batch size $16$ and learning rate 1e-4. To enable classifier-free guidance for both models, we randomly dropout conditioning inputs (text, mask, image, etc.) with probability of $0.1$ during training. The classifier-free guidance scales are set to $s_I = 2.5$ and $s_T = 7.0$. We use $t_{max} = 0.99$ for both in-painting and artifact removal and linearly decrease $t_{min}$ for the in-painting module from $0.98$ to $0.90$, and from $0.98$ to $0.70$ for the artifact removal module. We sample both the in-painted and clean renders for $s = 20$ DDIM sampling steps. We also linearly decay the weight of $\mathcal{L}_{sample}$ (c.f. \ref{subsec:distillation}) from $1$ to $0.1$ over $30k$ iterations. 

\vspace{-1.0mm}
\subsection{Comparison Results}
\begin{figure}[!htbp]
    % \captionsetup{justification=centering}
    \centering
        \centering
        \setlength{\lineskip}{0pt} % Reduce vertical space between rows
        \setlength{\lineskiplimit}{0pt} % Reduce vertical space between rows
        \begin{tabular}{@{}c@{}*{7}{>{\centering\arraybackslash}p{0.141\linewidth}@{}}}
            & 3DGS & RegNeRF & DNGaussian & DiffusioNeRF & ZeroNVS* & Ours & Ground Truth \\
            \raisebox{\height}{\rotatebox[origin=c]{90}{\scriptsize Bicycle (9)}} &
            \includegraphics[width=\linewidth]{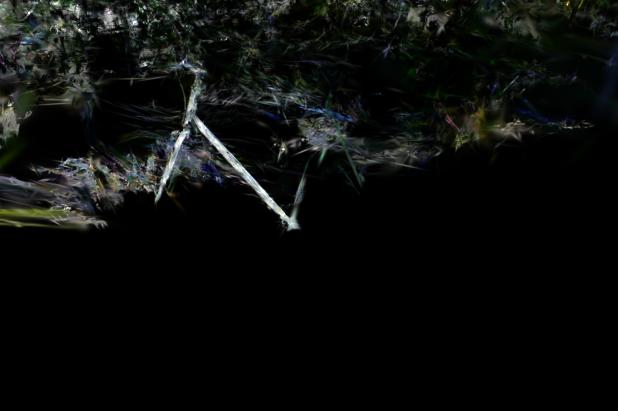} &
            \includegraphics[width=\linewidth]{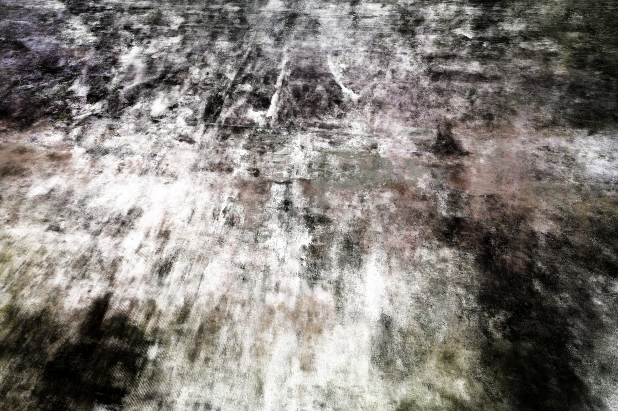} &
            \includegraphics[width=\linewidth]{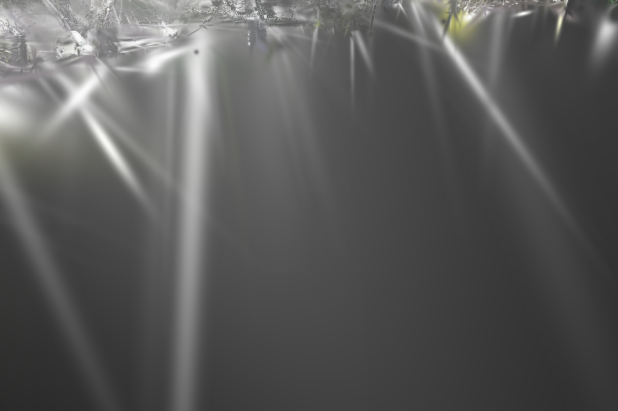} &
            \includegraphics[width=\linewidth]{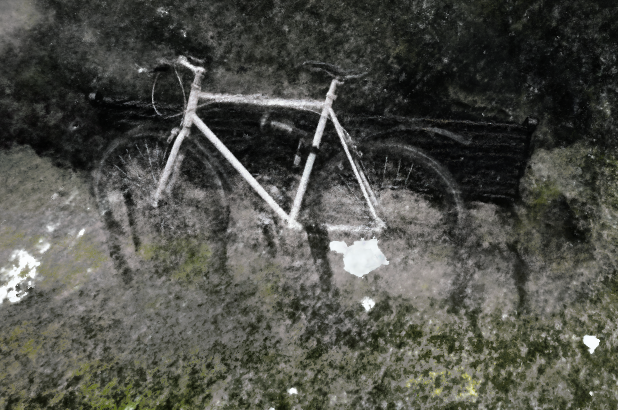} &
            \includegraphics[width=\linewidth]{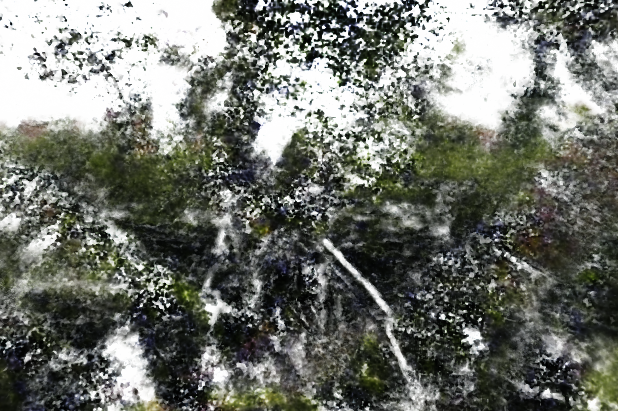} &
            \includegraphics[width=\linewidth]{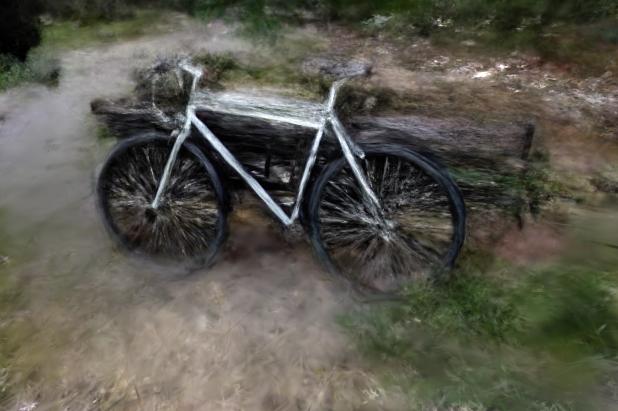} &
            \includegraphics[width=\linewidth]{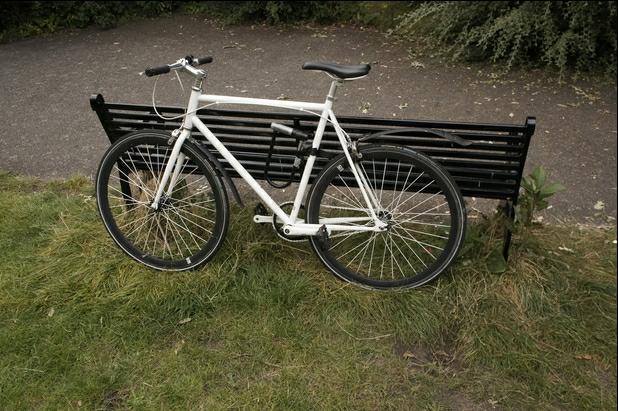} \\

            \raisebox{\height}{\rotatebox[origin=c]{90}{\scriptsize Treehill (9)}} &
            \includegraphics[width=\linewidth]{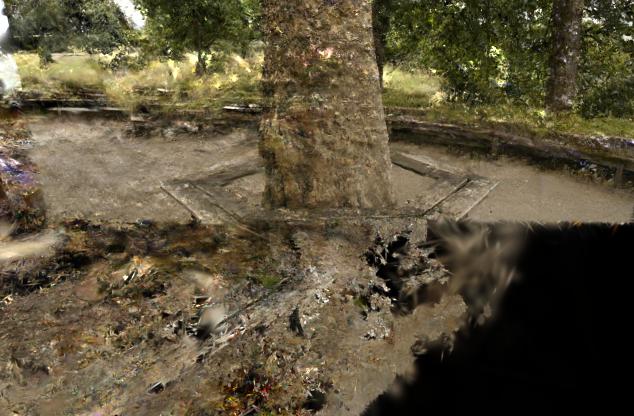} &
            \includegraphics[width=\linewidth]{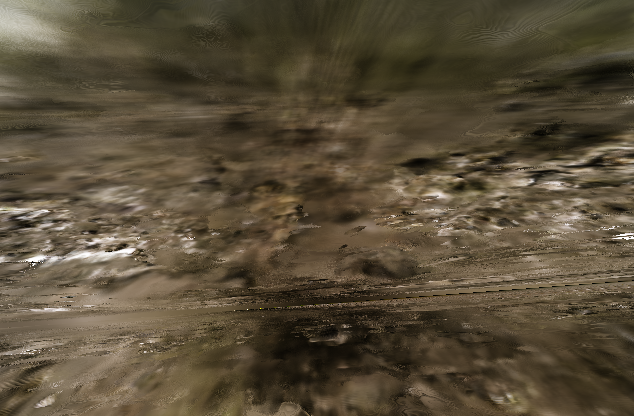}&
            \includegraphics[width=\linewidth]{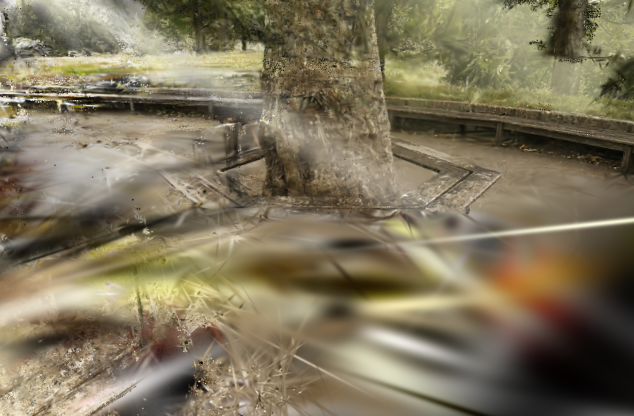} &
            \includegraphics[width=\linewidth]{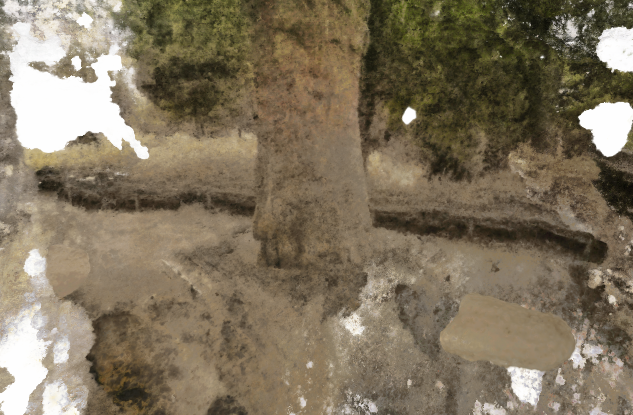} &
            \includegraphics[width=\linewidth]{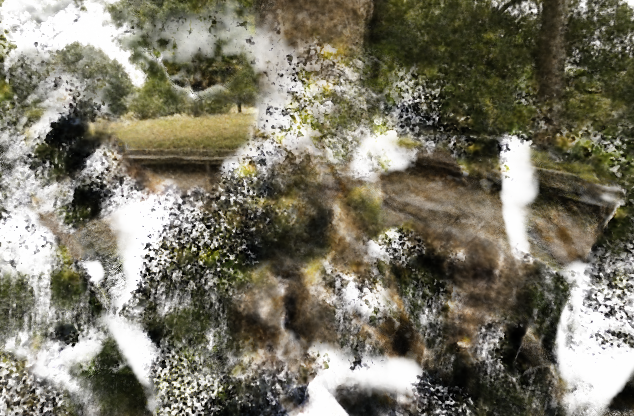} &
            \includegraphics[width=\linewidth]{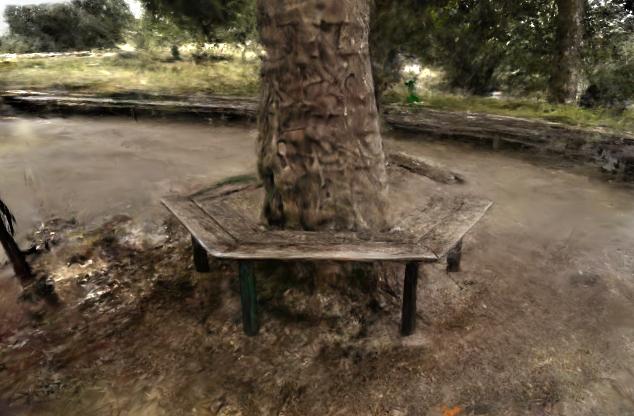} &
            \includegraphics[width=\linewidth]{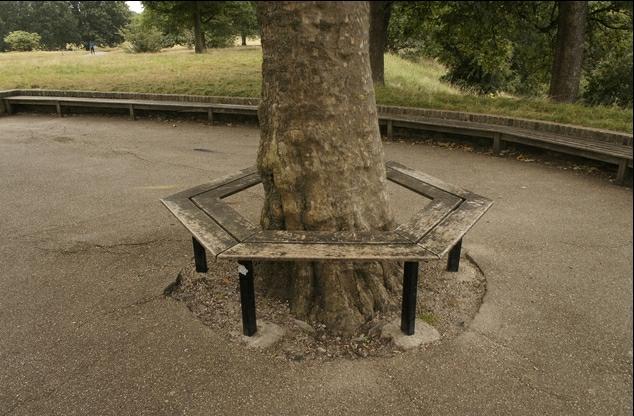} \\

            % \raisebox{\height}{\rotatebox[origin=c]{90}{\scriptsize Garden (9)}} &
            % \includegraphics[width=\linewidth]{images/3dgs/DSC07996.JPG} &
            % \includegraphics[width=\linewidth]{images/regnerf/color_005.png}&
            % \includegraphics[width=\linewidth]{images/dngaussian/00005.png} &
            % \includegraphics[width=\linewidth]{images/diffusionerf/ngp_ep0051_DSC07996.png} &
            % \includegraphics[width=\linewidth]{images/zeronvs/40.png} &
            % \includegraphics[width=\linewidth]{images/sp2360/DSC07996.JPG} &
            % \includegraphics[width=\linewidth]{images/gt/DSC07996.JPG} \\

            \raisebox{\height}{\rotatebox[origin=c]{90}{\scriptsize Garden (9)}} &
            \includegraphics[width=\linewidth]{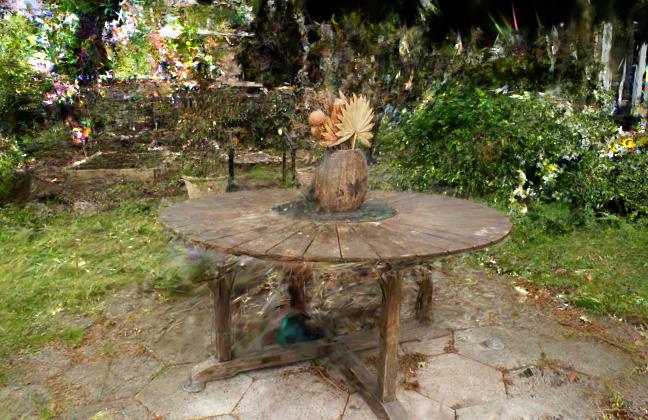} &
            \includegraphics[width=\linewidth]{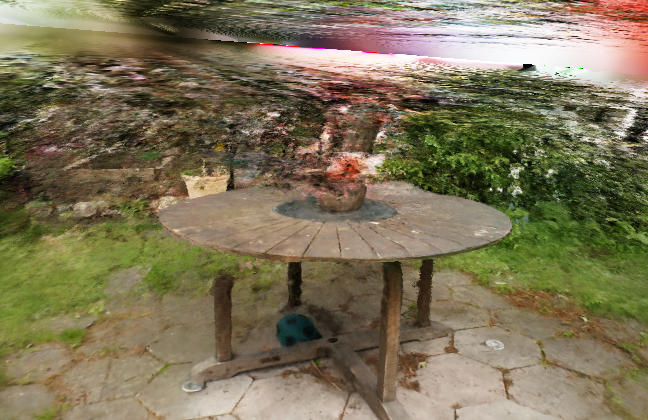}&
            \includegraphics[width=\linewidth]{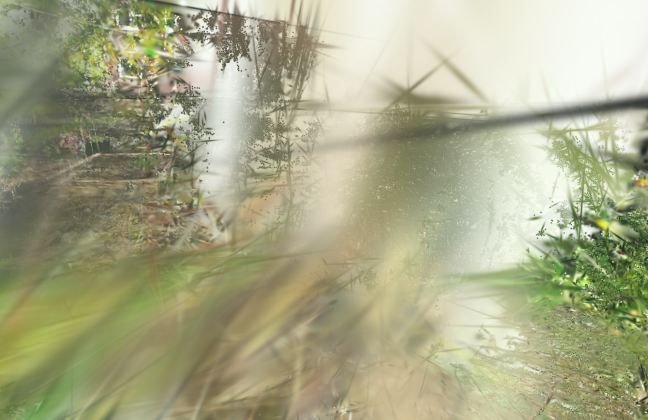} &
            \includegraphics[width=\linewidth]{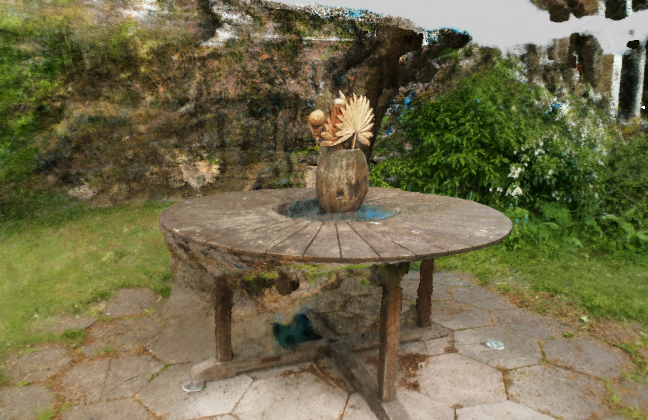} &
            \includegraphics[width=\linewidth]{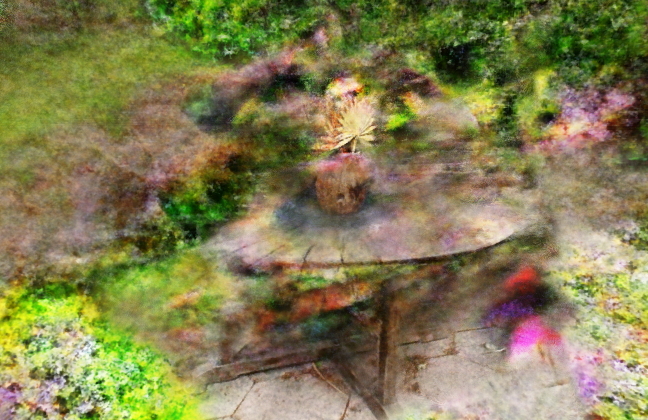} &
            \includegraphics[width=\linewidth]{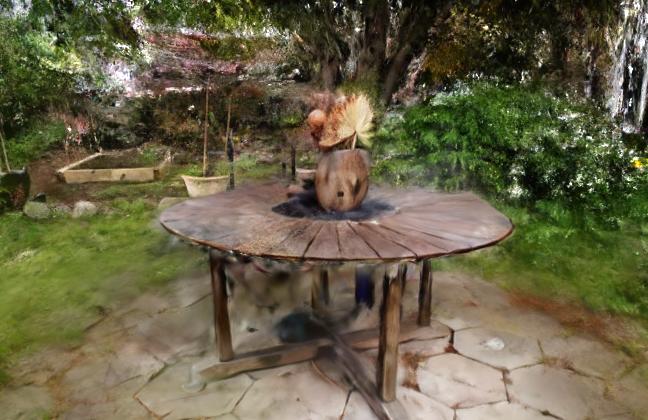} &
            \includegraphics[width=\linewidth]{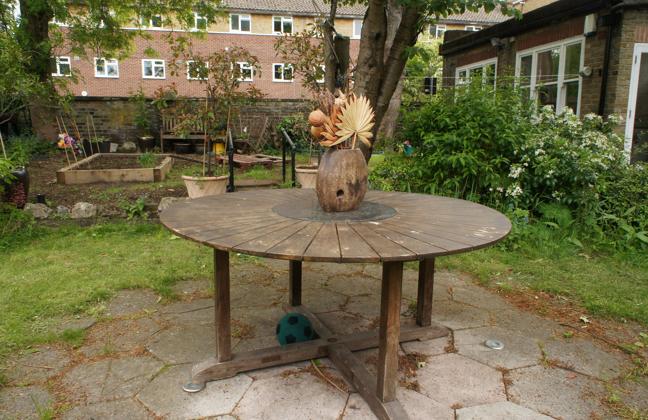} \\

            \raisebox{\height}{\rotatebox[origin=c]{90}{\scriptsize Stump (9)}} &
            \includegraphics[width=\linewidth]{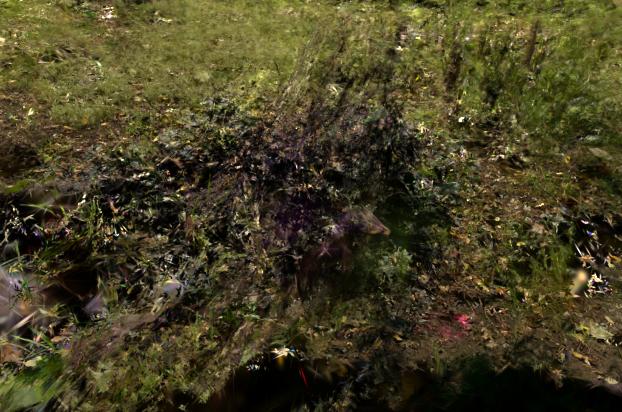} &
            \includegraphics[width=\linewidth]{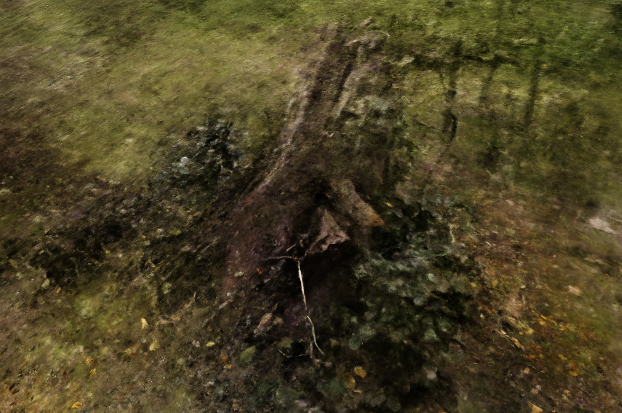}&
            \includegraphics[width=\linewidth]{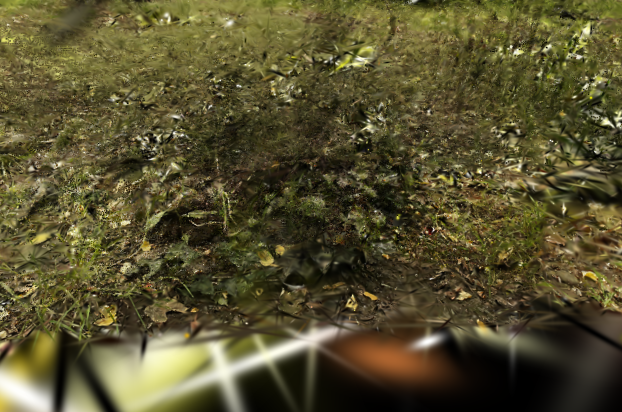} &
            \includegraphics[width=\linewidth]{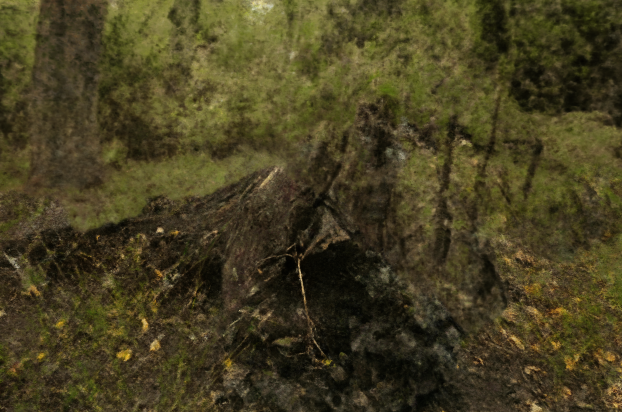} &
            \includegraphics[width=\linewidth]{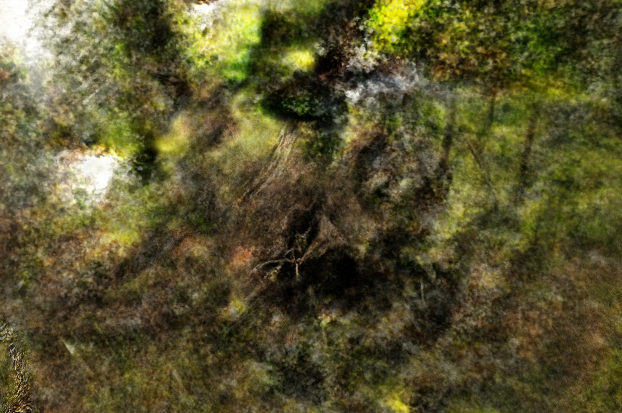} &
            \includegraphics[width=\linewidth]{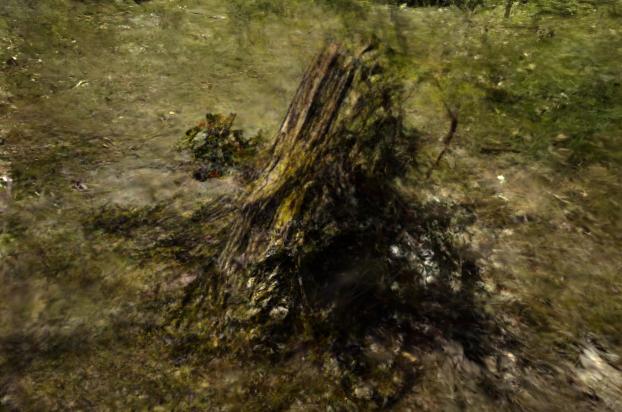} &
            \includegraphics[width=\linewidth]{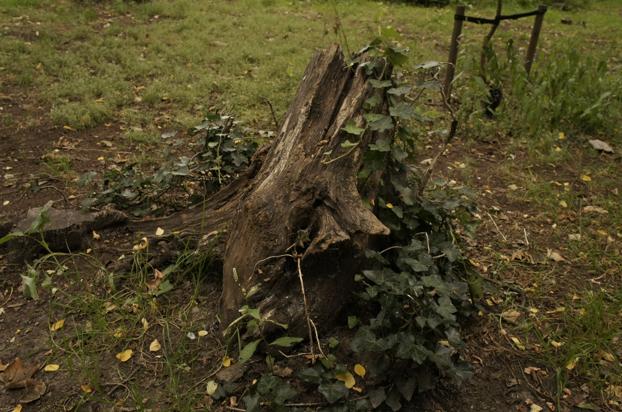} \\

            \raisebox{\height}{\rotatebox[origin=c]{90}{\scriptsize Kitchen (9)}} &
            \includegraphics[width=\linewidth]{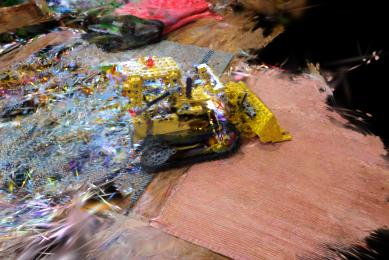} &
            \includegraphics[width=\linewidth]{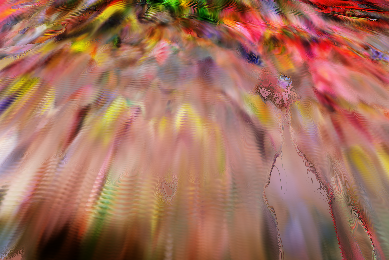}&
            \includegraphics[width=\linewidth]{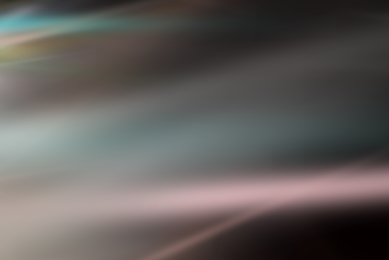} &
            \includegraphics[width=\linewidth]{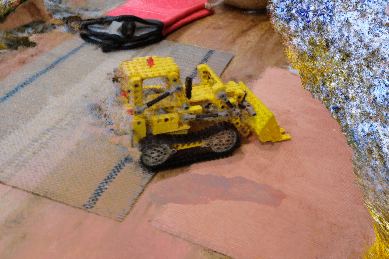} &
            \includegraphics[width=\linewidth]{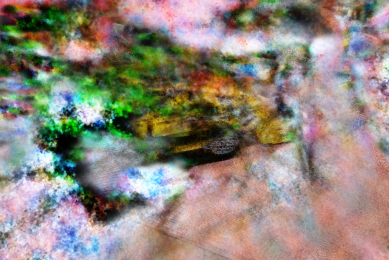} &
            \includegraphics[width=\linewidth]{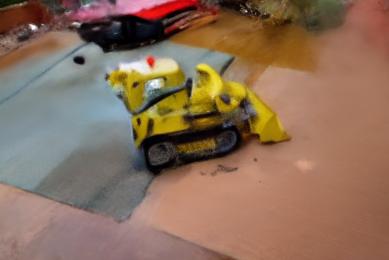} &
            \includegraphics[width=\linewidth]{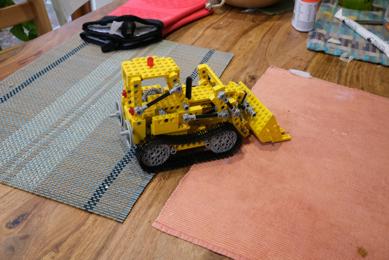} \\

            \raisebox{\height}{\rotatebox[origin=c]{90}{\scriptsize Flowers (9)}} &
            \includegraphics[width=\linewidth]{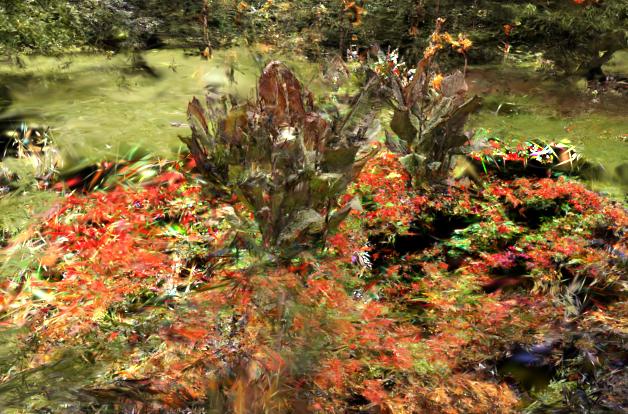} &
            \includegraphics[width=\linewidth]{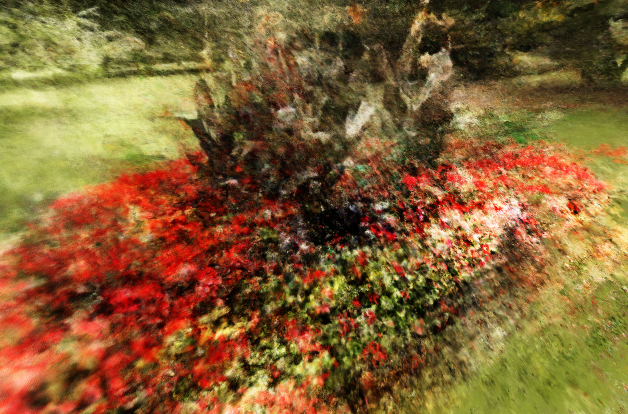}&
            \includegraphics[width=\linewidth]{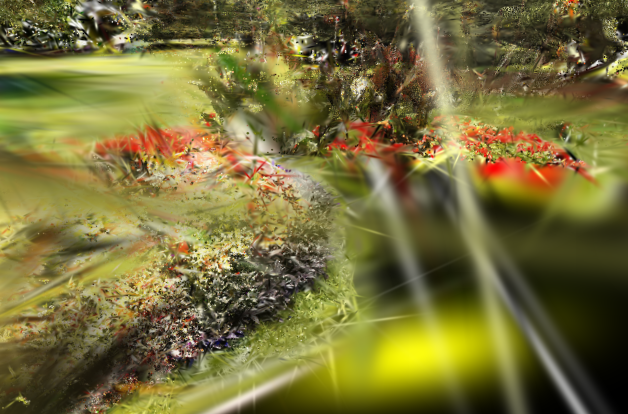} &
            \includegraphics[width=\linewidth]{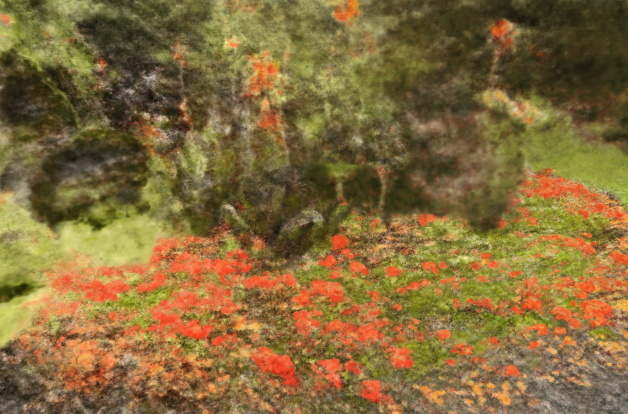} &
            \includegraphics[width=\linewidth]{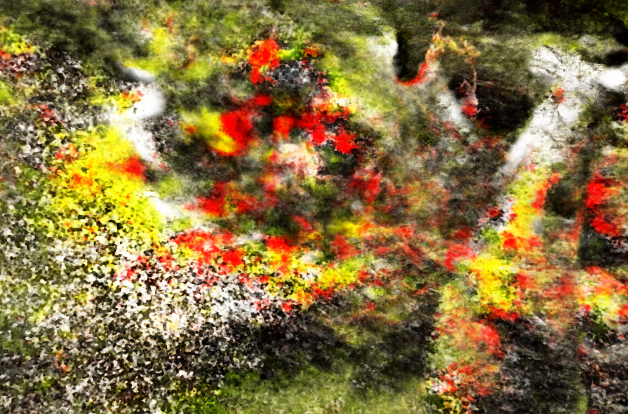} &
            \includegraphics[width=\linewidth]{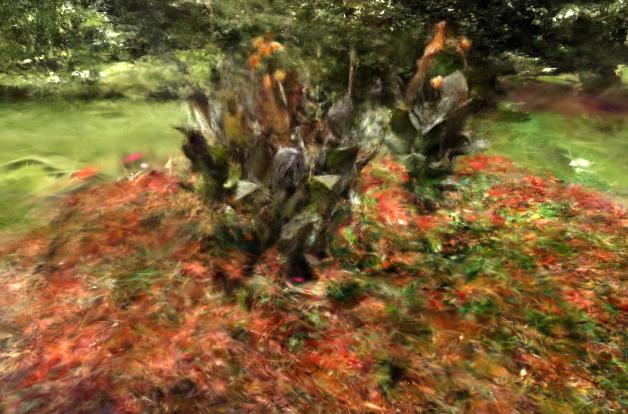} &
            \includegraphics[width=\linewidth]{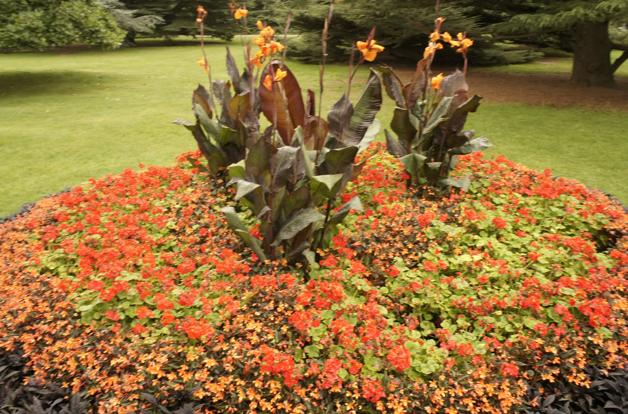} \\

            \raisebox{\height}{\rotatebox[origin=c]{90}{\scriptsize Bonsai (6)}} &
            \includegraphics[width=\linewidth]{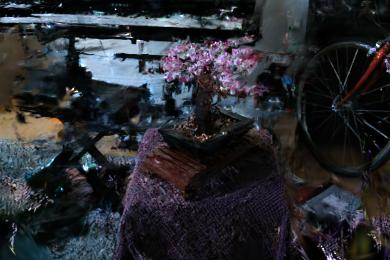} &
            \includegraphics[width=\linewidth]{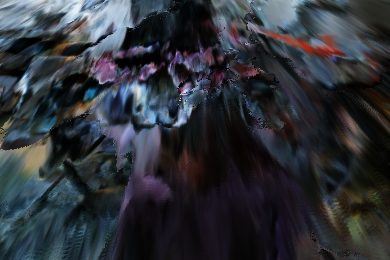}&
            \includegraphics[width=\linewidth]{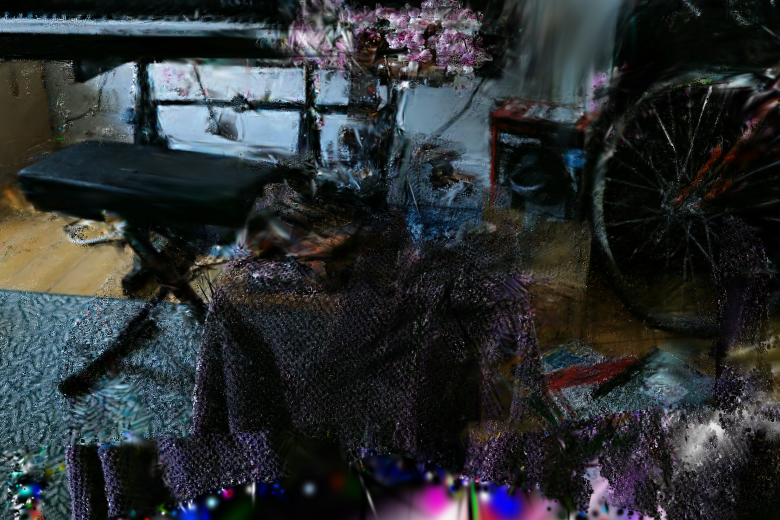} &
            \includegraphics[width=\linewidth]{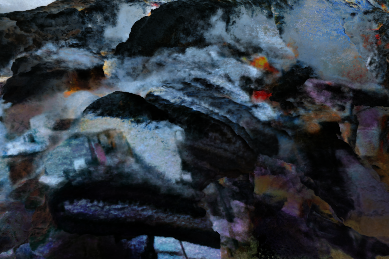} &
            \includegraphics[width=\linewidth]{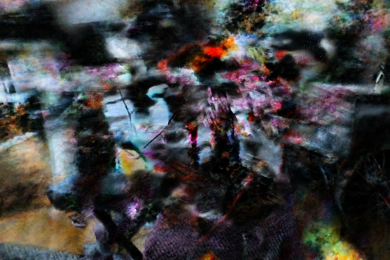} &
            \includegraphics[width=\linewidth]{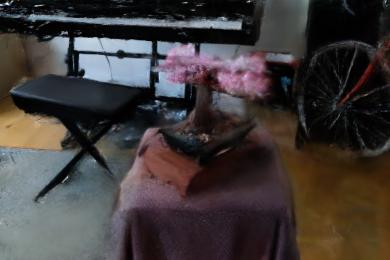} &
            \includegraphics[width=\linewidth]{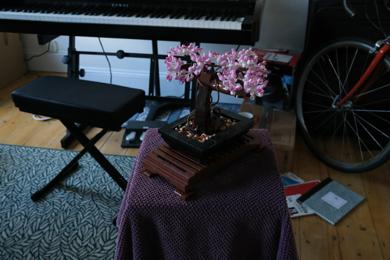}
        \end{tabular}
        
    \caption{\textbf{Qualitative comparison} of \emph{\method{}} with few-view methods. Our approach consistently fairs better in recovering image structure from foggy geometry, where baselines typically struggle with ``floaters'' and color artifacts. We encourage the reader to refer to our \textbf{supplemental 360° video}, where the benefits of our method can be observed along a smooth trajectory.}
    \label{fig:main_qual}
    \vspace{-2.5mm}
\end{figure}

We evaluate all baselines on our proposed splits for each scene. We report averaged quantitative results in Tables \ref{tab:main_table_classic} and \ref{tab:main_table_gen} and compare novel view rendering quality in Fig \ref{fig:main_qual}. We outperform all baselines for $9$-view reconstruction across all metrics and are second only to DiffusioNeRF on FID, KID, and DISTS for $3$ and $6$-view reconstruction. Unlike the baselines, our approach consistently improves with increasing $M$ across all metrics.

\begin{table}[!htbp]
\caption{\textbf{Quantitative comparison} with state-of-the-art sparse-view reconstruction techniques on classical metrics. Despite being a generative solution, we outperform all baselines across all view splits on both pixel-aligned and perceptual metrics.}
\label{tab:main_table_classic}
\centering
\resizebox{0.99\textwidth}{!}{%
\large
\begin{tabular}{lccccccccc}
\toprule
& \multicolumn{3}{c}{PSNR $\uparrow$} & \multicolumn{3}{c}{SSIM $\uparrow$} & \multicolumn{3}{c}{LPIPS $\downarrow$} \\
Method & 3-view  & 6-view & 9-view & 3-view  & 6-view & 9-view & 3-view  & 6-view  & 9-view \\ 
\midrule
% MipNeRF & \cellcolor{tabsecond}{10.261} & \cellcolor{tabthird}{10.375} & 10.656 & 0.118 & 0.128 & 0.138 & \cellcolor{tabthird}{0.684} & \cellcolor{tabthird}{0.684} & \cellcolor{tabsecond}{0.667} \\
3DGS & 10.288 & \cellcolor{tabthird}{11.628} & \cellcolor{tabsecond}{12.658} & 0.102 & \cellcolor{tabsecond}{0.141} & \cellcolor{tabsecond}{0.190} & 0.709 & \cellcolor{tabthird}{0.661} & \cellcolor{tabsecond}{0.605} \\

FreeNeRF & 9.948 & 9.599 & 10.641 &	\cellcolor{tabsecond}{0.124} &	0.129 &	0.145 &	\cellcolor{tabthird}{0.682} & 0.679 & 0.668 \\

RegNeRF & \cellcolor{tabsecond}{11.030} & 10.764 & 11.020 & \cellcolor{tabthird}{0.117} & 0.134 & 0.145 & \cellcolor{tabsecond}{0.663} & \cellcolor{tabsecond}{0.660} & 0.661 \\

DNGaussian & 9.867 & 10.671 & 11.275 & \cellcolor{tabsecond}{0.124} & \cellcolor{tabthird}{0.137} & \cellcolor{tabthird}{0.179} & 0.754 & 0.730 & 0.729 \\

DiffusioNeRF & \cellcolor{tabthird}{10.749} & \cellcolor{tabsecond}{11.728} & \cellcolor{tabthird}{11.430} & 0.093 & 0.116 & 0.112 & 0.709 & 0.678 & \cellcolor{tabthird}{0.654} \\

ZeroNVS* & 10.406 & 9.990 & 9.719 & 0.079 & 0.079 & 0.082 & 0.709 & 0.711 & 0.700 \\

\textbf{Ours} & \cellcolor{tabfirst}{12.927} & \cellcolor{tabfirst}{13.701} & \cellcolor{tabfirst}{14.121} & \cellcolor{tabfirst}{0.211} & \cellcolor{tabfirst}{0.231} & \cellcolor{tabfirst}{0.261} & \cellcolor{tabfirst}{0.647} & \cellcolor{tabfirst}{0.622} & \cellcolor{tabfirst}{0.591} \\
\bottomrule
\end{tabular}
}
\end{table}

\begin{table}[!htbp]
\caption{\textbf{Quantitative comparison} with few-view reconstruction techniques on metrics suited for generative reconstruction. We are second only to DiffusioNeRF for $3$ and $6$-view reconstruction but achieve higher scores for $9$ views.}
\label{tab:main_table_gen}
\centering
\resizebox{0.99\textwidth}{!}{%
\large
\begin{tabular}{lccccccccc}
\toprule
& \multicolumn{3}{c}{FID $\downarrow$} & \multicolumn{3}{c}{KID $\downarrow$} & \multicolumn{3}{c}{DISTS $\downarrow$} \\
Method & 3-view  & 6-view & 9-view & 3-view  & 6-view & 9-view & 3-view  & 6-view  & 9-view \\ 
\midrule
% MipNeRF & \cellcolor{tabsecond}{10.261} & \cellcolor{tabthird}{10.375} & 10.656 & 0.118 & 0.128 & 0.138 & \cellcolor{tabthird}{0.684} & \cellcolor{tabthird}{0.684} & \cellcolor{tabsecond}{0.667} \\
3DGS & 392.620 & 343.336 & \cellcolor{tabthird}{292.324} & 0.313 & 0.268 & \cellcolor{tabthird}{0.229} & 0.476 & 0.429 & \cellcolor{tabsecond}{0.321} \\

FreeNeRF & \cellcolor{tabthird}{347.625} & 343.833 & 342.917 & \cellcolor{tabsecond}{0.254} & \cellcolor{tabthird}{0.249} & 0.258 & \cellcolor{tabthird}{0.392} & \cellcolor{tabthird}{0.388} & 0.388 \\

RegNeRF & 362.856 & 347.045 & 349.043 & 0.291 & 0.266 & 0.247 & 0.399 & 0.403 & 0.404 \\

DNGaussian & 431.687 & 420.110 & 414.307 & 0.311 & 0.285 & 0.281 & 0.581 & 0.571 & 0.544 \\

DiffusioNeRF & \cellcolor{tabfirst}{273.096} & \cellcolor{tabfirst}{225.661} & \cellcolor{tabsecond}{290.184} & \cellcolor{tabfirst}{0.158} & \cellcolor{tabfirst}{0.104} & \cellcolor{tabsecond}{0.183} & \cellcolor{tabfirst}{0.362} & \cellcolor{tabfirst}{0.319} & \cellcolor{tabthird}{0.378} \\

ZeroNVS* & 351.090 & \cellcolor{tabthird}{335.155} & 337.457 & 0.283 & 0.282 & 0.290 & 0.437 & 0.429 & 0.428 \\

\textbf{Ours} & \cellcolor{tabsecond}{318.470} & \cellcolor{tabsecond}{283.504} & \cellcolor{tabfirst}{230.565} & \cellcolor{tabthird}{0.273} & \cellcolor{tabsecond}{0.229} & \cellcolor{tabfirst}{0.162} & \cellcolor{tabsecond}{0.384} & \cellcolor{tabsecond}{0.357} & \cellcolor{tabfirst}{0.315} \\
\bottomrule
\end{tabular}
}
\end{table}

\subsection{Ablation Studies}
\label{subsec:ablation_studies}

\begin{figure}[!htbp]
    % \captionsetup{justification=centering}
    \centering
        \centering
        \setlength{\lineskip}{0pt} % Reduce vertical space between rows
        \setlength{\lineskiplimit}{0pt} % Reduce vertical space between rows
        \begin{tabular}{@{}c@{}*{7}{>{\centering\small\arraybackslash}p{0.14\linewidth}@{}}}
            & Sparse 3DGS & w/o Artifact R. & w/o In-painting & w/o Schedule & 3DGS Loss & Ours & Ground Truth \\
            \raisebox{\height}{\rotatebox[origin=c]{90}{\scriptsize Render 1}}
            & \includegraphics[width=\linewidth]{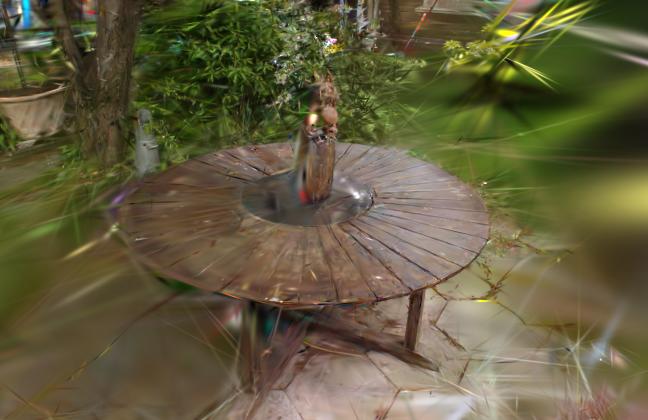} &
            \includegraphics[width=\linewidth]{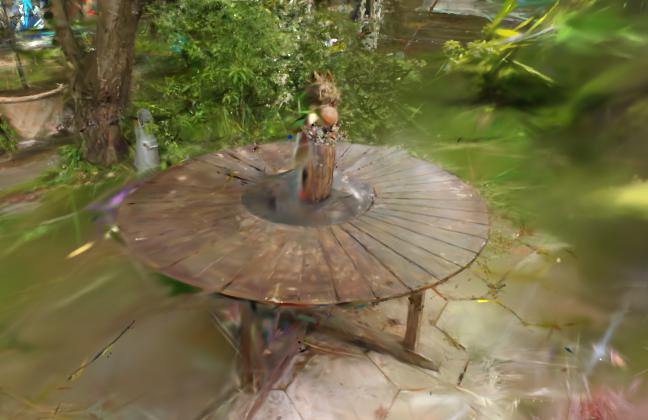} &
            \includegraphics[width=\linewidth]{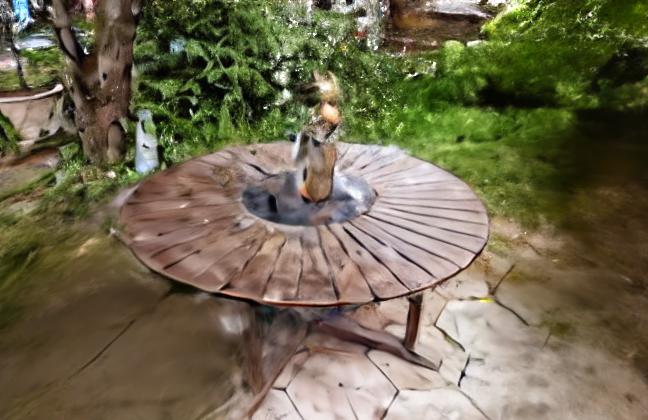} &
            \includegraphics[width=\linewidth]{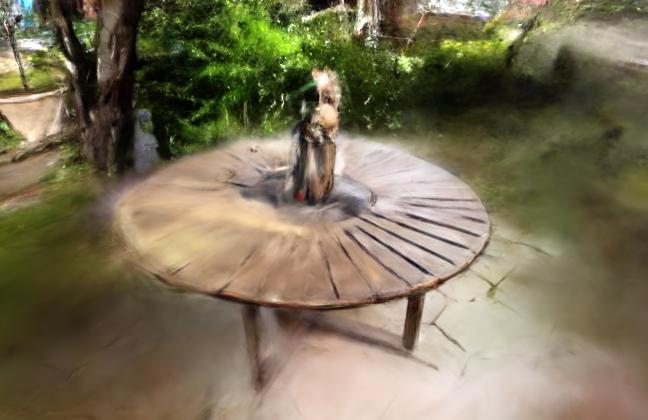} &
            \includegraphics[width=\linewidth]{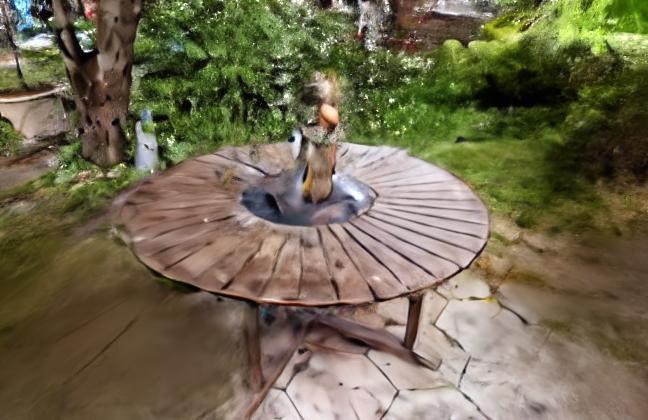} &
            \includegraphics[width=\linewidth]{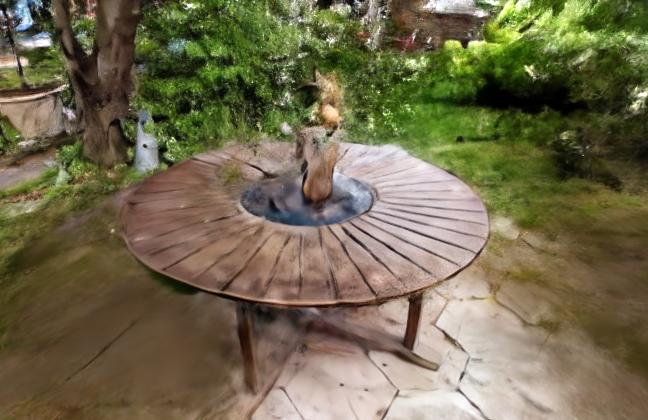} &
            \includegraphics[width=\linewidth]{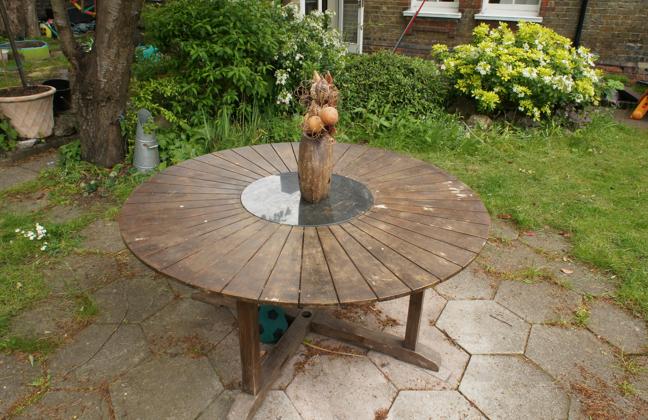} \\

            \raisebox{\height}{\rotatebox[origin=c]{90}{\scriptsize Render 2}}
            & \includegraphics[width=\linewidth]{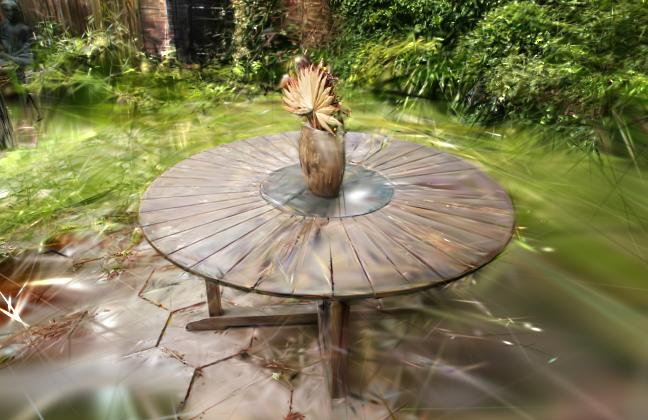} &
            \includegraphics[width=\linewidth]{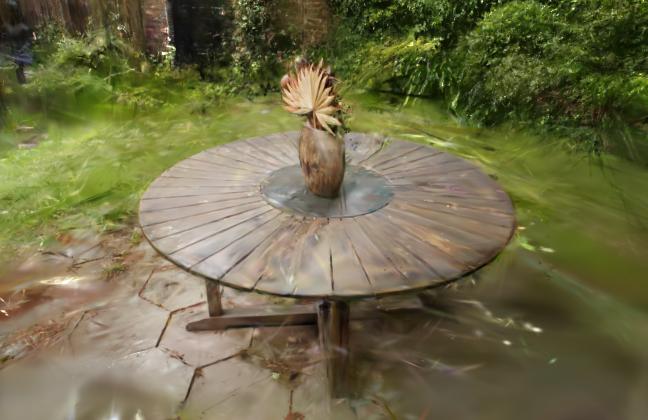} &
            \includegraphics[width=\linewidth]{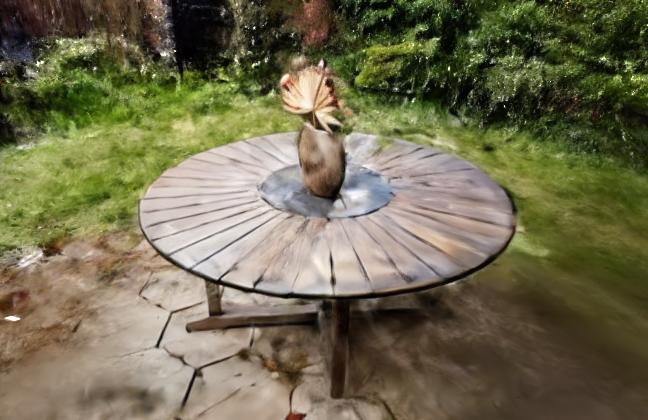} &
            \includegraphics[width=\linewidth]{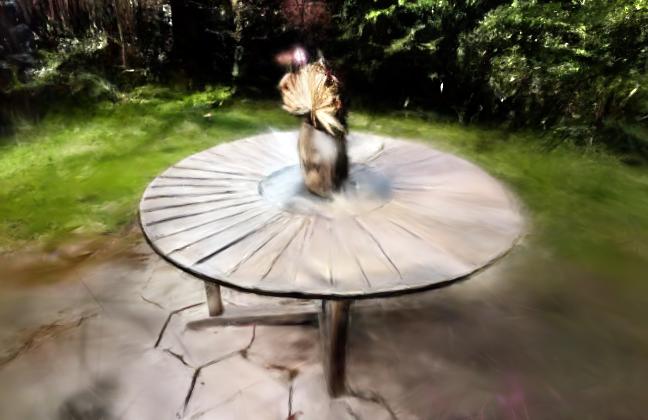} &
            \includegraphics[width=\linewidth]{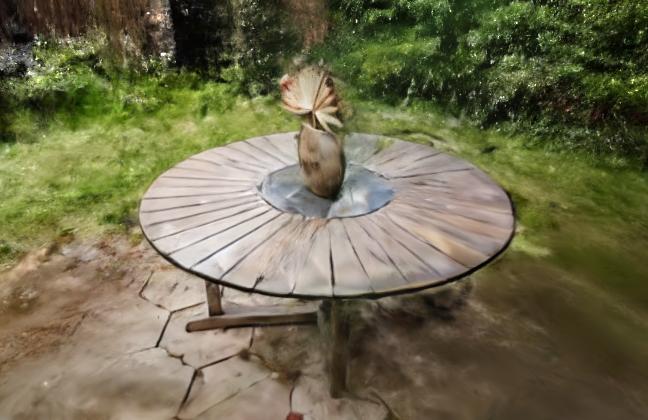} &
            \includegraphics[width=\linewidth]{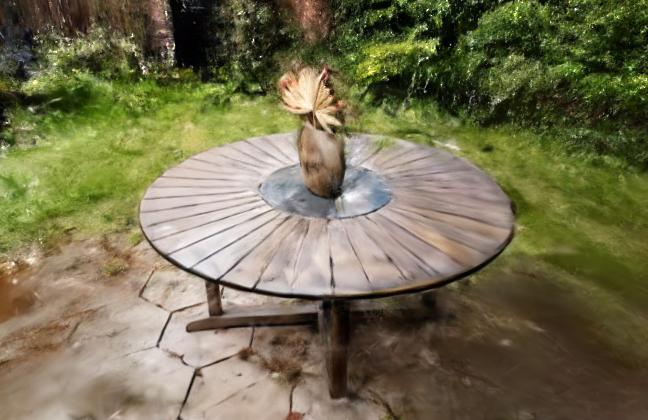} &
            \includegraphics[width=\linewidth]{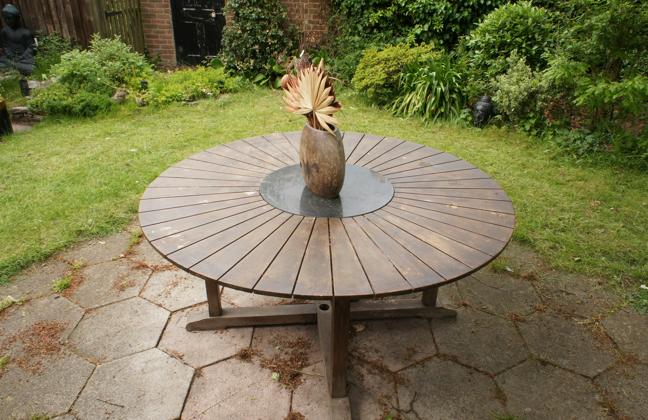} \\

        \end{tabular}
        
    \caption{\textbf{Ablation Study} on $9$-view reconstruction of \emph{garden} scene. Our fine-tuned artifact removal module and iterative schedule contribute the most toward quality of the final reconstruction. 3D Gaussians from \emph{Sparse 3DGS} act as suitable geometric prior in the absence of explicit view conditioning.}
    \label{fig:ablation_garden}
\end{figure}

In Tab \ref{tab:ablation} and Fig \ref{fig:ablation_garden}, we ablate the relative importance of each component towards $360^{\circ}$ reconstruction. We pick the garden scene and its $9$ view split for this study. We first show benefits of the regularization heuristics in \emph{Sparse 3DGS} over native 3DGS. We attempt reconstruction using either the in-painting or artifact removal module and show that their combination works best for both novel view synthesis and final reconstruction. Interestingly, leaving out the artifact removal module results in the best PSNR and SSIM scores across all variants, emphasizing that classical metrics reward blurry reconstructions, whereas FID and KID favor sharp, realistic details in novel views. We remove the iterative update formulation and add in-painted clean renders for all novel views simultaneously. This variant performs worse than \emph{Sparse 3DGS}, highlighting the need for autoregressive scene generation in our setting. Additionally, we try to supervise the renderings at novel views using the original 3DGS objective. However, this performs slightly worse than Eq \ref{eq:recon_obj}. Note that an SDS-based formulation~\cite{zeronvs} is not applicable here due to the two-step nature of generative view synthesis.

\begin{table}[!htbp]
\caption{\textbf{Ablation study} on the $9$ view split of \emph{garden} scene. Our combination of diffusion priors complements each other effectively. Without an iterative schedule to fuse novel views, our method fairs worse than a regularized baseline. Using the 3DGS objective for novel view renders leads to slightly worse performance on FID and KID.}
\centering
\begin{tabular}{lcccccc}
\toprule
Method & FID $\downarrow$ & KID $\downarrow$ & DISTS $\downarrow$ & LPIPS $\downarrow$ & PSNR $\uparrow$ & SSIM $\uparrow$\\
\midrule
3DGS & 223.594 & 0.146 & 0.247 & \cellcolor{tabfirst}{0.502} & 14.470 & 0.287 \\
Sparse 3DGS & \cellcolor{tabthird}{200.703} & 0.127 & 0.247 & \cellcolor{tabthird}{0.522} & \cellcolor{tabsecond}{16.589} & \cellcolor{tabsecond}{0.367} \\
\method{} w/o Artifact R. & 209.477 & 0.120 & 0.253 & 0.528 & \cellcolor{tabfirst}{16.850} & \cellcolor{tabfirst}{0.380} \\
\method{} w/o In-painting & \cellcolor{tabthird}{151.098} & \cellcolor{tabthird}{0.071} & \cellcolor{tabthird}{0.230} & 0.524 & 15.121 & 0.299 \\
\method{} w/o Schedule & 214.674 & 0.111 & 0.277 & 0.576 & 14.132 & 0.292 \\
\method{} w/ $\mathcal{L}_{D-SSIM}$ & \cellcolor{tabsecond}{133.875} & \cellcolor{tabsecond}{0.051} & \cellcolor{tabsecond}{0.225} & \cellcolor{tabfirst}{0.502} & 15.677 & \cellcolor{tabthird}{0.326} \\
\method{} & \cellcolor{tabfirst}{124.768} & \cellcolor{tabfirst}{0.048} & \cellcolor{tabfirst}{0.224} & \cellcolor{tabsecond}{0.504} & \cellcolor{tabthird}{15.759} & \cellcolor{tabthird}{0.326}\\
\bottomrule
\label{tab:ablation}
\end{tabular}
\end{table}

\vspace{-1.0mm}
\subsection{Scaling to More Views}

\begin{figure}[!t]
    \centering
    \begin{tabular}{c@{}*{6}{>{\centering\arraybackslash}p{0.165\linewidth}@{}}}
        & \includegraphics[width=\linewidth]{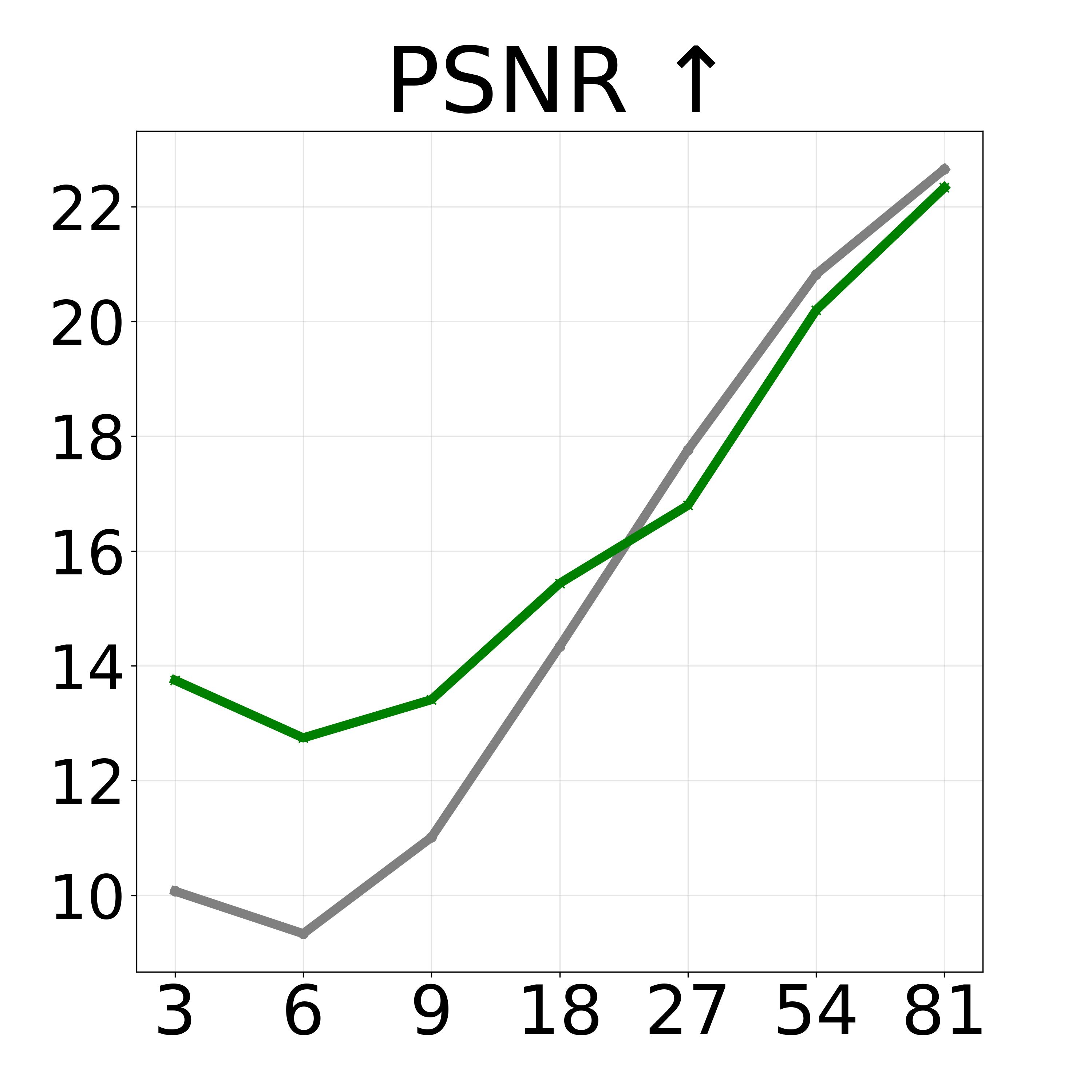} &
        \includegraphics[width=\linewidth]{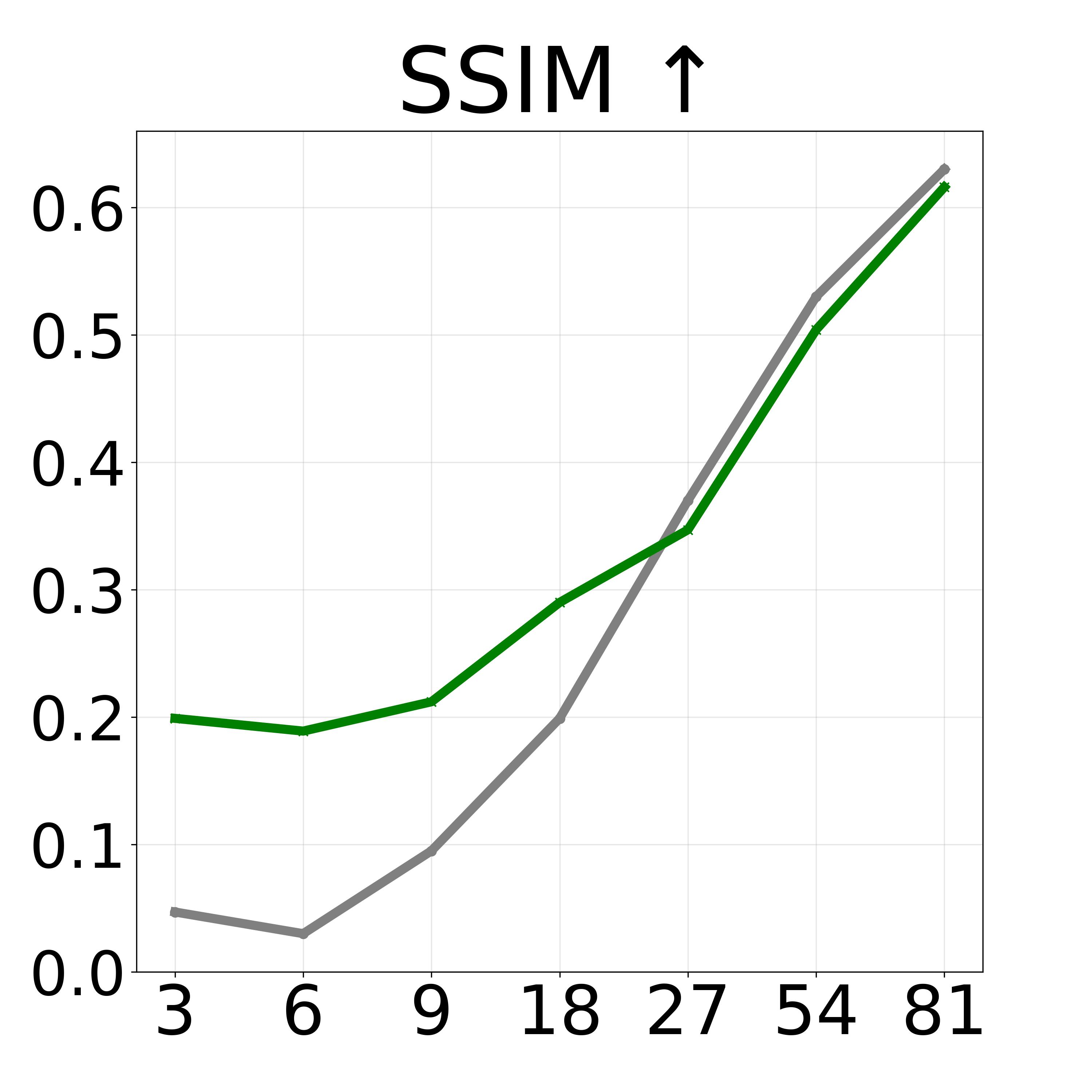} &
        \includegraphics[width=\linewidth]{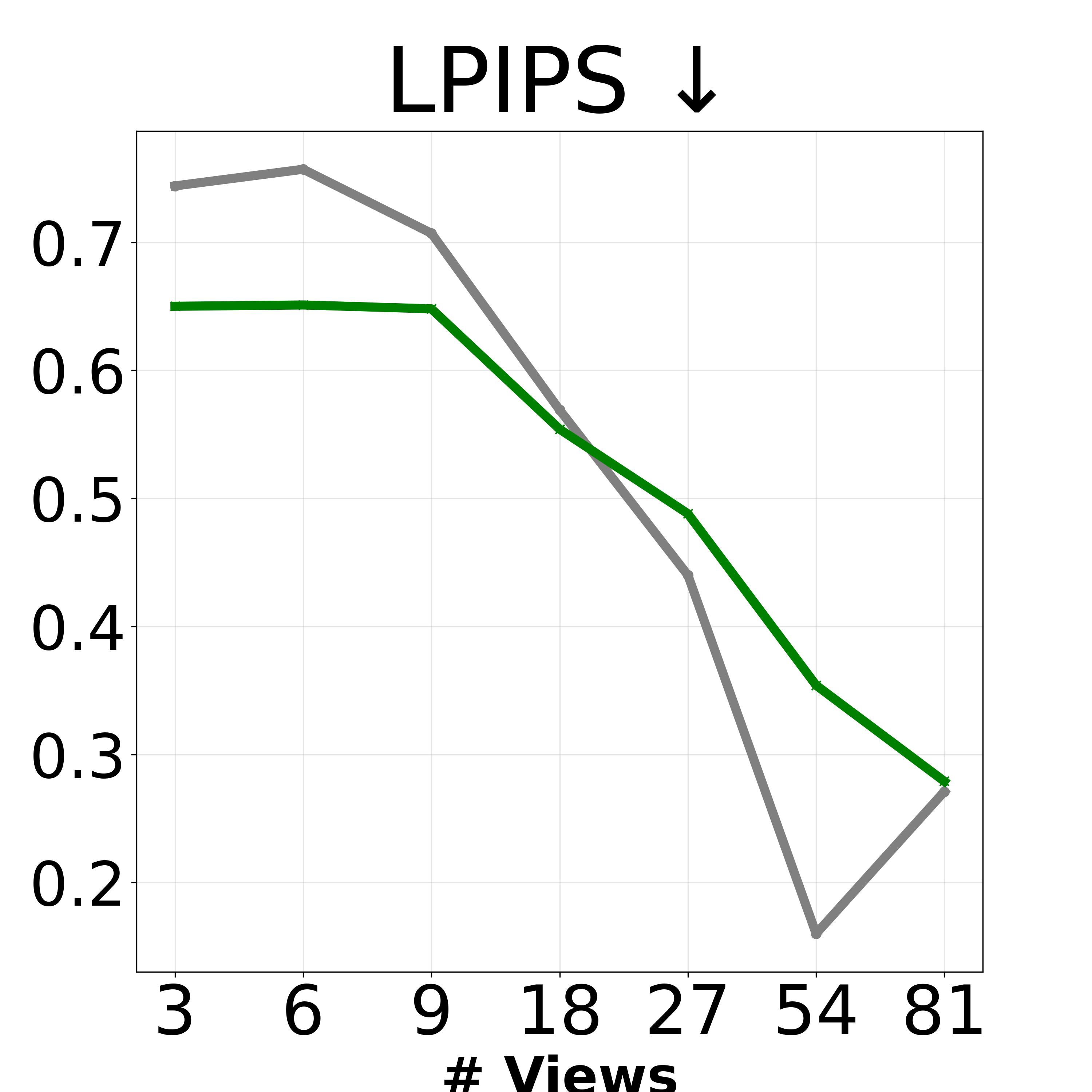} & \includegraphics[width=\linewidth]{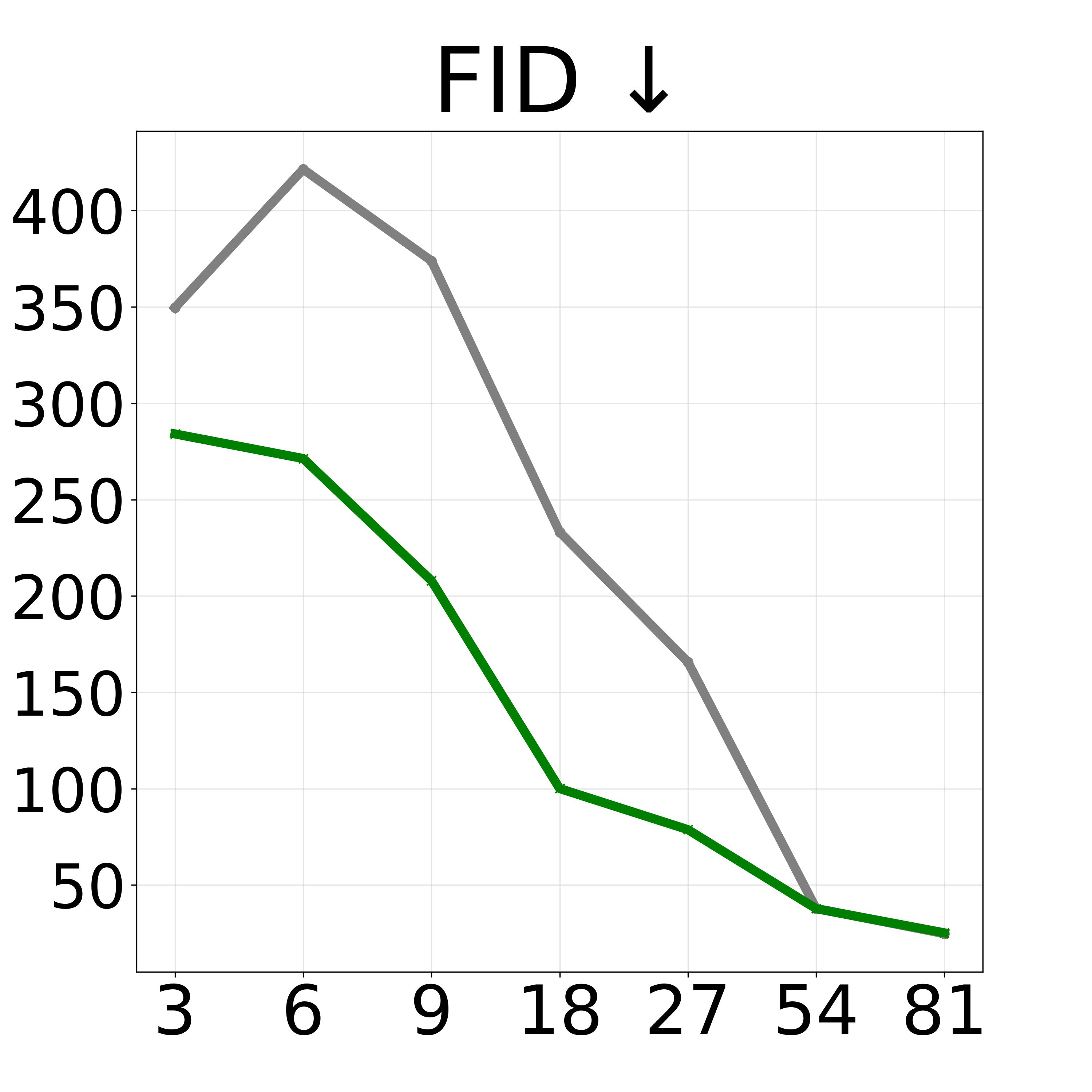} &
        \includegraphics[width=\linewidth]{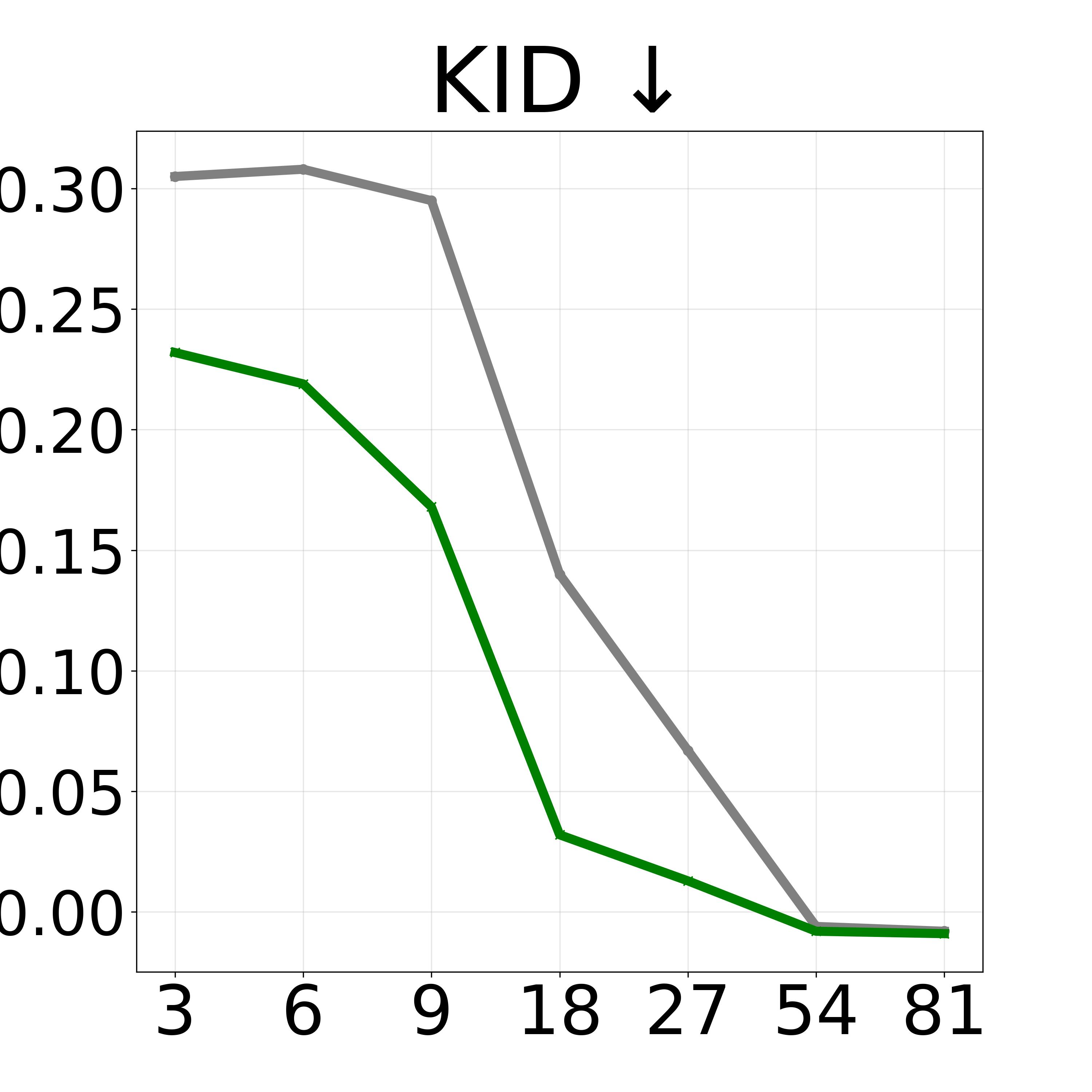} &
        \includegraphics[width=\linewidth]{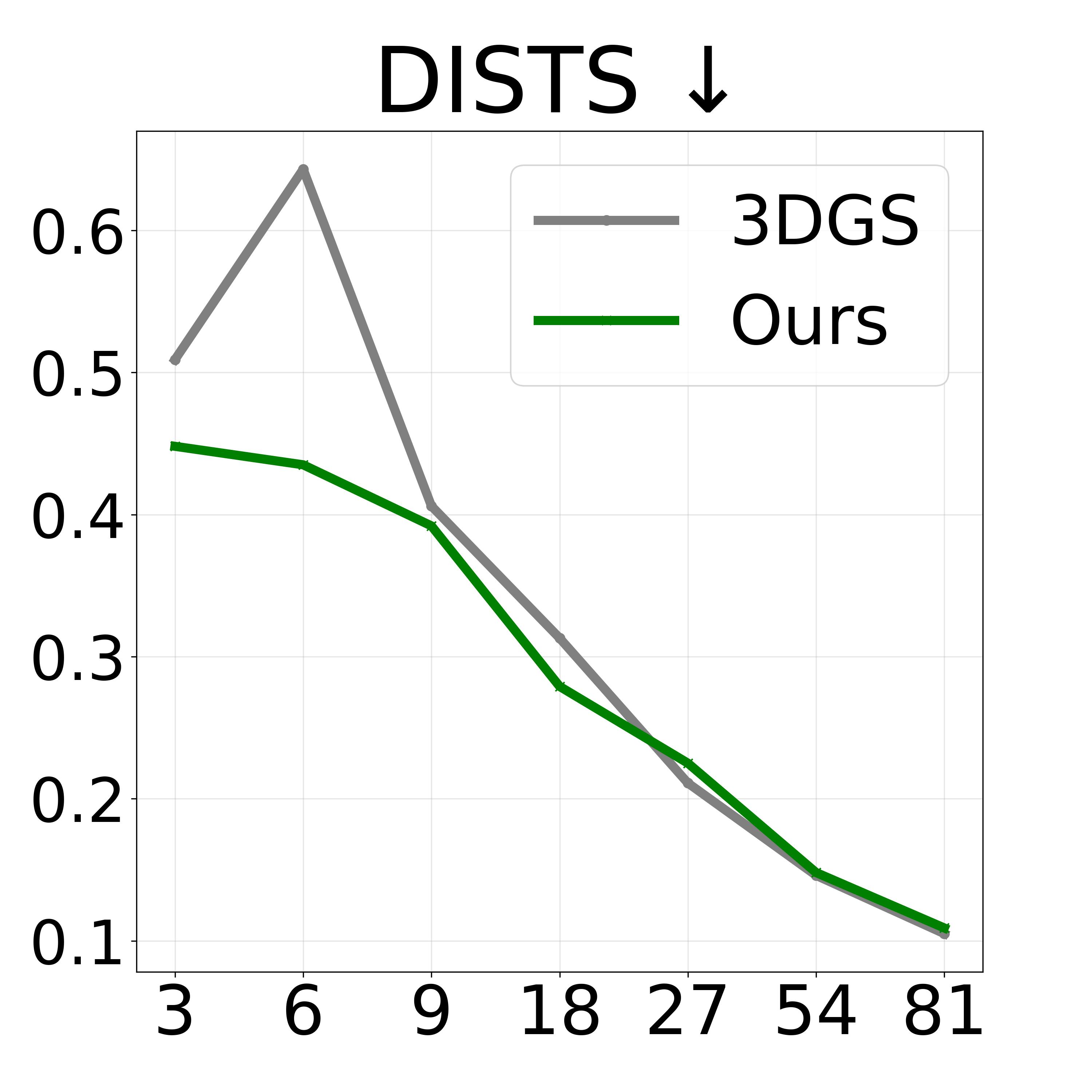}        
    \end{tabular}
    \caption{\textbf{Scalability of \method{} with input views.} Our combination of fine-tuned diffusion priors improves performance of 3DGS up to $27$ input views of the \emph{bicycle} scene, alleviating the need for dense captures.}
    \label{fig:scaling}
    \vspace{-3.0mm}
\end{figure}

\looseness=-1 In Fig \ref{fig:scaling}, we analyze the relevance of diffusion priors with increasing $M$. We evaluate \method{} and 3DGS on view splits of increasing size with  $M \in \{3, 6, 9, 18, 27, 54, 81\}$. For $M \leq 27$, our method consistently improves 3DGS' generalization at novel views. We observe that as ambiguities resolve with increasing scene coverage, diffusion-based regularization becomes less important, and for $M \geq 54$, our method performs either equally or slightly worse across the 6 performance measures. 

\vspace{-1.0mm}
\subsection{Efficiency}
\label{sec:efficiency}
Our approach focuses on limited use of multi-view data, minimal training times, and fast inference. We fit 3D Gaussians to our initial $M$ inputs inside $1$ hr and fine-tune the in-painting module in $\sim 2$ hrs. \method{} trains inside $30$ mins on a single A100 GPU, significantly faster than all baselines with NeRF backbones. FreeNeRF takes $\sim 1$ day while RegNeRF takes $>2$ days to train on a single A40 GPU. Both ZeroNVS* and DiffusioNeRF require $\sim 3$ hrs to distill 2D diffusion priors into Instant-NGP \cite{instantngp}, whereas ReconFusion trains in 1 hour on 8 A100 GPUs \footnote{Source: ReconFusion authors}. All 3 approaches use custom diffusion models that take several days to train on large-scale 3D datasets. On the contrary, we only need \emph{10.5K} samples to obtain a generalized artifact removal module that efficiently eliminates Gaussian artifacts and recovers image details in the foreground and background. We currently use a leave-one-out mechanism for the $9$ scenes in MipNeRF360 to train the artifact removal module for a particular MipNeRF360 scene. However, we only do this to have enough diverse data pairs across the $3$ datasets. This requirement can be easily alleviated using multi-view training data from other sources. Given the availability of a generalized artifact removal module, our entire pipeline can reconstruct a 360 scene in under $3.5$ hrs on a single A100 GPU. Courtesy of our 3D representation, we retain the real-time rendering capabilities of 3DGS.
\vspace{-2.0mm}
\section{Limitations \& Conclusion}
\label{sec:limitations}

\paragraph{Limitations}
Despite being a low-cost and efficient approach for reconstructing complex $360^{\circ}$ scenes from a few views, \method{} has certain limitations. Our approach is limited by the sparse geometry prior from an SfM point cloud, estimated from few views. For example, the point cloud for 9 views of the \emph{bicycle} scene only has $628$ points. This prevents our method from achieving even higher fidelity and restricting artifacts in distant, ambiguous novel views. DUSt3R \cite{wang2023dust3r} is a recent stereo reconstruction pipeline that can potentially provide a stronger geometry prior.

%DUSt3R estimates dense point cloud of a scene from as few as 2 input views. For the 9 views of our $bicycle$ scene, a DUSt3R point cloud contains $321k$ points. However, integrating this prior into our pipeline is not straightforward as the point cloud has heavy scale and alignment mismatch with ground truth camera poses \SP{Not sure how to frame this in a different way. ICP only solves our problem partly}. Registration algorithms like ICP \citep{Besl1992AMF} typically cannot produce a tight enough alignment, and DUSt3R only functions properly without any preset poses. Future research in this direction should drastically improve our reconstruction quality with dense priors for under-constrained settings.
%\JEL{No need to explain Dust3r here.}
\paragraph{Conclusion}

We present \method{}, a low-cost, data-efficient approach to reconstruct complex $360^{\circ}$ scenes from a few input images. We proposed a system that combines diffusion priors specialized for in-painting and Gaussian artifact removal to generate artificial novel views, which are iteratively added to our training image set. In our experiments we show that our approach approves over previous methods on the challenging MipNeRF360 dataset and illustrate the contributions of individual components in our ablation studies. For future work, we see the potential of our system to be improved with additional geometry cues from 3D vision foundation models.

% \begin{ack}
% Use unnumbered first level headings for the acknowledgments. All acknowledgments
% go at the end of the paper before the list of references. Moreover, you are required to declare funding (financial activities supporting the submitted work) and competing interests (related financial activities outside the submitted work).
% More information about this disclosure can be found at: \url{https://neurips.cc/Conferences/2024/PaperInformation/FundingDisclosure}. Do {\bf not} include this section in the anonymized submission, only in the final paper. You can use the \texttt{ack} environment provided in the style file to automatically hide this section in the anonymized submission.
% \end{ack}

\medskip
{
\small
\bibliographystyle{plain}
\bibliography{bibliography}

\begin{thebibliography}{10}

\bibitem{achiam2023gpt}
Josh Achiam, Steven Adler, Sandhini Agarwal, Lama Ahmad, Ilge Akkaya,
  Florencia~Leoni Aleman, Diogo Almeida, Janko Altenschmidt, Sam Altman,
  Shyamal Anadkat, et~al.
\newblock Gpt-4 technical report.
\newblock In {\em arXiv}, 2023.

\bibitem{barron2022mipnerf360}
Jonathan~T. Barron, Ben Mildenhall, Dor Verbin, Pratul~P. Srinivasan, and Peter
  Hedman.
\newblock Mip-nerf 360: Unbounded anti-aliased neural radiance fields.
\newblock In {\em CVPR}, 2022.

\bibitem{bińkowski2018demystifying}
Mikołaj Bińkowski, Dougal~J. Sutherland, Michael Arbel, and Arthur Gretton.
\newblock Demystifying {MMD} {GAN}s.
\newblock In {\em ICLR}, 2018.

\bibitem{instructpix2pix}
Tim Brooks, Aleksander Holynski, and Alexei~A. Efros.
\newblock Instructpix2pix: Learning to follow image editing instructions.
\newblock In {\em CVPR}, 2023.

\bibitem{genvs}
Eric~R. Chan, Koki Nagano, Matthew~A. Chan, Alexander~W. Bergman, Jeong~Joon
  Park, Axel Levy, Miika Aittala, Shalini~De Mello, Tero Karras, and Gordon
  Wetzstein.
\newblock {GeNVS}: Generative novel view synthesis with {3D}-aware diffusion
  models.
\newblock In {\em ICCV}, 2023.

\bibitem{pixelsplat}
David Charatan, Sizhe Li, Andrea Tagliasacchi, and Vincent Sitzmann.
\newblock pixelsplat: 3d gaussian splats from image pairs for scalable
  generalizable 3d reconstruction.
\newblock In {\em CVPR}, 2024.

\bibitem{mvsnerf}
Anpei Chen, Zexiang Xu, Fuqiang Zhao, Xiaoshuai Zhang, Fanbo Xiang, Jingyi Yu,
  and Hao Su.
\newblock Mvsnerf: Fast generalizable radiance field reconstruction from
  multi-view stereo.
\newblock In {\em ICCV}, 2021.

\bibitem{ssdnerf}
Hansheng Chen, Jiatao Gu, Anpei Chen, Wei Tian, Zhuowen Tu, Lingjie Liu, and
  Hao Su.
\newblock Single-stage diffusion nerf: A unified approach to 3d generation and
  reconstruction.
\newblock In {\em ICCV}, 2023.

\bibitem{fantasia3d}
Rui Chen, Yongwei Chen, Ningxin Jiao, and Kui Jia.
\newblock Fantasia3d: Disentangling geometry and appearance for high-quality
  text-to-3d content creation.
\newblock In {\em ICCV}, 2023.

\bibitem{mvsplat}
Yuedong Chen, Haofei Xu, Chuanxia Zheng, Bohan Zhuang, Marc Pollefeys, Andreas
  Geiger, Tat-Jen Cham, and Jianfei Cai.
\newblock Mvsplat: Efficient 3d gaussian splatting from sparse multi-view
  images.
\newblock 2024.

\bibitem{depthreggs}
Jaeyoung Chung, Jeongtaek Oh, and Kyoung~Mu Lee.
\newblock Depth-regularized optimization for 3d gaussian splatting in few-shot
  images.
\newblock In {\em arXiv}, 2023.

\bibitem{npg}
Devikalyan Das, Christopher Wewer, Raza Yunus, Eddy Ilg, and Jan~Eric Lenssen.
\newblock Neural parametric gaussians for monocular non-rigid object
  reconstruction.
\newblock In {\em CVPR}, 2024.

\bibitem{nerdi}
Congyue Deng, Chiyu Jiang, Charles~R Qi, Xinchen Yan, Yin Zhou, Leonidas
  Guibas, Dragomir Anguelov, et~al.
\newblock Nerdi: Single-view nerf synthesis with language-guided diffusion as
  general image priors.
\newblock In {\em CVPR}, 2023.

\bibitem{dsnerf}
Kangle Deng, Andrew Liu, Jun-Yan Zhu, and Deva Ramanan.
\newblock Depth-supervised {NeRF}: Fewer views and faster training for free.
\newblock In {\em CVPR}, 2022.

\bibitem{guideddiffusion}
Prafulla Dhariwal and Alexander Nichol.
\newblock Diffusion models beat gans on image synthesis.
\newblock In {\em NeurIPS}, 2021.

\bibitem{dists}
Keyan Ding, Kede Ma, Shiqi Wang, and Eero~P Simoncelli.
\newblock Image quality assessment: Unifying structure and texture similarity.
\newblock In {\em IEEE TPAMI}, 2020.

\bibitem{gal2022textual}
Rinon Gal, Yuval Alaluf, Yuval Atzmon, Or~Patashnik, Amit~H. Bermano, Gal
  Chechik, and Daniel Cohen-Or.
\newblock An image is worth one word: Personalizing text-to-image generation
  using textual inversion, 2022.

\bibitem{sugar}
Antoine Gu{\'e}don and Vincent Lepetit.
\newblock Sugar: Surface-aligned gaussian splatting for efficient 3d mesh
  reconstruction and high-quality mesh rendering.
\newblock In {\em CVPR}, 2024.

\bibitem{fe-nvs}
Pengsheng Guo, Miguel~Angel Bautista, Alex Colburn, Liang Yang, Daniel
  Ulbricht, Joshua~M. Susskind, and Qi~Shan.
\newblock Fast and explicit neural view synthesis.
\newblock In {\em WACV}, 2022.

\bibitem{instructnerf}
Ayaan Haque, Matthew Tancik, Alexei Efros, Aleksander Holynski, and Angjoo
  Kanazawa.
\newblock Instruct-nerf2nerf: Editing 3d scenes with instructions.
\newblock In {\em ICCV}, 2023.

\bibitem{10.1145/3272127.3275084}
Peter Hedman, Julien Philip, True Price, Jan-Michael Frahm, George Drettakis,
  and Gabriel Brostow.
\newblock Deep blending for free-viewpoint image-based rendering.
\newblock In {\em ACM TOG}, 2018.

\bibitem{wcr}
Philipp Henzler, Jeremy Reizenstein, Patrick Labatut, Roman Shapovalov, Tobias
  Ritschel, Andrea Vedaldi, and David Novotny.
\newblock Unsupervised learning of 3d object categories from videos in the
  wild.
\newblock In {\em CVPR}, 2021.

\bibitem{heusel2017gans}
Martin Heusel, Hubert Ramsauer, Thomas Unterthiner, Bernhard Nessler, and Sepp
  Hochreiter.
\newblock Gans trained by a two time-scale update rule converge to a local nash
  equilibrium.
\newblock In {\em NeurIPS}, 2017.

\bibitem{ddpm}
Jonathan Ho, Ajay Jain, and Pieter Abbeel.
\newblock Denoising diffusion probabilistic models.
\newblock In {\em NeurIPS}, 2020.

\bibitem{lora}
Edward~J Hu, Yelong Shen, Phillip Wallis, Zeyuan Allen-Zhu, Yuanzhi Li, Shean
  Wang, Lu~Wang, and Weizhu Chen.
\newblock Lo{RA}: Low-rank adaptation of large language models.
\newblock In {\em ICLR}, 2022.

\bibitem{2dgs}
Binbin Huang, Zehao Yu, Anpei Chen, Andreas Geiger, and Shenghua Gao.
\newblock 2d gaussian splatting for geometrically accurate radiance fields.
\newblock In {\em SIGGRAPH}, 2024.

\bibitem{neo360}
Muhammad~Zubair Irshad, Sergey Zakharov, Katherine Liu, Vitor Guizilini, Thomas
  Kollar, Adrien Gaidon, Zsolt Kira, and Rares Ambrus.
\newblock Neo 360: Neural fields for sparse view synthesis of outdoor scenes.
\newblock In {\em ICCV}, 2023.

\bibitem{dietnerf}
Ajay Jain, Matthew Tancik, and Pieter Abbeel.
\newblock Putting nerf on a diet: Semantically consistent few-shot view
  synthesis.
\newblock In {\em ICCV}, 2021.

\bibitem{jensen2014large}
Rasmus Jensen, Anders Dahl, George Vogiatzis, Engil Tola, and Henrik Aan{\ae}s.
\newblock Large scale multi-view stereopsis evaluation.
\newblock In {\em CVPR}, 2014.

\bibitem{edm}
Tero Karras, Miika Aittala, Timo Aila, and Samuli Laine.
\newblock Elucidating the design space of diffusion-based generative models.
\newblock In {\em NeurIPS}, 2022.

\bibitem{3dgs}
Bernhard Kerbl, Georgios Kopanas, Thomas Leimk{\"u}hler, and George Drettakis.
\newblock 3d gaussian splatting for real-time radiance field rendering.
\newblock In {\em ACM TOG}, 2023.

\bibitem{10.1145/3072959.3073599}
Arno Knapitsch, Jaesik Park, Qian-Yi Zhou, and Vladlen Koltun.
\newblock Tanks and temples: benchmarking large-scale scene reconstruction.
\newblock In {\em ACM TOG}, 2017.

\bibitem{dngaussian}
Jiahe Li, Jiawei Zhang, Xiao Bai, Jin Zheng, Xin Ning, Jun Zhou, and Lin Gu.
\newblock Dngaussian: Optimizing sparse-view 3d gaussian radiance fields with
  global-local depth normalization.
\newblock In {\em CVPR}, 2024.

\bibitem{magic3d}
Chen-Hsuan Lin, Jun Gao, Luming Tang, Towaki Takikawa, Xiaohui Zeng, Xun Huang,
  Karsten Kreis, Sanja Fidler, Ming-Yu Liu, and Tsung-Yi Lin.
\newblock Magic3d: High-resolution text-to-3d content creation.
\newblock In {\em CVPR}, 2023.

\bibitem{visionnerf}
Kai-En Lin, Lin Yen-Chen, Wei-Sheng Lai, Tsung-Yi Lin, Yi-Chang Shih, and Ravi
  Ramamoorthi.
\newblock Vision transformer for nerf-based view synthesis from a single input
  image.
\newblock In {\em WACV}, 2023.

\bibitem{zero1to3}
Ruoshi Liu, Rundi Wu, Basile Van~Hoorick, Pavel Tokmakov, Sergey Zakharov, and
  Carl Vondrick.
\newblock Zero-1-to-3: Zero-shot one image to 3d object.
\newblock In {\em ICCV}, 2023.

\bibitem{dynamicgaussians}
Jonathon Luiten, Georgios Kopanas, Bastian Leibe, and Deva Ramanan.
\newblock Dynamic 3d gaussians: Tracking by persistent dynamic view synthesis.
\newblock In {\em 3DV}, 2024.

\bibitem{pc2}
Luke Melas-Kyriazi, Christian Rupprecht, and Andrea Vedaldi.
\newblock Pc2: Projection-conditioned point cloud diffusion for single-image 3d
  reconstruction.
\newblock In {\em CVPR}, 2023.

\bibitem{meng2022sdedit}
Chenlin Meng, Yutong He, Yang Song, Jiaming Song, Jiajun Wu, Jun-Yan Zhu, and
  Stefano Ermon.
\newblock {SDE}dit: Guided image synthesis and editing with stochastic
  differential equations.
\newblock In {\em International Conference on Learning Representations}, 2022.

\bibitem{nerf}
Ben Mildenhall, Pratul~P. Srinivasan, Matthew Tancik, Jonathan~T. Barron, Ravi
  Ramamoorthi, and Ren Ng.
\newblock Nerf: Representing scenes as neural radiance fields for view
  synthesis.
\newblock In {\em ECCV}, 2020.

\bibitem{diffrf}
Norman M{\"u}ller, Yawar Siddiqui, Lorenzo Porzi, Samuel~Rota Bulo, Peter
  Kontschieder, and Matthias Nie{\ss}ner.
\newblock Diffrf: Rendering-guided 3d radiance field diffusion.
\newblock In {\em CVPR}, 2023.

\bibitem{instantngp}
Thomas M\"uller, Alex Evans, Christoph Schied, and Alexander Keller.
\newblock Instant neural graphics primitives with a multiresolution hash
  encoding.
\newblock In {\em ACM TOG}, 2022.

\bibitem{glide}
Alexander~Quinn Nichol, Prafulla Dhariwal, Aditya Ramesh, Pranav Shyam, Pamela
  Mishkin, Bob Mcgrew, Ilya Sutskever, and Mark Chen.
\newblock {GLIDE}: Towards photorealistic image generation and editing with
  text-guided diffusion models.
\newblock In {\em ICML}, 2022.

\bibitem{regnerf}
Michael Niemeyer, Jonathan~T. Barron, Ben Mildenhall, Mehdi S.~M. Sajjadi,
  Andreas Geiger, and Noha Radwan.
\newblock Regnerf: Regularizing neural radiance fields for view synthesis from
  sparse inputs.
\newblock In {\em CVPR}, 2022.

\bibitem{dreamfusion}
Ben Poole, Ajay Jain, Jonathan~T. Barron, and Ben Mildenhall.
\newblock Dreamfusion: Text-to-3d using 2d diffusion.
\newblock In {\em ICLR}, 2023.

\bibitem{clip}
Alec Radford, Jong~Wook Kim, Chris Hallacy, Aditya Ramesh, Gabriel Goh,
  Sandhini Agarwal, Girish Sastry, Amanda Askell, Pamela Mishkin, Jack Clark,
  et~al.
\newblock Learning transferable visual models from natural language
  supervision.
\newblock In {\em ICML}, 2021.

\bibitem{dalle2}
Aditya Ramesh, Prafulla Dhariwal, Alex Nichol, Casey Chu, and Mark Chen.
\newblock Hierarchical text-conditional image generation with clip latents.
\newblock In {\em arXiv}, 2022.

\bibitem{Ranftl2020}
Ren\'{e} Ranftl, Katrin Lasinger, David Hafner, Konrad Schindler, and Vladlen
  Koltun.
\newblock Towards robust monocular depth estimation: Mixing datasets for
  zero-shot cross-dataset transfer.
\newblock In {\em IEEE TPAMI}, 2020.

\bibitem{reizenstein2021common}
Jeremy Reizenstein, Roman Shapovalov, Philipp Henzler, Luca Sbordone, Patrick
  Labatut, and David Novotny.
\newblock Common objects in 3d: Large-scale learning and evaluation of
  real-life 3d category reconstruction.
\newblock In {\em ICCV}, 2021.

\bibitem{roessle2022depthpriorsnerf}
Barbara Roessle, Jonathan~T. Barron, Ben Mildenhall, Pratul~P. Srinivasan, and
  Matthias Nie{\ss}ner.
\newblock Dense depth priors for neural radiance fields from sparse input
  views.
\newblock In {\em CVPR}, 2022.

\bibitem{ldm}
Robin Rombach, Andreas Blattmann, Dominik Lorenz, Patrick Esser, and Bj\"orn
  Ommer.
\newblock High-resolution image synthesis with latent diffusion models.
\newblock In {\em CVPR}, 2022.

\bibitem{dreambooth}
Nataniel Ruiz, Yuanzhen Li, Varun Jampani, Yael Pritch, Michael Rubinstein, and
  Kfir Aberman.
\newblock Dreambooth: Fine tuning text-to-image diffusion models for
  subject-driven generation.
\newblock In {\em CVPR}, 2023.

\bibitem{imagen}
Chitwan Saharia, William Chan, Saurabh Saxena, Lala Li, Jay Whang, Emily~L
  Denton, Kamyar Ghasemipour, Raphael Gontijo~Lopes, Burcu Karagol~Ayan, Tim
  Salimans, Jonathan Ho, David~J Fleet, and Mohammad Norouzi.
\newblock Photorealistic text-to-image diffusion models with deep language
  understanding.
\newblock In {\em NeurIPS}, 2022.

\bibitem{zeronvs}
Kyle Sargent, Zizhang Li, Tanmay Shah, Charles Herrmann, Hong-Xing Yu, Yunzhi
  Zhang, Eric~Ryan Chan, Dmitry Lagun, Li~Fei-Fei, Deqing Sun, and Jiajun Wu.
\newblock {ZeroNVS}: Zero-shot 360-degree view synthesis from a single real
  image.
\newblock In {\em CVPR}, 2024.

\bibitem{schoenberger2016sfm}
Johannes~Lutz Sch\"{o}nberger and Jan-Michael Frahm.
\newblock Structure-from-motion revisited.
\newblock In {\em CVPR}, 2016.

\bibitem{schoenberger2016mvs}
Johannes~Lutz Sch\"{o}nberger, Enliang Zheng, Marc Pollefeys, and Jan-Michael
  Frahm.
\newblock Pixelwise view selection for unstructured multi-view stereo.
\newblock In {\em ECCV}, 2016.

\bibitem{npcd}
Philipp Schröppel, Christopher Wewer, Jan~Eric Lenssen, Eddy Ilg, and Thomas
  Brox.
\newblock Neural point cloud diffusion for disentangled 3d shape and appearance
  generation.
\newblock In {\em CVPR}, 2024.

\bibitem{laion5b}
Christoph Schuhmann, Romain Beaumont, Richard Vencu, Cade Gordon, Ross
  Wightman, Mehdi Cherti, Theo Coombes, Aarush Katta, Clayton Mullis, Mitchell
  Wortsman, Patrick Schramowski, Srivatsa Kundurthy, Katherine Crowson, Ludwig
  Schmidt, Robert Kaczmarczyk, and Jenia Jitsev.
\newblock Laion-5b: An open large-scale dataset for training next generation
  image-text models.
\newblock In {\em NeurIPS}, 2022.

\bibitem{triplanediffusion}
J.~Ryan Shue, Eric~Ryan Chan, Ryan Po, Zachary Ankner, Jiajun Wu, and Gordon
  Wetzstein.
\newblock 3d neural field generation using triplane diffusion.
\newblock In {\em CVPR}, 2023.

\bibitem{ddim}
Jiaming Song, Chenlin Meng, and Stefano Ermon.
\newblock Denoising diffusion implicit models.
\newblock In {\em ICLR}, 2021.

\bibitem{dreamgaussian}
Jiaxiang Tang, Jiawei Ren, Hang Zhou, Ziwei Liu, and Gang Zeng.
\newblock Dreamgaussian: Generative gaussian splatting for efficient 3d content
  creation.
\newblock In {\em ICLR}, 2024.

\bibitem{tang2023realfill}
Luming Tang, Nataniel Ruiz, Chu Qinghao, Yuanzhen Li, Aleksander Holynski,
  David~E Jacobs, Bharath Hariharan, Yael Pritch, Neal Wadhwa, Kfir Aberman,
  and Michael Rubinstein.
\newblock Realfill: Reference-driven generation for authentic image completion.
\newblock In {\em arXiv}, 2023.

\bibitem{grf}
Alex Trevithick and Bo~Yang.
\newblock Grf: Learning a general radiance field for 3d scene representation
  and rendering.
\newblock In {\em ICCV}, 2021.

\bibitem{sparsenerf}
Guangcong Wang, Zhaoxi Chen, Chen~Change Loy, and Ziwei Liu.
\newblock Sparsenerf: Distilling depth ranking for few-shot novel view
  synthesis.
\newblock In {\em ICCV}, 2023.

\bibitem{scorejacobianchaining}
Haochen Wang, Xiaodan Du, Jiahao Li, Raymond~A Yeh, and Greg Shakhnarovich.
\newblock Score jacobian chaining: Lifting pretrained 2d diffusion models for
  3d generation.
\newblock In {\em CVPR}, 2023.

\bibitem{neus}
Peng Wang, Lingjie Liu, Yuan Liu, Christian Theobalt, Taku Komura, and Wenping
  Wang.
\newblock Neus: Learning neural implicit surfaces by volume rendering for
  multi-view reconstruction.
\newblock In {\em NeurIPS}, 2021.

\bibitem{wang2023dust3r}
Shuzhe Wang, Vincent Leroy, Yohann Cabon, Boris Chidlovskii, and Jerome Revaud.
\newblock Dust3r: Geometric 3d vision made easy.
\newblock In {\em CVPR}, 2024.

\bibitem{simnp}
Christopher Wewer, Eddy Ilg, Bernt Schiele, and Jan~Eric Lenssen.
\newblock {SimNP}: Learning self-similarity priors between neural points.
\newblock In {\em ICCV}, 2023.

\bibitem{latentsplat}
Christopher Wewer, Kevin Raj, Eddy Ilg, Bernt Schiele, and Jan~Eric Lenssen.
\newblock latentsplat: Autoencoding variational gaussians for fast
  generalizable 3d reconstruction.
\newblock In {\em arXiv}, 2024.

\bibitem{4dgs}
Guanjun Wu, Taoran Yi, Jiemin Fang, Lingxi Xie, Xiaopeng Zhang, Wei Wei, Wenyu
  Liu, Qi~Tian, and Wang Xinggang.
\newblock 4d gaussian splatting for real-time dynamic scene rendering.
\newblock 2024.

\bibitem{reconfusion}
Rundi Wu, Ben Mildenhall, Philipp Henzler, Keunhong Park, Ruiqi Gao, Daniel
  Watson, Pratul~P. Srinivasan, Dor Verbin, Jonathan~T. Barron, Ben Poole, and
  Aleksander Holynski.
\newblock Reconfusion: 3d reconstruction with diffusion priors.
\newblock In {\em CVPR}, 2024.

\bibitem{diffusionerf}
Jamie Wynn and Daniyar Turmukhambetov.
\newblock {DiffusioNeRF: Regularizing Neural Radiance Fields with Denoising
  Diffusion Models}.
\newblock In {\em CVPR}, 2023.

\bibitem{sparsegs}
Haolin Xiong, Sairisheek Muttukuru, Rishi Upadhyay, Pradyumna Chari, and Achuta
  Kadambi.
\newblock Sparsegs: Real-time 360° sparse view synthesis using gaussian
  splatting.
\newblock In {\em arXiv}, 2023.

\bibitem{Yang_2023_CVPR}
Jiawei Yang, Marco Pavone, and Yue Wang.
\newblock Freenerf: Improving few-shot neural rendering with free frequency
  regularization.
\newblock In {\em CVPR}, 2023.

\bibitem{pixelnerf}
Alex Yu, Vickie Ye, Matthew Tancik, and Angjoo Kanazawa.
\newblock {pixelNeRF}: Neural radiance fields from one or few images.
\newblock In {\em CVPR}, 2021.

\bibitem{lpips}
Richard Zhang, Phillip Isola, Alexei~A Efros, Eli Shechtman, and Oliver Wang.
\newblock The unreasonable effectiveness of deep features as a perceptual
  metric.
\newblock In {\em CVPR}, 2018.

\bibitem{pvdiffusion}
Linqi Zhou, Yilun Du, and Jiajun Wu.
\newblock 3d shape generation and completion through point-voxel diffusion.
\newblock In {\em ICCV}, 2021.

\bibitem{zhou2018stereo}
Tinghui Zhou, Richard Tucker, John Flynn, Graham Fyffe, and Noah Snavely.
\newblock Stereo magnification: Learning view synthesis using multiplane
  images.
\newblock In {\em SIGGRAPH}, 2018.

\bibitem{sparsefusion}
Zhizhuo Zhou and Shubham Tulsiani.
\newblock Sparsefusion: Distilling view-conditioned diffusion for 3d
  reconstruction.
\newblock In {\em CVPR}, 2023.

\bibitem{fsgs}
Zehao Zhu, Zhiwen Fan, Yifan Jiang, and Zhangyang Wang.
\newblock Fsgs: Real-time few-shot view synthesis using gaussian splatting.
\newblock In {\em arXiv}, 2023.

\end{thebibliography}
}

%%%%%%%%%%%%%%%%%%%%%%%%%%%%%%%%%%%%%%%%%%%%%%%%%%%%%%%%%%%
\appendix
\section{Appendix / Supplemental Material}

In this section, we provide further details on certain heuristics and design choices of our pipeline not covered in the main paper. We begin with our strategy for creating $M$-view subsets of a scene (c.f~\ref{subsec:view_split}). We then provide details on the regularization introduced in our improved baseline to adapt 3DGS to sparse-view reconstruction (c.f~\ref{subsec:geometry_baseline_app}). Next, we compare our Gaussian artifact removal module with existing image-to-image diffusion models and show the importance of finetuned priors for adapting to this task (c.f~\ref{subsec:img2img_baselines}). We perform a toy experiment to determine the best choice of parameters for our iterative schedule (c.f.~\ref{subsec:iterative_update}). We end with some minor details on the distillation procedure to lift 2D diffusion priors to 3D Gaussians (c.f.~\ref{subsec:distillation}). Please also note our supplemental video, showing 360° reconstruction results.

\vspace{-1.0mm}
\subsection{Creating M-view Sparse Sets}
\label{subsec:view_split}

3DGS relies on COLMAP for a sparse point cloud initialization, and COLMAP requires sufficient feature correspondence across the input images to produce a 3D consistent sparse scene geometry. To produce a subset where the views are far apart to have sufficient scene coverage and at the same time have enough feature correspondence among the images to produce a 3D consistent (albeit sparse) point cloud, we devise a well-defined strategy such that view subsets across different scenes follow a certain heuristic and are not just randomly picked.
% Among related literature, only ReconFusion proposes a heuristic to systemically synthesize sparse subsets from the full set, while SparseGS \cite{sparsegs} and FSGS \cite{fsgs} randomly pick 12 and 24 images, respectively, for MipNeRF360. However, the implementation details in ReconFusion are too scant for us to reproduce the exact splits, and there is no open-source code available yet.\\

We are given a set $\mathcal{X} = \{\mathbf{I}_{i}, \pi_{i}\}_{i=1}^{N}$ of images and corresponding camera poses of a scene where $\mathbf{I}_{i} \in \mathbb{R}^{h \times w \times 3}$ and $\pi_{i} = [\mathrm{K}^{i} | \mathrm{R}^{i} | \mathrm{t}^{i} ] \in SE(3)$. Here $K^{i}$ is the camera intrinsics matrix, $\mathrm{R}^{i} \in \mathbb{R}^{3 \times 3}$ is the rotation matrix, and $t^{i} \in \mathbb{R}^{3 \times 1}$ is the translation vector. From this dense set, we aim to create $M$-view sparse sets $ (M < N) $ $\{\mathbf{I}_{k}, \mathrm{\pi}_{k}\}_{k=1}^{M} $ with a corresponding sparse point cloud $P_M \in \mathbb{R}^{S \times 3}$, obtained via COLMAP. To maximize the scene coverage of the $M$-view subset, we use the SE(3) (Special Euclidean group in three dimensions) representation to define a geodesic distance between a pair of camera viewpoints. SE(3) represents all rigid transformations (translations and rotations) that can be applied to a three-dimensional space while preserving distances and angles. Mathematically, SE(3) is defined as:

\begin{equation}
\label{eq_se3}
\begin{aligned}
SE(3) = \{(\mathrm{R}, t) \vert \mathrm{R} \in SO(3),t \in \mathbb{R}^{3} \}
\end{aligned}    
\end{equation}

\noindent
Here, $SO(3)$ is the special orthogonal group in three dimensions, representing all possible 3D rotations without reflection or scaling. To quantify ``closeness'' between 2 SE(3) elements $\mathrm{\pi}_{1}$ and $\mathrm{\pi}_{2}$, we use the geodesic distance that measures the shortest path between 2 viewpoints on a curved surface - in this case, the minimum distance along a curve in SE(3) that connects $\mathrm{\pi}_{1}$ and $\mathrm{\pi}_{2}$. One common approach of measuring the geodesic distance involves separating the rotational and translational components and combining individual distances through a weighted average. If $\mathrm{R_1}$, $\mathrm{T_1}$ and $\mathrm{R_2}$, $\mathrm{T_2}$ are the rotation and translation matrices corresponding to $\mathrm{\pi}_{1}$ and $\mathrm{\pi}_{2}$ respectively, the geodesic distance can be defined as:

\begin{equation}
\label{eq_geodesic_distance}
\begin{aligned}
\mathcal{D}_{geodesic} = \lVert \textnormal{Rodrigues}(\mathrm{R}_{rel}) \rVert + w_T \lVert\mathrm{T_1} - \mathrm{T_2}\rVert
\end{aligned}
\end{equation}

\noindent
where the Rodrigues formula determines the relative angle of rotation $\theta$ between 2 matrices $\mathrm{R_1}$ and $\mathrm{R_2}$ by measuring the angle of the difference rotation matrix $\mathrm{R}_{rel} = \mathrm{R_1}\mathrm{R_2}^{T}$ as follows: 

\begin{equation}
\label{eq_rodrigues}
\begin{aligned}
\textnormal{Rodrigues}(\mathrm{R}_{rel}) = \mathcal{\theta} = \cos^{-1}\left( \frac{\text{trace}(R_1 R_2^{T}) - 1}{2}  \right)
\end{aligned}
\end{equation}

$w_T$ is a weighting factor. We set $w_T = 0.1$ for all our experiments as we usually find that rotational distance has more impact on scene coverage given an $M$-view constraint.

Suppose the size of the current view stack is $n_s < M$, and now we want to add a new viewpoint $\mathrm{\pi}_{k}$ from a stack of $(N - n_s)$ available viewpoints. Now, we can either pick the closest, the $2^{nd}$ closest, or the $n^{th}$ closest viewpoint with respect to current stack $\{\mathrm{\pi}_{j}\}_{j=1}^{n_s}$. We calculate a matrix of geodesic distances of dimension $n_s \times (N - n_s)$ between $\{\mathrm{\pi}_{j}\}_{j=1}^{n_s}$ and $\{\mathrm{\pi}_{j}\}_{j=n_s + 1}^{N}$ and pick the $n^{th}$ closest viewpoint from $\{\mathrm{\pi}_{j}\}_{j=n_s + 1}^{N}$. This process continues until we have $M$ total viewpoints in our stack. We observe experimentally that, as we go on increasing $n$, $\underset{i, j}{\max} \, \mathcal{D}^{\mathrm{\pi}}_{i,j}(n)$, i.e., the maximum geodesic distance across all pairs of viewpoints in $\{\mathrm{\pi}_{j}\}_{j=1}^{M}$, also shows an increasing trend (non-monotonous), while the size of $P_M$ shows a decreasing trend (also non-monotonous). After a certain value of $n$, COLMAP can no longer find enough feature correspondences among image pairs to register all images to a single point cloud $P_M$. This is where we terminate and pick the $M$-view subset corresponding to $\underset{n}{\max} \, \mathcal{D}^{\mathrm{\pi}}(n)$.

\begin{figure}[!htbp]
    \centering
    \begin{subfigure}[!htbp]{0.49\linewidth}
        \centering
        \includegraphics[width=\linewidth]{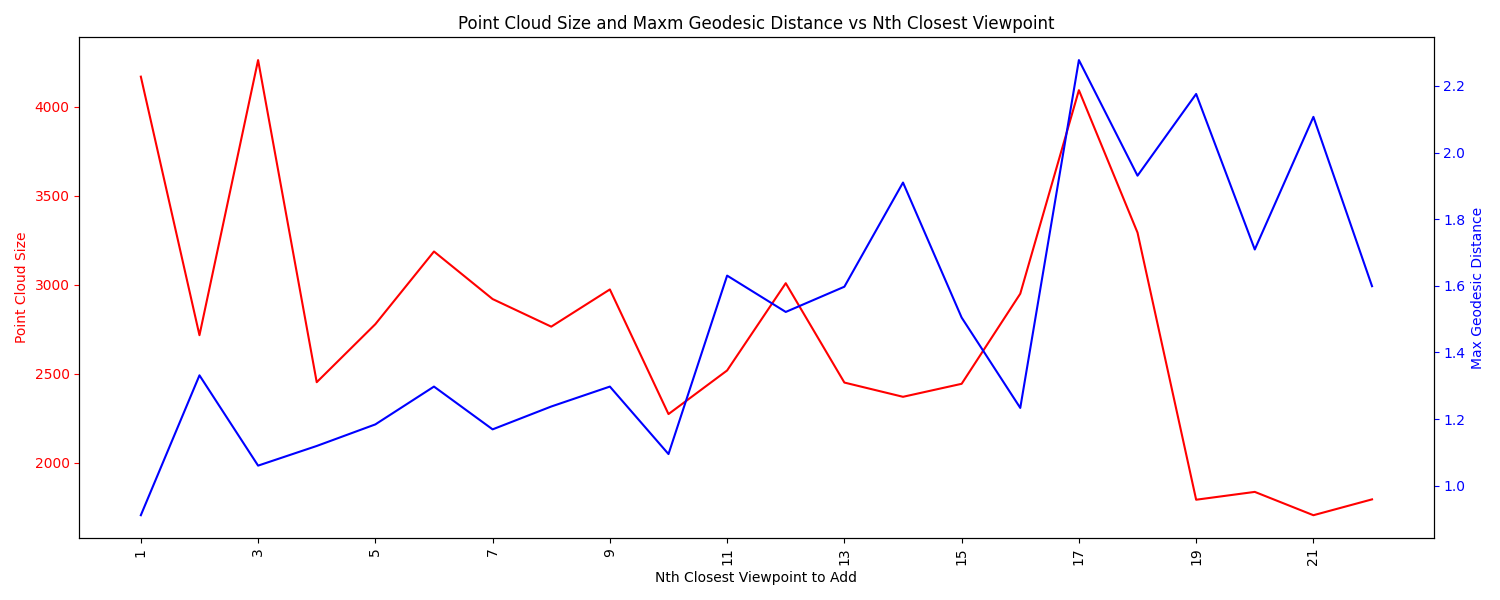}
        \caption{$M=9$ in the $treehill$ scene}
    \end{subfigure}\hfill
    \begin{subfigure}[!htbp]{0.49\linewidth}
        \centering
        \includegraphics[width=\linewidth]{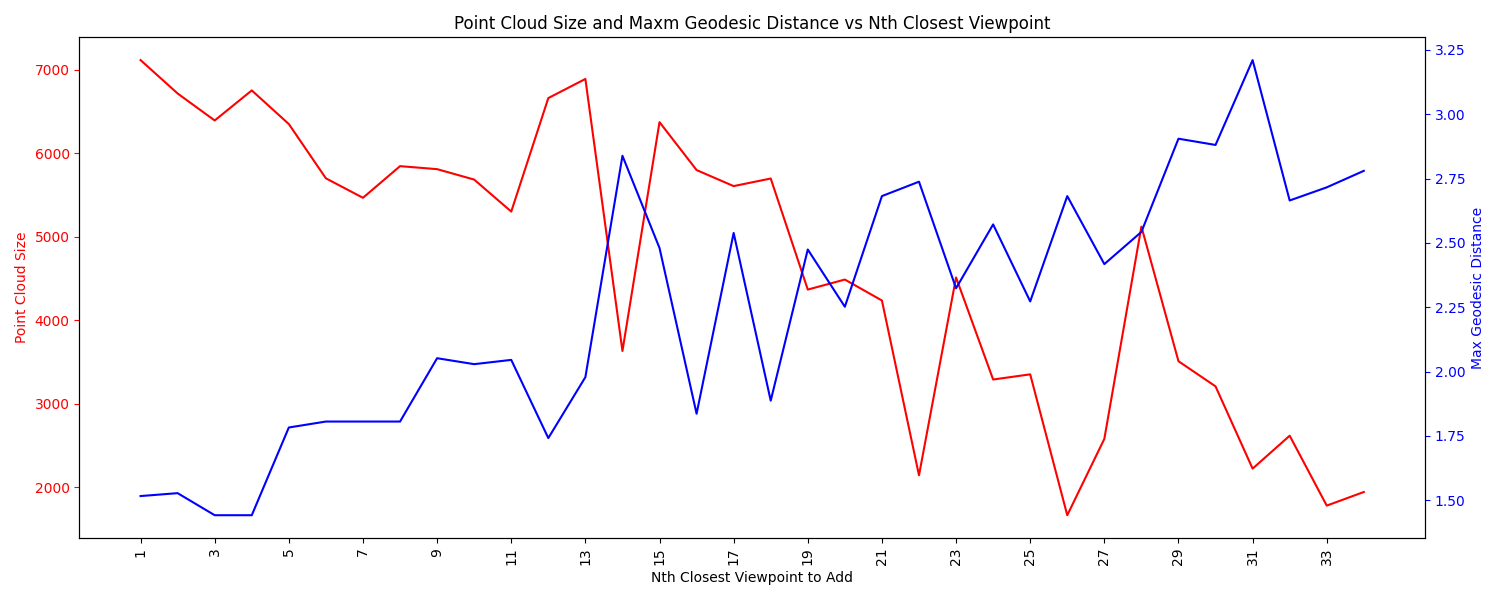}
        \caption{$M=18$ in the $stump$ scene}
    \end{subfigure}
    % Single caption and label for the whole figure
    \caption{Plot of \textcolor{blue}{maximum geodesic distance} across training viewpoints and \textcolor{red}{SfM point cloud size} vs. $n$ - the index of the $n^{th}$ closest viewpoint from the available training stack. The locations of the maxima in the \textcolor{blue}{blue plots} are used to create our view subset of size $M$. }
    \label{fig_view_split}
\end{figure}

\begin{figure}[!htbp]
    \centering
    \begin{subfigure}[!htbp]{0.4\linewidth}
        \centering
        \includegraphics[width=0.9\linewidth]{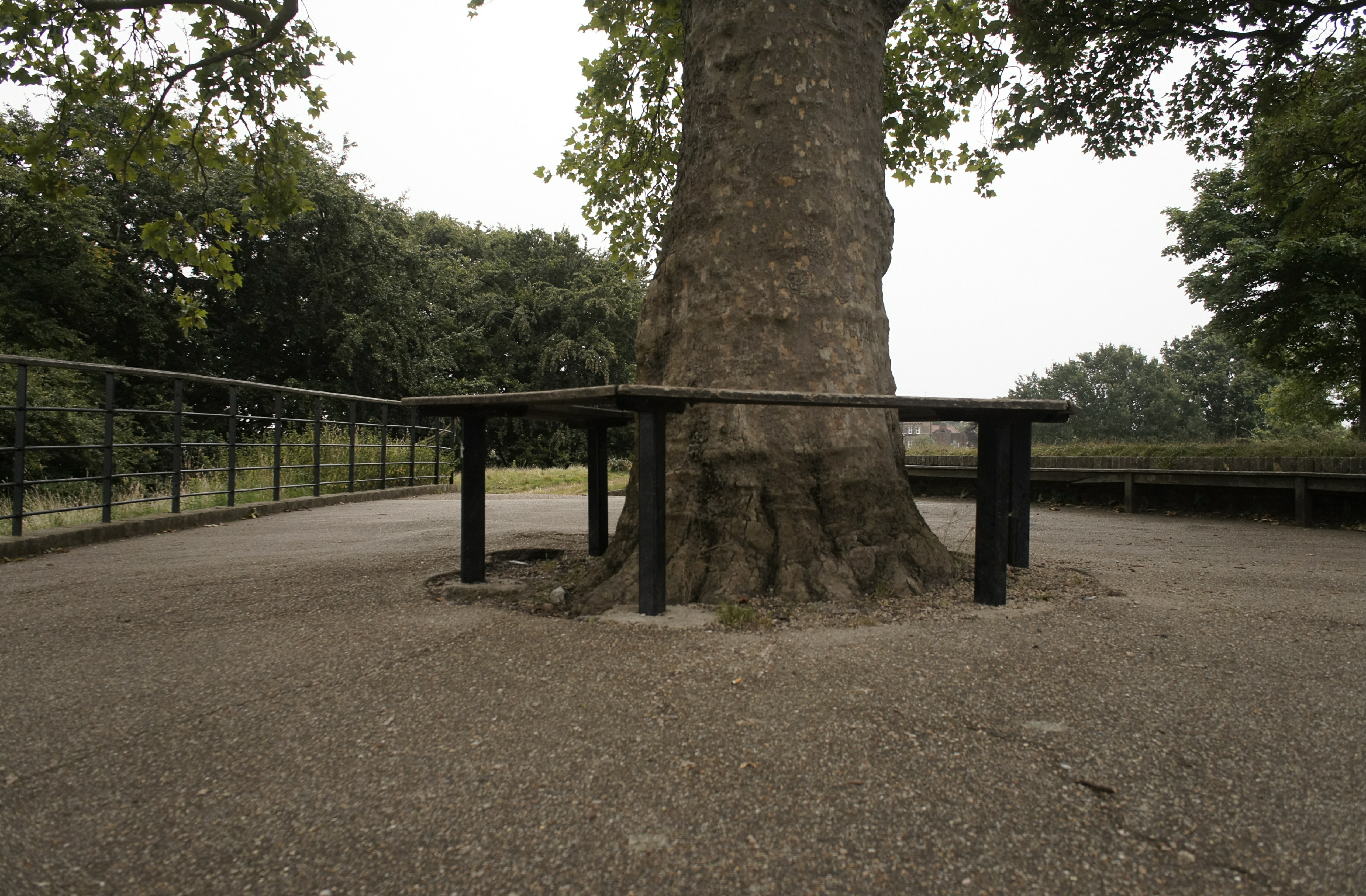}
    \end{subfigure}
    \begin{subfigure}[!htbp]{0.4\linewidth}
        \centering
        \includegraphics[width=0.9\linewidth]{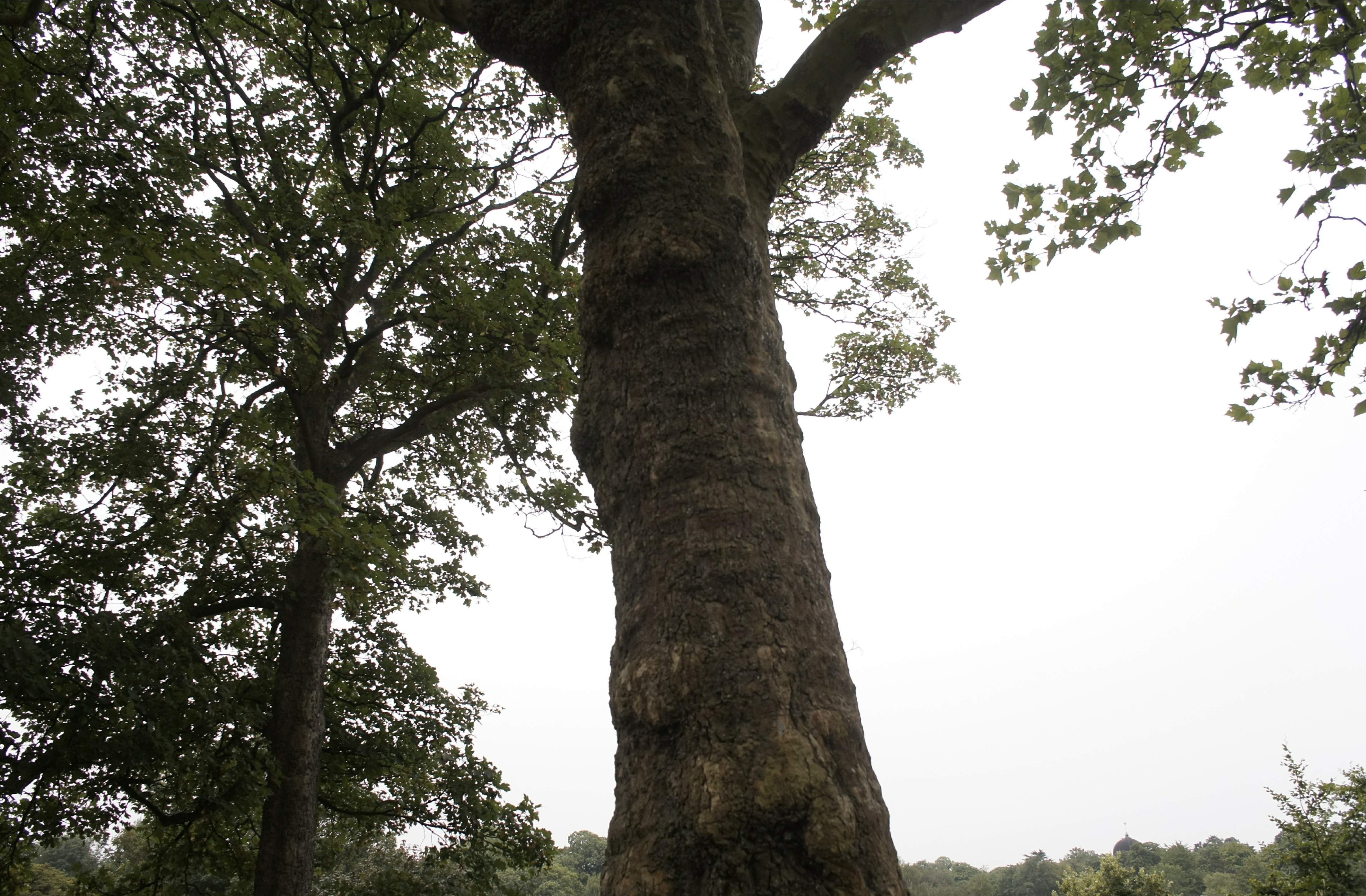}
    \end{subfigure}
    % Single caption and label for the whole figure
    \caption{We find that the maximum geodesic distance between all possible pairs of cameras in a view subset directly correlates with how much of a $360^{\circ}$ scene we can observe. Here we show two farthest views (in terms of $\mathcal{D}_{geodesic}$) of the \emph{treehill} scene for $M = 9$.}
    \label{fig:view_span}
    \vspace{-1.0mm}
\end{figure}

An example plot is shown in Fig \ref{fig_view_split} where we observe the change in $\mathcal{D}^{\mathrm{\pi}}(n)$ and the point cloud size $P_M$ as we steadily increase $n$ by 1. For $M = 9$ in the \emph{treehill} scene (left), we pick the subset corresponding to $n = 17$, the global maxima for $\mathcal{D}^{\mathrm{\pi}}(n)$. Similarly, for $M = 18$ in the \emph{stump} scene (right), we pick the subset corresponding to $n = 31$. Visually, in Fig \ref{fig:view_span}, we observe that this heuristic gives us considerable scene coverage.

\subsection{Sparse 3DGS}
\label{subsec:geometry_baseline_app}

Inspired by recent works \cite{sparsegs, fsgs}, we introduce 4 different regularization techniques to prevent 3DGS from overfitting to the observed few views. 

\paragraph{Tunable Hyperparameters} Similar to \cite{depthreggs}, we observe that the periodic resetting of Gaussian opacities, designed to restrict over-reconstruction in a dense setting, harms generalization to novel views in a sparse setting. Either several Gaussians are removed through pruning or get trapped in local minima of the optimization process, not receiving enough updates to match the given scene. Visually, we observe much lower artifacts in novel views for extremely sparse settings ($M \leq 18$) if opacities $\alpha$ are optimized with no resets during training. The densification threshold for positional gradients - $\tau_{pos}$ is another hyperparameter that controls the overall number of Gaussians. Experimentally, we observe varying $\tau_{pos}$ between 2-5 times the default value for dense setting in 3DGS leads to smoother geometry and prevents overfitting in a sparse setting for $M \leq 54$. We show visual evidence in Fig~\ref{fig:opreset_effect} and \ref{fig:dense_thres_effect} below.

\begin{figure}[!htbp]
    \centering
    \begin{subfigure}[!htbp]{0.4\linewidth}
        \centering
        \includegraphics[width=0.9\linewidth]{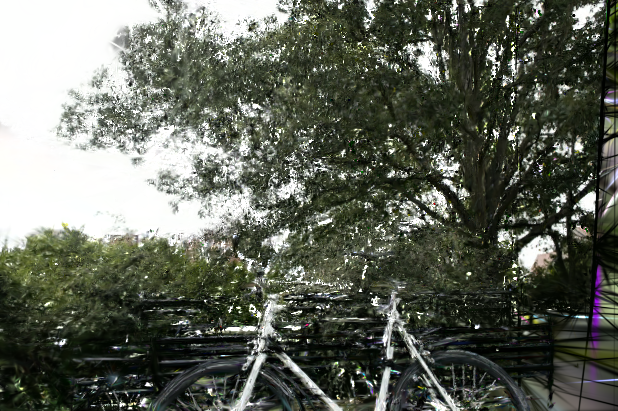}
        \caption{No opacity ($\alpha$) reset for $30k$ iterations.}
    \end{subfigure}
    \begin{subfigure}[!htbp]{0.4\linewidth}
        \centering
        \includegraphics[width=0.9\linewidth]{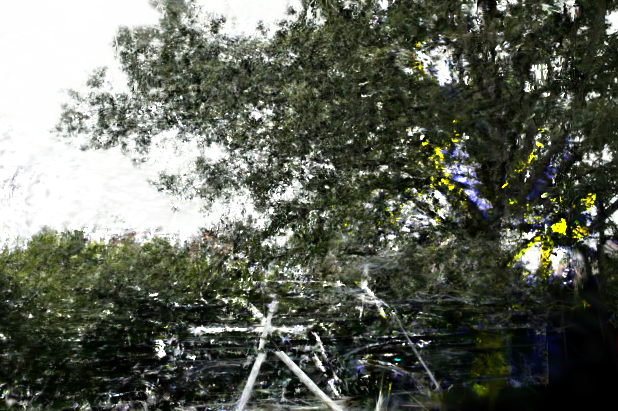}
        \caption{opacity ($\alpha$) reset every $3k$ iterations.}
    \end{subfigure}
    \caption{Periodic reset of the opacity ($\alpha$) of Gaussians has a detrimental effect on novel view synthesis in a sparse reconstruction setting. For example, scene details are more clearly observed in the left image, where there is no $\alpha$ reset for $30k$ iterations. ($M = 9$)}
    \label{fig:opreset_effect}
\end{figure}

\begin{figure}[!htbp]
    \centering
    \begin{subfigure}[!htbp]{0.4\linewidth}
        \centering
        \includegraphics[width=0.9\linewidth]{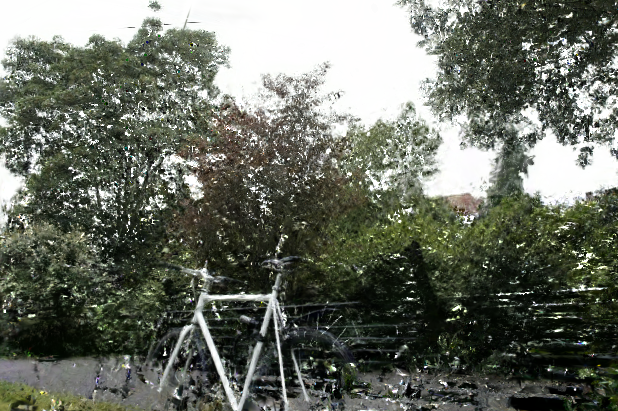}
        \caption{$\tau_{pos} = 0.0002$}
    \end{subfigure}
    \begin{subfigure}[!htbp]{0.4\linewidth}
        \centering
        \includegraphics[width=0.9\linewidth]{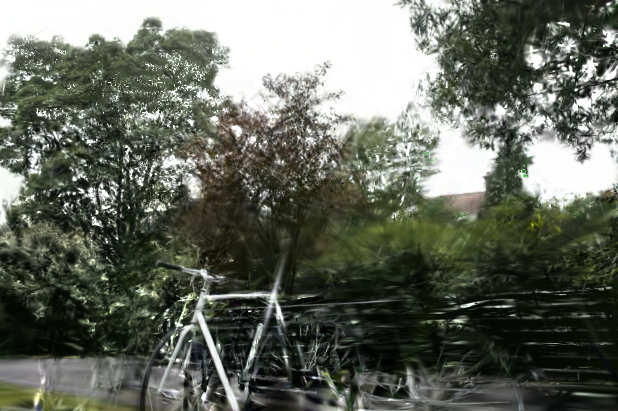}
        \caption{$\tau_{pos} = 0.001$}
    \end{subfigure}
    \caption{Effect of densification threshold ($\tau_{pos}$) for $M=9$ in the $bicycle$ scene. A higher value of $\tau_{pos}$ in a view-constrained scenario limits artifacts and favors smoother geometry. }
    \label{fig:dense_thres_effect}
    \vspace{-4.0mm}
\end{figure}

\paragraph{Depth Regularization} Inspired by depth regularization techniques in NeRF \cite{dsnerf, regnerf} and 3DGS literature \cite{dngaussian}, we introduce dense depth priors from pretrained monocular depth estimators (MDEs) \cite{Ranftl2020}. Due to the severe sparsity of the initial point cloud $P_M$, we observe that this regularization is essential to make Gaussians grow towards a reasonable approximation of the true scene geometry. If $\mathcal{D}_{est}$ is the estimated monocular depth from a training view, and $\mathcal{D}_{ras}$ is the rendered depth, we employ the Pearson correlation coefficient (PCC) as a loss function to align the two depth map distributions \cite{sparsegs, fsgs}. Specifically, we add a third loss term to the 3DGS optimization as:

\begin{equation}
\label{eq_depth_loss}
\begin{aligned}
\mathcal{L}_{depth} = 1 - PCC (\hat{\mathcal{D}}_{ras}, \hat{\mathcal{D}}_{est}) = 1 - \frac{Cov(\hat{\mathcal{D}}_{ras}, \hat{\mathcal{D}}_{est})}{\sqrt{Var(\hat{\mathcal{D}}_{ras})Var(\hat{\mathcal{D}}_{est})}}
\end{aligned}    
\end{equation}

where $\hat{\mathcal{D}}_{ras}$, $\hat{\mathcal{D}}_{est}$ are obtained after normalizing $\mathcal{D}_{ras}$ and $\mathcal{D}_{est}$ to the range $[0, 1]$.

\paragraph{Pseudo Views} As a final regularization, we generate pseudo views using camera interpolation and employ depth regularization as above to improve generalization in unobserved areas. We synthesize novel views on a path by interpolating between a viewpoint $\pi_1$ and its closest view $\pi_2$ in SE3 space. Specifically, we sample $\mathbf{t}$ on a B-spline fitted between the translation matrices and estimate an averaged orientation $\mathbf{R}$ through slerp interpolation on the rotation matrices. This yields an augmented viewpoint $\hat{\pi}$ with shared camera intrinsics and camera extrinsics $(\mathbf{R} \vert \mathbf{t})$. Without a ground truth image for $\hat{\pi}$, we only enable depth regularization on the rendered RGB image. Overall, the optimization objective becomes:

\begin{equation}
\label{eq_overall_obj}
\begin{aligned}
\mathcal{L} = (1 - \lambda_1)\mathcal{L}_1 + \lambda_1\mathcal{L}_{D-SSIM} + \lambda_{depth}\mathcal{L}_{depth} + \lambda_{pseudo}\mathcal{L}_{pseudo} \\
\end{aligned}    
\end{equation}

\begin{figure}[!htbp]
    \centering
    \begin{subfigure}[!htbp]{0.4\linewidth}
        \centering
        \includegraphics[width=0.9\linewidth]{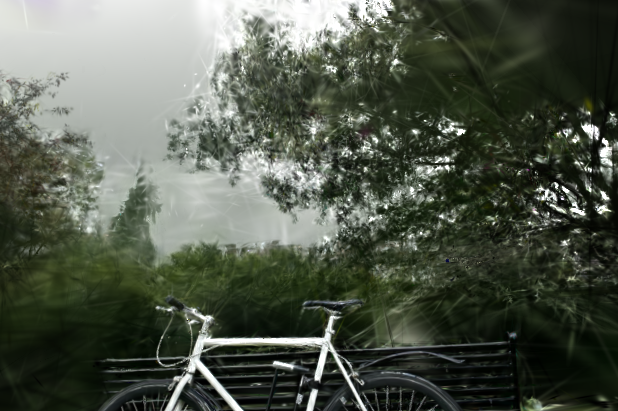}
        \caption{$\lambda_{depth} = 0.05, \lambda_{pseudo} = 0.05$}
    \end{subfigure}
    \begin{subfigure}[!htbp]{0.4\linewidth}
        \centering
        \includegraphics[width=0.9\linewidth]{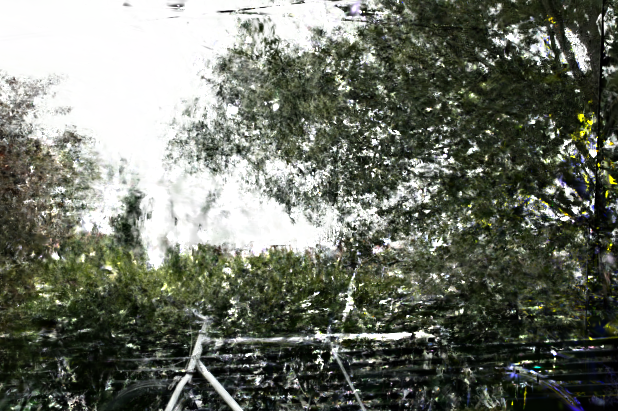}
        \caption{$\lambda_{depth} = \lambda_{pseudo} = 0.0$}
    \end{subfigure}
    \caption{Effect of $\mathcal{L}_{depth}$ and $\mathcal{L}_{pseudo}$ for $M=9$ in $bicycle$ scene. Enabling a depth-based regularization loss during training helps resolve depth ambiguities and learn true scene geometry at nearby novel viewpoints.}
    \label{fig:depth_reg_effect}
    \vspace{-4.0mm}
\end{figure}

\noindent
We switch on $\mathcal{L}_{pseudo}$ after 2000 iterations when the optimization has stabilized and Gaussians have learned at least a coarse scene geometry. We call this improved baseline \emph{Sparse 3DGS}. Visually, we observe in Fig~\ref{fig:depth_reg_effect} that depth regularization through both ground truth and synthesized views helps 3DGS learn plausible geometry of the scene - like the structure of the bicycle and bench while smoothening out a lot of the grainy noise in the background.

\subsection{Baselines for Artifact Removal Module}
\label{subsec:img2img_baselines}

\begin{figure}[!htbp]
    % \captionsetup{justification=centering}
    \centering
        \centering
        \setlength{\lineskip}{0pt} % Reduce vertical space between rows
        \setlength{\lineskiplimit}{0pt} % Reduce vertical space between rows
        \begin{tabular}{@{}c@{}*{6}{>{\centering\small\arraybackslash}p{0.16\linewidth}@{}}}
            & Render & In-painted render & SD Img2Img & Instruct Pix2Pix & Ours & Ground Truth \\

            \raisebox{\height}{\rotatebox[origin=c]{90}{\scriptsize View 1}}
            & \includegraphics[width=\linewidth]{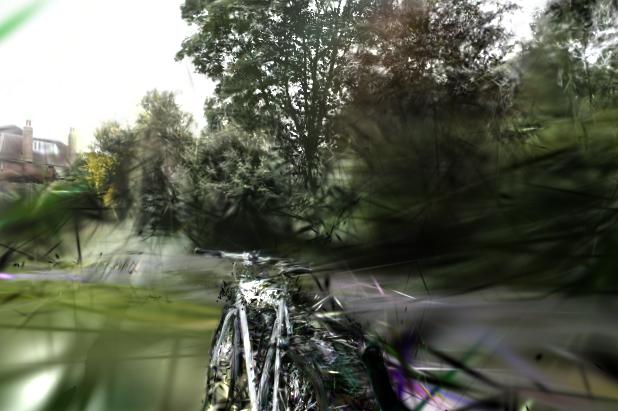} &
            \includegraphics[width=\linewidth]{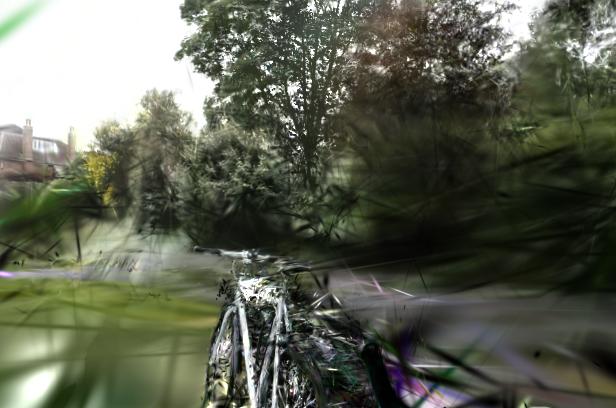} &
            \includegraphics[width=\linewidth]{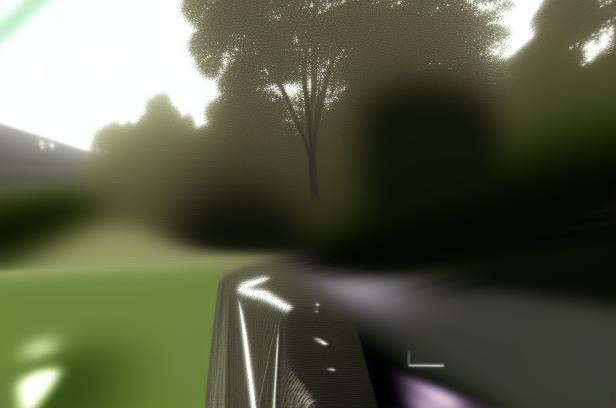} &
            \includegraphics[width=\linewidth]{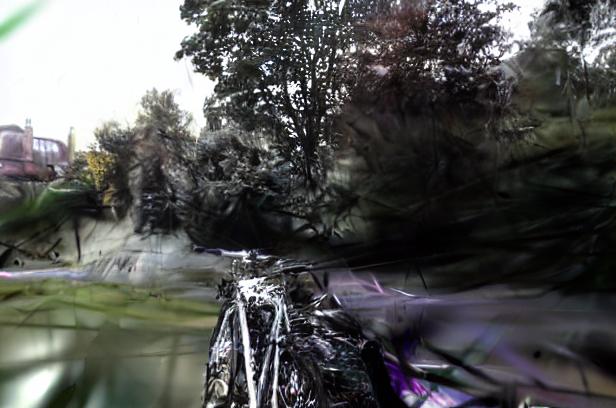} &
            \includegraphics[width=\linewidth]{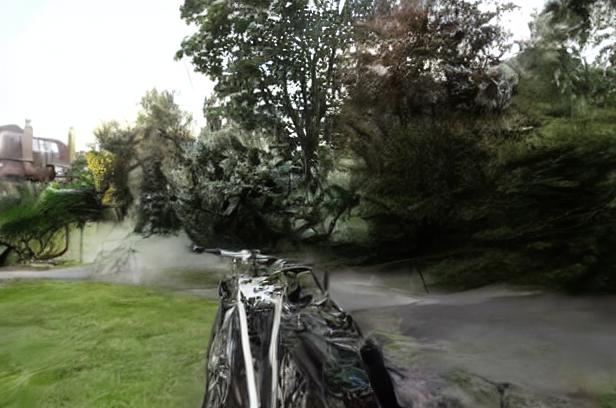} &
            \includegraphics[width=\linewidth]{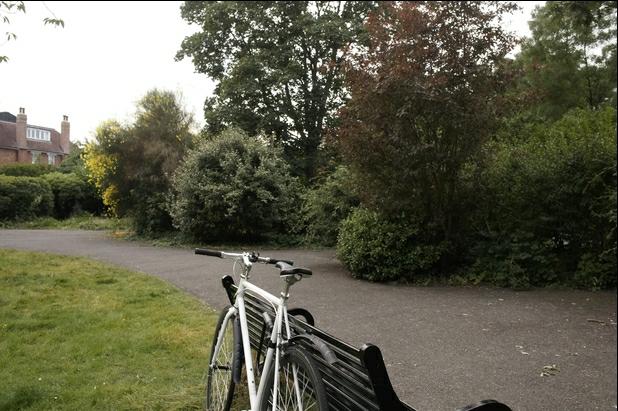}
            \\

            \raisebox{\height}{\rotatebox[origin=c]{90}{\scriptsize View 2}}
            & \includegraphics[width=\linewidth]{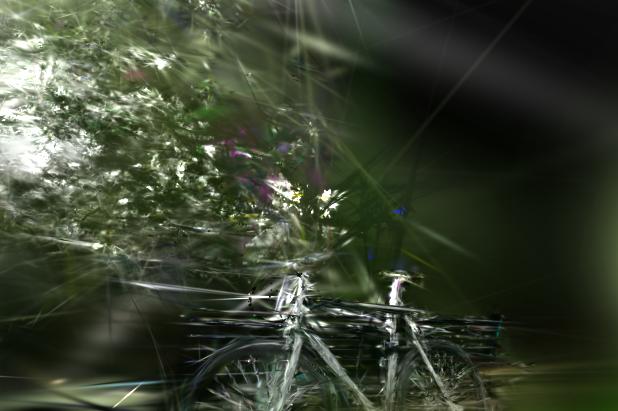} &
            \includegraphics[width=\linewidth]{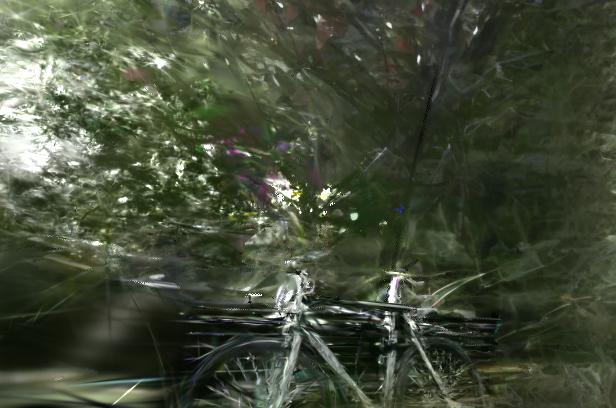} &
            \includegraphics[width=\linewidth]{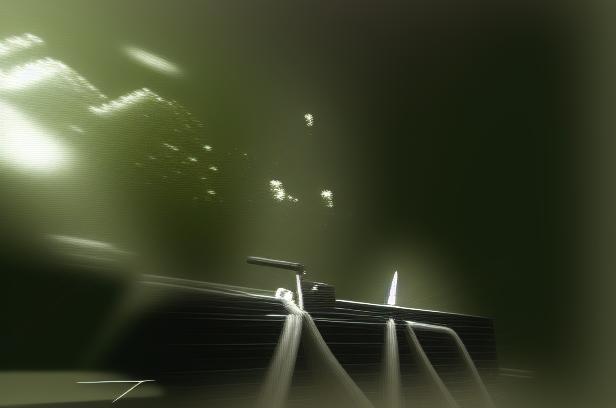} &
            \includegraphics[width=\linewidth]{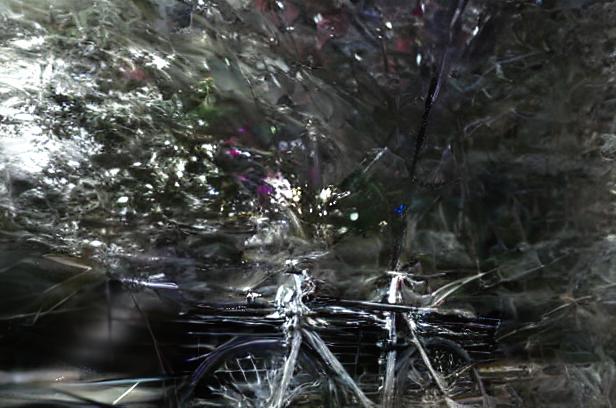} &
            \includegraphics[width=\linewidth]{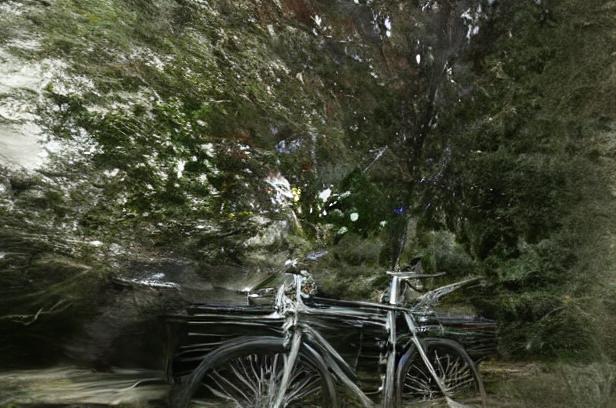} &
            \includegraphics[width=\linewidth]{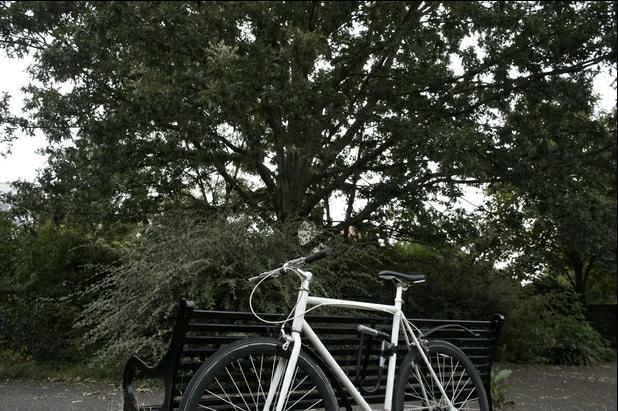}
            \\

            \raisebox{\height}{\rotatebox[origin=c]{90}{\scriptsize View 3}}
            & \includegraphics[width=\linewidth]{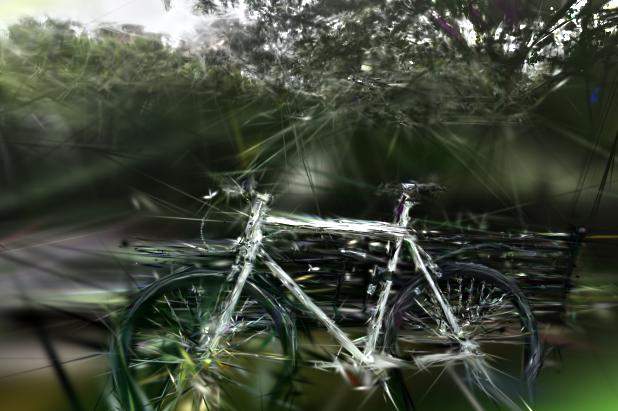} &
            \includegraphics[width=\linewidth]{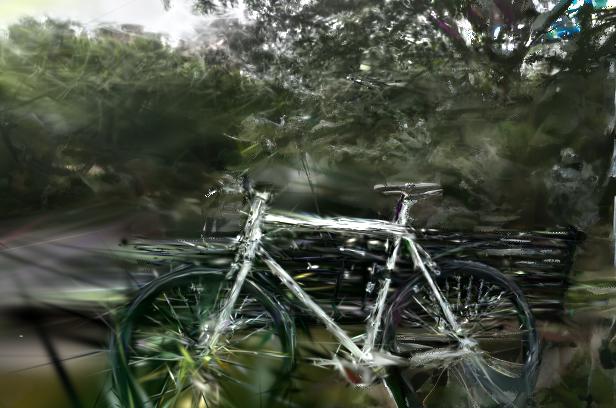} &
            \includegraphics[width=\linewidth]{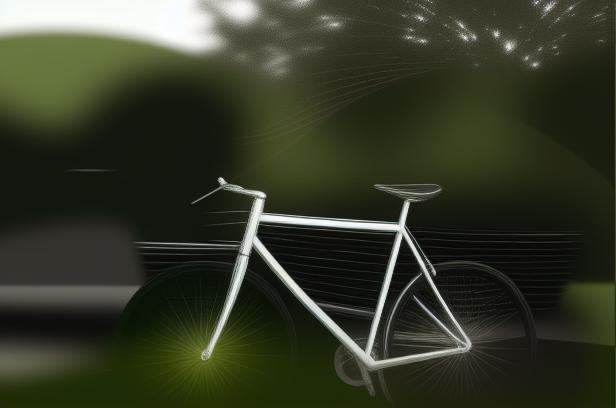} &
            \includegraphics[width=\linewidth]{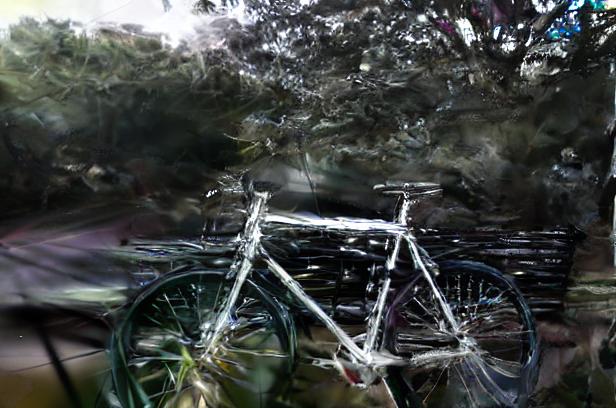} &
            \includegraphics[width=\linewidth]{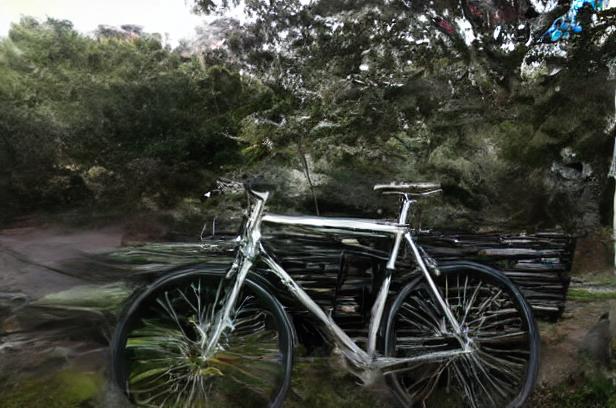} &
            \includegraphics[width=\linewidth]{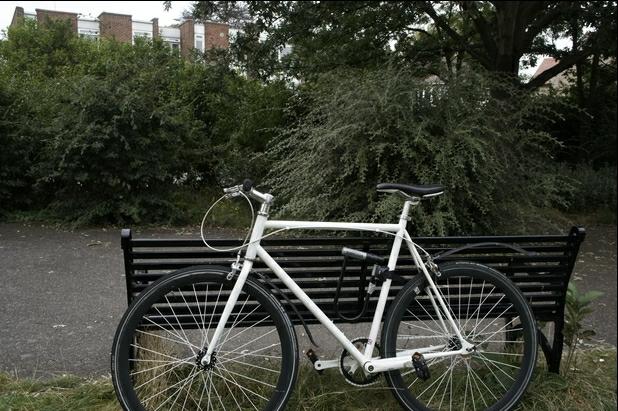}
            \\
            
            \raisebox{\height}{\rotatebox[origin=c]{90}{\scriptsize View 4}}
            & \includegraphics[width=\linewidth]{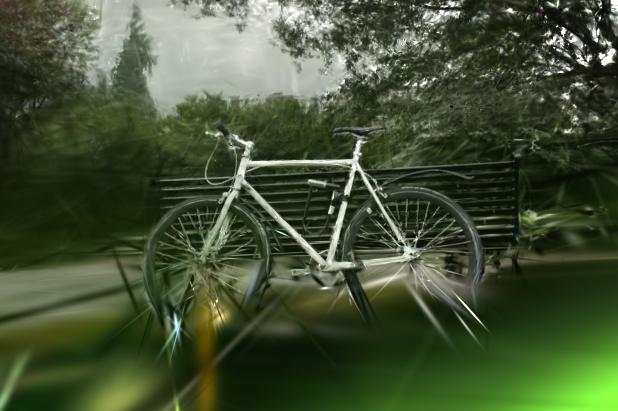} &
            \includegraphics[width=\linewidth]{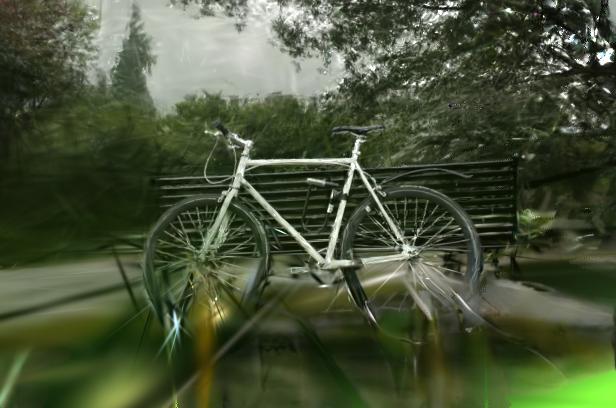} &
            \includegraphics[width=\linewidth]{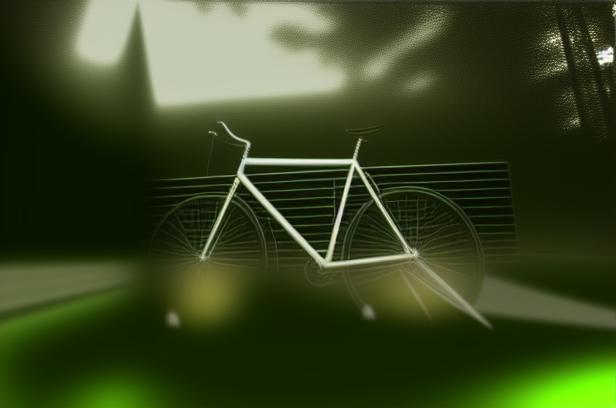} &
            \includegraphics[width=\linewidth]{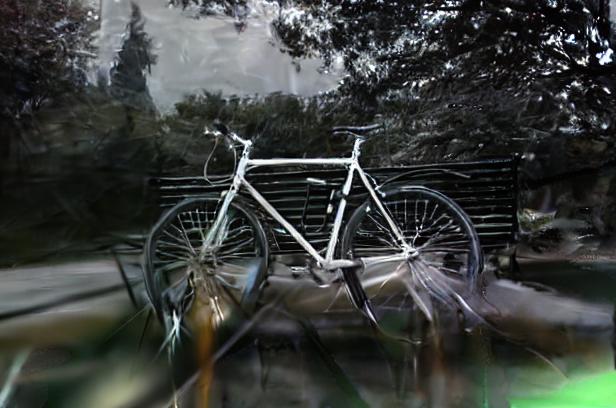} &
            \includegraphics[width=\linewidth]{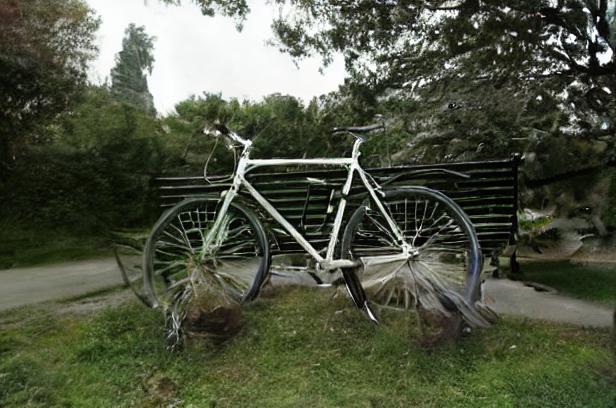} &
            \includegraphics[width=\linewidth]{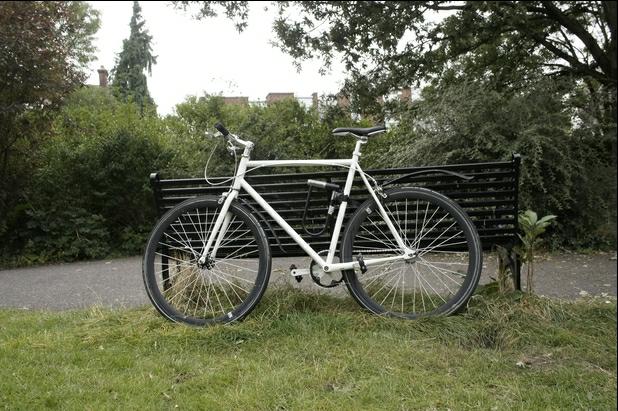}
            \\

        \end{tabular}
        
    \caption{\looseness=-1 \textbf{Qualitative comparison} of \emph{Artifact Removal Module} with image-to-image diffusion baselines. The baselines either produce unrealistic images with background collapse or fail to restore areas in the render corrupted by artifacts. Our finetuned prior, on the contrary, identifies Gaussian artifacts and removes them more efficiently to recover both foreground and background details in the scene.}
    \label{fig:img2img_compare}
    \vspace{-4.0mm}
\end{figure}

In Fig \ref{fig:img2img_compare}, we compare the quality of diffusion samples from our proposed Artifact Removal Module (Sec \ref{subsec:img2img_diffusion}) and two possible baseline solutions based on image-to-image diffusion. For this, we pick the $9$-view split of the \emph{bicycle} scene and run our two-step view synthesis pipeline with the baselines. All $3$ methods take inpainted renders obtained from the Inpainting Module as input (Sec~\ref{subsec:in-painting}).

\paragraph{Stable Diffusion Image-to-Image} Image-to-image diffusion is similar to text-to-image synthesis, but in addition to a text prompt, the final generation is also conditioned on an initial image. We use Stable Diffusion Image-to-Image~\cite{meng2022sdedit} to try and remove Gaussian artifacts from inpainted renders. If \(\mathbf{x}_0\) is the inpainted render, it is first encoded as \(\mathbf{z}_0 = \mathcal{E}(\mathbf{x}_0)\), and then noised corresponding to a certain timestep \(t\) to obtain \(\mathbf{z}_t\). The UNet $\boldsymbol{\epsilon_\theta}$ predicts the noise in \(\mathbf{z}_t\) as:

\begin{equation}
\label{eq:noise_pred_img2img}
\begin{aligned}
    \hat{\epsilon_t} = \boldsymbol{\epsilon_\theta}(\mathbf{z}_t; t, c_\theta(y)),
\end{aligned}
\end{equation}

Here \(y = \)``A photo of [V]'', where \([V]\) is a special token by textual inversion~\cite{gal2022textual} from the \(M\) reference images, to capture visual and semantic details of the scene. We set \(t_{max} = 0.3\) to ensure the output does not deviate too much from \(\mathbf{x}_0\) and keeps at least geometry intact. However, the model has trouble differentiating Gaussian artifacts from image structures and hence cannot recover any scene details lost to artifacts. Instead, it creates images with oversmoothed geometry that look unrealistic and lack any high-frequency details of the original scene.

\paragraph{Instruct Pix2Pix} We try to obtain clean renders with the pretrained Instruct Pix2Pix (IP2P)~\cite{instructpix2pix} checkpoint using the same base instruction as our finetuned artifact removal module. However, IP2P was originally designed to perform image edits like replacing objects, changing the style or setting of a given image, etc, and not an image restoration task we aim to solve. As such, both the conditioning artifact image and the edit instruction are out-of-distribution for IP2P, and hence it fails to identify or remove any of the Gaussian artifacts. \\

Our finetuned artifact removal module, on the other hand, efficiently deals with blur, floaters, and color artifacts in \(\mathbf{x}_0\) and recovers image structures lost to Gaussian artifacts. Artifacts induced at novel views by a 3D Gaussian representation are not observed in images used to train generic image-to-image diffusion models, necessitating the creation of our synthetically curated fine-tuning dataset.

\subsection{Iterative 3DGS}
\label{subsec:iterative_update}

Our iterative schedule fuses clean, in-painted novel view renders with the 3D Gaussians fitted by \emph{Sparse 3DGS}. This part of our method is inspired, in part, by Instruct NeRF2NeRF \cite{instructnerf}. However, unlike their setting, we do not have ground truths for novel views to condition our diffusion models, making it more challenging. We propose to find the best combination of parameters for our algorithm in a simplified setting. \\

We first fit 3D Gaussians to the $9$ view split of the \emph{bicycle} scene and then autoregressively sample closest novel views to fuse corresponding ground truth image with existing scene Gaussians. We argue that this variant should closely match the performance of a \emph{full model} fitted to dense views. Both models see the same training views but with different initializations - SfM point cloud of the entire scene for the \emph{full model} vs 3D Gaussians fitted to $M$ views for \emph{Iterative 3DGS}. The \emph{full model} sees all training views simultaneously, while the iterative version does so incrementally, adjusting the number and position of Gaussians with information from each novel view.

We fuse the heldout $N - M$ novel views over 30k iterations, sampling $m$ views at a time where $m \in \{1, 2, 3\}$. After each incremental update of the training stack, the Gaussians are optimized for $n_k$ iterations where $\{n_k\}_{k=1}^{\lceil \frac{N - M}{m} \rceil}$ is sampled from a schedule, which is a function of the total iterations (30k) and the number of novel views $N - M$. Specifically, $n_k$'s satisfy the following constraint:

\begin{equation}
\label{eq:iterative_update}
\begin{aligned}
    \sum_{k=1}^{\lceil \frac{N - M}{m} \rceil} n_k &= 30000 \\
    n_{k+1} &= f(k) \cdot n_k \hspace{2em} k \in \{1, 2, \ldots, \lceil \frac{N - M}{m} \rceil - 1\}
\end{aligned}
\end{equation}

In the simplest case, when we have a constant number of iterations per train stack update, $f(k) = 1$ and $n_k = \left\lfloor \frac{30000}{\frac{N - M}{m}} \right\rfloor$. For linear and quadratic schedules, $f(k) = a \cdot k$ or $f(k) = a^2 \cdot k$, where $a$ is calculated from the summation constraint (Eq \ref{eq:iterative_update}). A generic solver for Eq \ref{eq:iterative_update} typically finds the largest $a$ for which $\sum_{k} n_k \leq 30000$ is satisfied, and the residual iterations are added to one of the $n_k$'s, usually $n_{\lceil \frac{N - M}{m} \rceil}$. 

\begin{table}[!htbp]
\small
\caption{Ablation Study on design choices of Iterative 3DGS for $M = 9$ in $bicycle$ scene. We fuse ground truth images for heldout novel views and attempt to match the performance of a \emph{full model}.}
\label{tab:iteration_algo_ablation}
\centering
\begin{tabular}{l|ccc}
\toprule
Method & PSNR$\uparrow$ & SSIM$\uparrow$ & LPIPS$\downarrow$\\ 
\hline
$m = 1$ + constant & 21.13 & 0.51 & 0.39 \\ 
$m = 2$ + constant & 21.87 & 0.52 & 0.36 \\ 
$m = 3$ + constant & 21.29 & 0.53 & 0.39 \\
$m = 2$ + cosine & 21.01 & 0.52 & 0.37 \\
$m = 2$ + linear & 22.26 & 0.57 & 0.34 \\
$m = 2$ + quadratic & 22.43 & 0.58 & 0.33 \\
$m = 2$ + quadratic + no $\alpha$-reset & 20.37 & 0.48 & 0.39 \\
$m = 2$ + quadratic + $\eta = 0.95$ & \cellcolor{tabthird}{22.72} & \cellcolor{tabthird}{0.60} & \cellcolor{tabthird}{0.32} \\
$m = 2$ + quadratic + $\eta = 0.97$ ($\mathcal{O}$) & \cellcolor{tabsecond}{22.95} & \cellcolor{tabsecond}{0.63} & \cellcolor{tabsecond}{0.29} \\
\hline
\emph{Full Model} & \cellcolor{tabfirst}{24.34} & \cellcolor{tabfirst}{0.73} & \cellcolor{tabfirst}{0.22} \\ 
\bottomrule
\end{tabular}
\end{table}

\paragraph{Findings} We summarize the findings of our toy experiment in Tab \ref{tab:iteration_algo_ablation}. We experiment with \textit{constant}, \textit{linear}, \textit{quadratic}, and \textit{cosine} schedules and find that a \textit{quadratic} schedule gets us closest to the performance of the \textit{full model}. We find $m = 2$ gives better performance compared to adding just 1 novel viewpoint at a time. We hypothesize that adding 2 viewpoints at a time provides an implicit depth prior that better guides incremental optimization of the Gaussians. Adding $m > 2$ viewpoints does not provide any further improvement. Each Gaussian has a 3D scaling vector $s$ as one of its attributes that determines the scaling of the ellipsoid in 3 axial directions in world space. Our experiments show that shrinking $s$ by a scale adjustment factor $\eta < 1$, once every set of viewpoint(s) is added, impacts generalization to novel views. Intuitively, some of the Gaussians, fitted to sparse views, would have grown towards large volumes to create over-smoothed geometry in empty space. So, shrinking the extent of the ellipsoids, as further observations are introduced, gives more incentive to the \emph{adaptive control module} to clone Gaussians in under-reconstructed regions and reconstruct small-scale geometry better.

Lastly, we find that regularizations introduced in \emph{Sparse 3DGS} are no longer needed during the iterative process due to the growing set of observations and diminishing ambiguities. As such, setting $\tau_{pos} = 0.0002$, resetting opacity($\alpha$) every $3k$ iterations, and removing $\mathcal{L}_{depth}$, $\mathcal{L}_{pseudo}$ gets us closest to the performance of the \emph{full model}. We call the best-performing variant our oracle $\mathcal{O}$, which gives us the experimental maximum for performance on the held-out test set. $\mathcal{O}$ also has similar storage requirements as the \emph{full model} ($3.82M$ v $3.80M$ Gaussians), meaning the \emph{iterative 3DGS} does not lead to a drop in FPS at inference time.

% \subsection{In-Painting Module}
%\begin{figure}[!htbp]
%    \centering
%    \begin{tabular}{c@{}*{3}{>{\centering\arraybackslash}p{0.25\linewidth}@{}}}
%        & Render & Opacity & in-painting Mask \\
%        & \includegraphics[width=\linewidth]{images/in-painting/_DSC8734_render.JPG} &
%        \includegraphics[width=\linewidth]{images/in-painting/_DSC8734_alpha.JPG} &
%        \includegraphics[width=\linewidth]{images/in-painting/_DSC8734_mask.JPG}
%    \end{tabular}
%    \caption{Opacity maps detect empty regions in novel view renders and hence can be repurposed as masks for image in-painting.}
%    \label{fig_opacity_mask}
%\end{figure}

% We find that empty regions in novel view renders with little to no overlap with the scene Gaussians can be automatically detected using the opacity map obtained from the 3DGS rasterizer. Formally, if $I_{\pi} \in \mathbb{R}^{h \times w \times 3}$ and $\alpha_{\pi} \in \mathbb{R}^{h \times w}, 0 \leq \alpha_{\pi} \leq 1$ are the rendered RGB image and opacity map at a novel viewpoint $\pi$ of a 3DGS point cloud $\psi$, we obtain an in-painting mask $\phi$ by binarizing $\alpha_{\pi}$ at a certain threshold $\tau_\phi$ and then inverting it as follows:

% \begin{equation}
% \label{eq:mask_creation}
% \phi = 
% \begin{cases}
%     0 & \text{if } \alpha_{\pi} \geq \tau_\phi, \\
%     1 & \text{if } \alpha_{\pi} < \tau_\phi
% \end{cases}
% \end{equation}

% $\phi = 1$ denotes regions where an LDM needs to in-paint details coherent with the known regions in $I_{\pi}$ ($\phi = 0$). 
%%%%%%%%%%%%%%%%%%%%%%%%%%%%%%%%%%%%%%%%%%%%%%%%%%%%%%%%%%%%

\subsection{Distilling 2D priors to 3D}
\label{subsec:distillation}

Our diffusion priors infer plausible detail in unobserved regions and the iterative update algorithm incrementally grows and updates scene Gaussians by fusing information at novel viewpoints. We initialize the scene with 3D Gaussians fitted by \emph{Sparse 3DGS} to $M$ input views, autoregressively sample closest novel viewpoints (based on SE3 distance), obtain pseudo ground truths for novel views using our combination of diffusion priors, add them to the training stack and optimize for certain iterations (determined by the schedule in Sec \ref{subsec:iterative_update}). At every iteration, we sample either an observed or unobserved viewpoint from the current training stack. We optimize Gaussian attributes using the 3DGS objective for known poses and the SparseFusion \cite{sparsefusion} objective for novel views:

\begin{equation}
\label{eq:recon_obj}
\begin{aligned}
    \mathcal{L}_{sample}(\psi) &= \mathbb{E}_{\pi, t}\Big[w(t) (\Vert I_{\pi} - \hat{I}_{\pi}\Vert_1 + \mathcal{L}_p(I_{\pi}, \hat{I}_{\pi}))\Big] \\
\end{aligned}
\end{equation}

where $\mathcal{L}_p$ is the perceptual loss \cite{lpips}, $w(t)$ a noise-dependent weighting function, $I_{\pi}$ is the 3DGS render at novel viewpoint $\pi$, and $\hat{I}_{\pi}$ is the inpainted, clean render version of $I_{\pi}$ obtained with our cascaded diffusion priors.

\end{document}